\documentclass{bmvc2k}

\usepackage{times}
\usepackage{epsfig}
\usepackage{graphicx}
\usepackage{amsmath}
\usepackage{amssymb}
\usepackage{makecell}
\usepackage{float}

\title{Semantically Adaptive Image-to-image Translation for Domain Adaptation of Semantic Segmentation}

\addauthor{Luigi Musto}{luigi.musto@studenti.unipr.it}{1}
\addauthor{Andrea Zinelli}{andrea.zinelli1@studenti.unipr.it}{1}

\addinstitution{
 University of Parma\\
 Parma, IT
}

\runninghead{Musto, Zinelli}{Semantically Adaptive Image-to-image Translation}

\def\eg{\emph{e.g}\bmvaOneDot}

\begin{document}

\maketitle

\begin{abstract}
	Domain shift is a very challenging problem for semantic segmentation. Any model can be easily trained on synthetic data, where images and labels are artificially generated, but it will perform poorly when deployed on real environments. In this paper, we address the problem of domain adaptation for semantic segmentation of street scenes. Many state-of-the-art approaches focus on translating the source image while imposing that the result should be semantically consistent with the input. However, we advocate that the image semantics can also be exploited to guide the translation algorithm. To this end, we rethink the generative model to enforce this assumption and strengthen the connection between pixel-level and feature-level domain alignment. We conduct extensive experiments by training common semantic segmentation models with our method and show that the results we obtain on the synthetic-to-real benchmarks surpass the state-of-the-art.
\end{abstract}

\section{Introduction}
Deep neural networks for the semantic segmentation of street scenes require to be trained on large and heterogeneous datasets to achieve good accuracy and generalize well. Nevertheless, they still might fail in unseen scenarios and environments (e.g. because of adverse weather). Collecting and manually annotating datasets which can cover all these scenarios requires a huge effort, since the cost of per-pixel labeling is too high.

Simulators, instead, allow to generate unlimited labeled data with low effort. Driving simulators, for example, only require to setup the needed scene and to drive in it to collect the required data. Despite the advances and the photorealism of modern computer graphics, simulators still fail at generating images visually similar to the real ones, which is why models trained naively on such kind of data perform poorly when deployed in the real world.

This setting falls in the more general problem of Domain Adaptation: we have access to two domains, source and target, and we want to exploit the source domain to maximize the accuracy in the target domain, for a given task. When we do not have access to the target labels, but only source ones, we call this Unsupervised Domain Adaptation (UDA). In our case we can formalize the source and target domains to be a synthetic and real dataset.

The most recent solutions to this problem adopt a two-steps approach. The first step is to perform image-to-image translation, where a generative model (\eg CycleGAN~\cite{cyclegan}) or a stylization method~\cite{domstylization} is employed to translate the source images to the target domain. The second step involves training the segmentation network on the translated images, where various methods can be employed to align the features extracted in the two domains.

We focus on improving the first step, making the translation model aware of the task that has to be performed on the resulting images. Different loss functions have been introduced to impose that the task network gives the same result on the two domains~\cite{cycada, crdoco, bdl}. Here, instead, we rethink the generator architecture itself and design it to condition the image translation according to the predicted classes. This not only enhances the capabilities of the generator, but also strengthens the connection between translation and segmentation, since the generated features are connected to the corresponding class by the network itself.

Similar to the related work~\cite{cycada, adaptsegnet, clan, crdoco, bdl} we test our method by adapting both the GTA5~\cite{gta} and SYNTHIA~\cite{synthia} synthetic datasets to Cityscapes~\cite{cityscapes} and show that our results surpass the current state-of-the-art for the commonly used segmentation networks.

\begin{figure}
	\begin{center}
		\includegraphics[width=.8\linewidth]{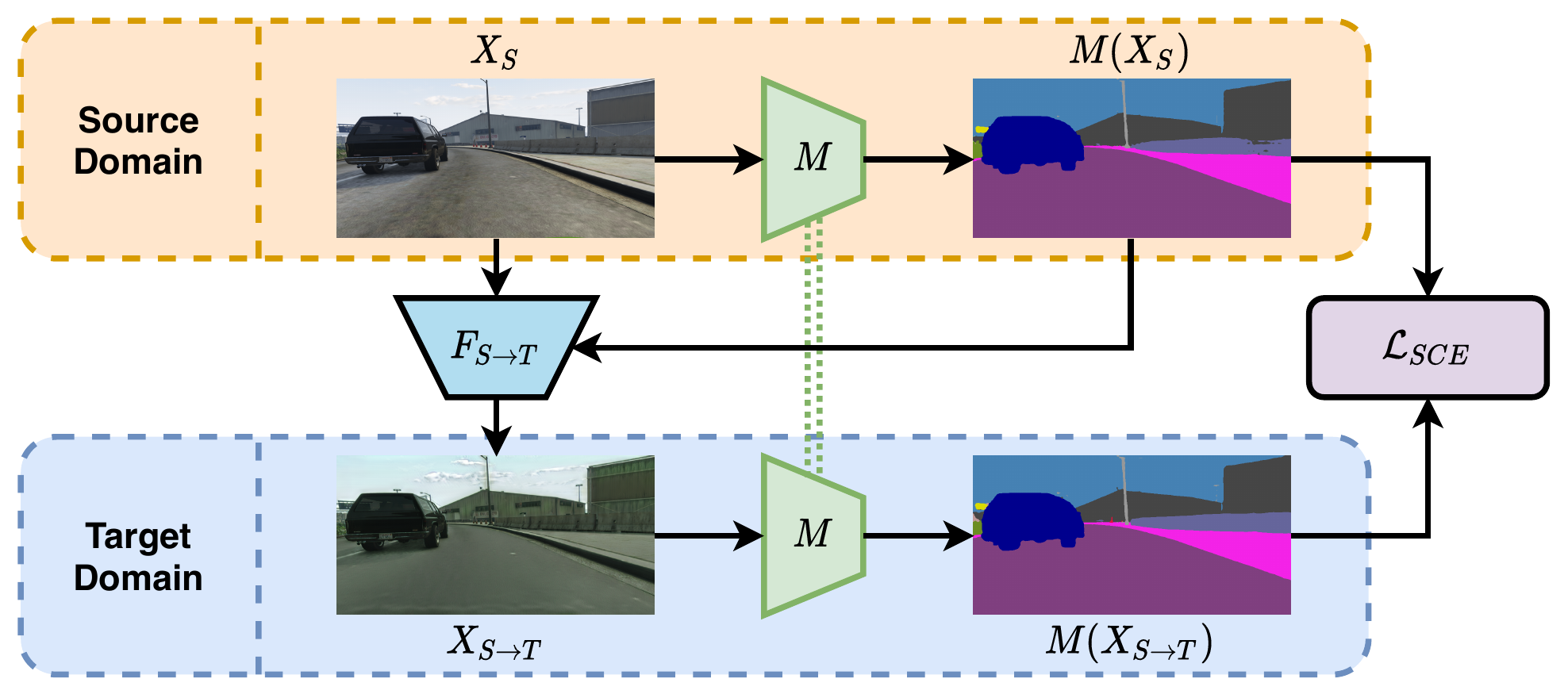}
	\end{center}
	\caption{Core idea of our image-to-image translation system. We use the segmentation network $M$ to get the semantic map $M(X_S)$ from the source image $X_S$. The semantic map acts as guidance for the translation model $F_{S \rightarrow T}$, which translates $X_S$ to the target domain. The translated image $X_{S \rightarrow T}$ is then fed to $M$ again and we get $M(X_{S \rightarrow T})$. Finally we impose the cross-domain semantic consistency by using the Symmetric Cross-Entropy Loss $\mathcal{L}_{SCE}$.}
	\label{fig:i2i-flow}
\end{figure}

\section{Related Work}
Our work can be split into two cooperating parts: UDA for semantic segmentation and image-to-image translation. Here we separately review the most relevant approaches to these tasks, highlighting our contributions.

\paragraph{Unsupervised domain adaptation}
We aim at using synthetic data to perform semantic segmentation on real images, where no labels are available. This can be framed as a problem of UDA, where the main idea is to align the source and target distributions at either feature level, pixel level, or both. This has been applied to image classification~\cite{ganin2015unsupervised, tzeng2015simultaneous, learning_transferable, ganin2016domain, long2016unsupervised, adda, raan} by minimizing the Maximum Mean Discrepancy~\cite{daml, learning_transferable}, measuring the correlation distance~\cite{coral} or with adversarial learning~\cite{adda}.

\begin{figure}
	\centering
	\includegraphics[width=.9\linewidth]{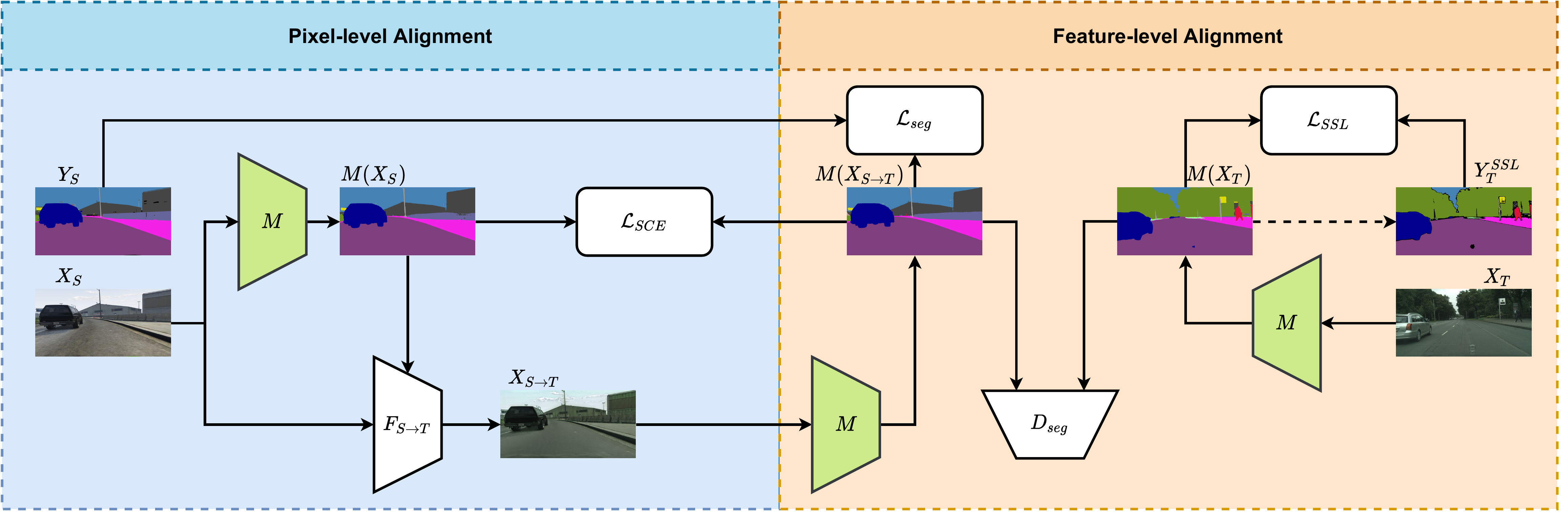}
	\caption{\textbf{Overview of our method}. $M$ is the segmentation network, which is shared for all the steps. $F$ is the translation network. $D_{seg}$ is the segmentation discriminator. The losses are detailed in Section~\ref{sec:method}. We omitted the reconstruction and consistency pipelines, together with the target image discriminator. The pixel-level alignment for $T \rightarrow S$ is symmetric to $S \rightarrow T$. The dashed arrow between $M(X_T)$ and $Y_T^{SSL}$ is used to indicate that the pseudo-label generation is performed offline, before feature-level alignment.}
	\label{fig:full_scheme}
\end{figure}

Semantic segmentation is a much more complex task and the first solution for domain adaptation has been proposed in~\cite{fcnsitw} with global distribution alignment at feature level. Curriculum learning was used in~\cite{cda}, where the authors proposed first to learn the global distribution of the image and the local distribution of superpixels, and then train the segmentation network according to these properties. Global feature alignment was also adopted in~\cite{adaptsegnet}, where different discriminators are employed for features at different levels. Other works introduced class-wise adversarial learning~\cite{nomoredis} and pseudo-labels~\cite{nomoredis, cbst, bdl}. CLAN~\cite{clan} improves feature level alignment by reweighting the adversarial loss with a discrepancy map based on categories. Feature level alignment, however, is not enough to adapt to different domains, which is why the most recent approaches~\cite{domstylization, cycada, dcan, crdoco, bdl} introduced also pixel-level alignment. CyCADA~\cite{cycada} trains the segmentation network on images translated with CycleGAN~\cite{cyclegan} and a semantic consistency loss. DCAN~\cite{dcan} adopts a custom image-to-image translation network and performs feature alignment both in the translation and in the segmentation step. CrDoCo~\cite{crdoco} uses a cross-domain consistency loss to improve the translation. Similarly BDL~\cite{bdl} links translation and segmentation with a perceptual loss, where the training is iterated to gradually improve both tasks.

Our work builds on top of these ideas, but we rethink the generator architecture to condition the image-to-image translation with the semantic guidance of the segmentation network.

\paragraph{Image-to-image translation}
In order to translate synthetic images into real looking ones without using paired couples, the most common approach is to use Generative Adversarial Networks~\cite{gan}. By learning how to trick the discriminator, the generator network becomes able to generate images aligned with the target distribution.

Nevertheless, only with the introduction of the cycle consistency~\cite{cyclegan} the generated images look realistic. UNIT~\cite{unit} develops a more complex assumption: by combining VAEs~\cite{vae_kingma, rezende2014stochastic} with CoGANs~\cite{cogan}, they enforce that two domains share a common latent space which can be used to move from one domain to the other and back. This approach evolves in MUNIT~\cite{munit}, where the shared latent space is formalized as the content and combined with the target style to generate multimodal realistic outputs. 

\paragraph{Normalization layers}
The key insight for image-to-image translation is in the ability to disentangle style and content. In fact, in order to move from one domain to the other, one has to be able to change the style while preserving the image content.

It has been noted~\cite{adain} that the most effective way to swap styles is by using normalization layers. Batch Normalization~\cite{batchnorm} has been used in~\cite{texture}, but~\cite{improved-texture} found that replacing it with Instance Normalization (IN)~\cite{instancenorm} leads to significant improvements. IN works in the feature space in the same way Contrast Normalization works in the pixel space, which makes it much more effective. A more general approach has been introduced by Adaptive IN (AdaIN), which computes the affine transformation from a style input. UNIT~\cite{unit} uses IN to swap the source and target style. Instead of performing a global translation with IN, we exploit the task network to translate each region of the image according to its semantic meaning. To this end, we choose to denormalize the generator activations with the SPADE layer~\cite{spade}, giving a result that naturally cooperates with the learning of the semantic segmentation task.

\begin{figure}
	\centering
	\begin{tabular}{@{\hskip3pt}c@{\hskip3pt}c@{\hskip3pt}c@{\hskip3pt}c@{\hskip3pt}c}
		\bmvaHangBox{\includegraphics[width=.19\textwidth]{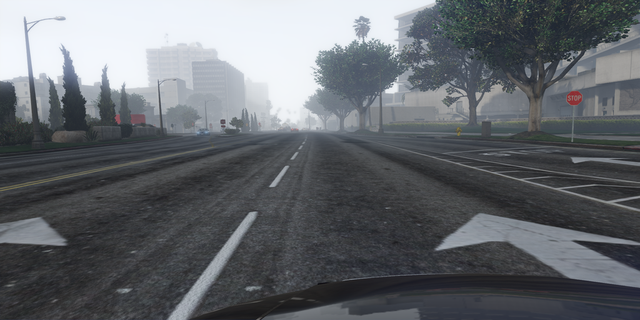}}&
		\bmvaHangBox{\includegraphics[width=.19\textwidth]{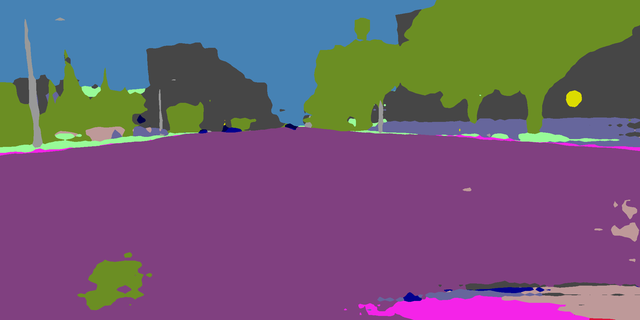}}&
		\bmvaHangBox{\includegraphics[width=.19\textwidth]{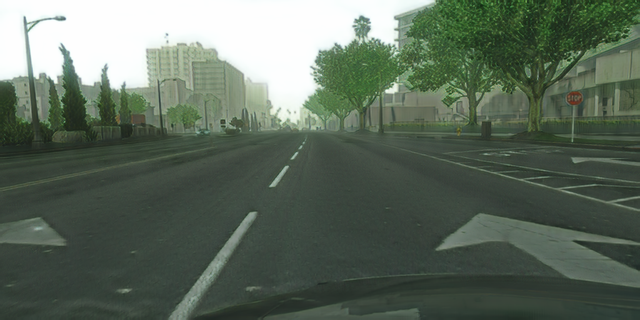}}&
		\bmvaHangBox{\includegraphics[width=.19\textwidth]{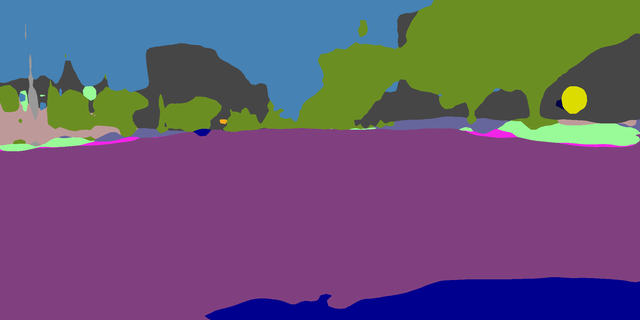}}&
		\bmvaHangBox{\includegraphics[width=.19\textwidth]{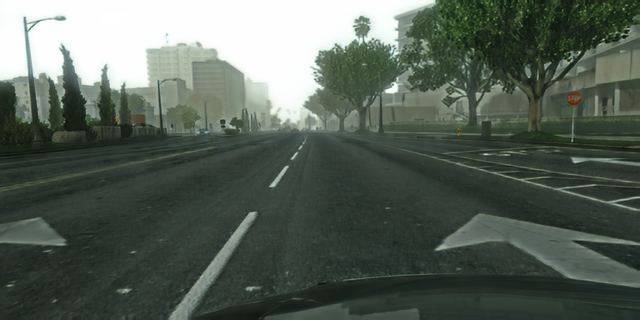}}\\
		\bmvaHangBox{\includegraphics[width=.19\textwidth]{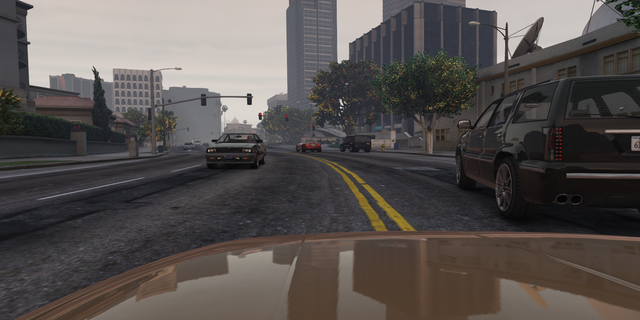}}&
		\bmvaHangBox{\includegraphics[width=.19\textwidth]{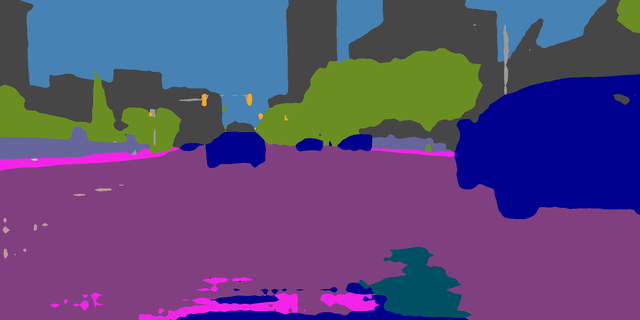}}&
		\bmvaHangBox{\includegraphics[width=.19\textwidth]{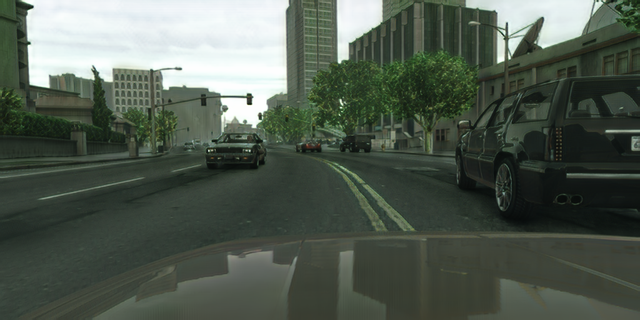}}&
		\bmvaHangBox{\includegraphics[width=.19\textwidth]{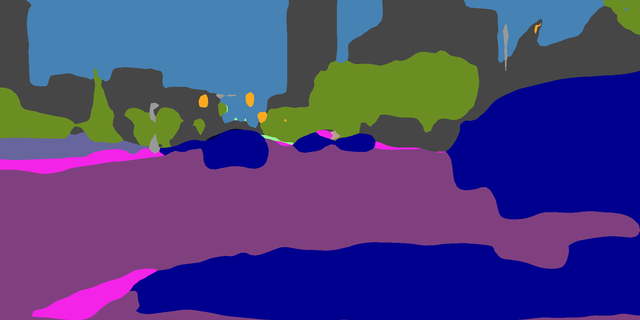}}&
		\bmvaHangBox{\includegraphics[width=.19\textwidth]{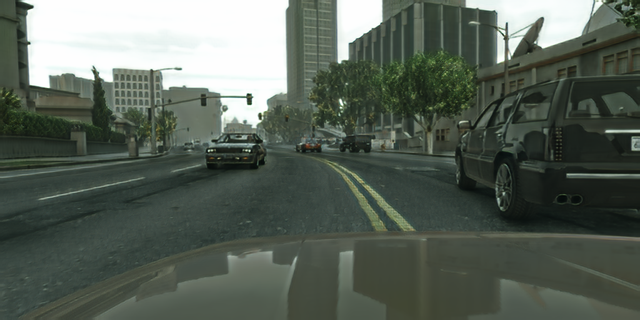}}\\
		\bmvaHangBox{\includegraphics[width=.19\textwidth]{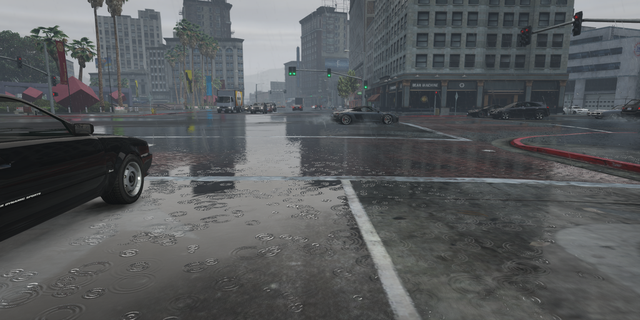}}&
		\bmvaHangBox{\includegraphics[width=.19\textwidth]{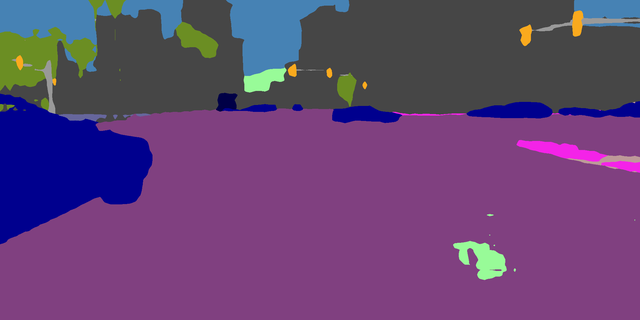}}&
		\bmvaHangBox{\includegraphics[width=.19\textwidth]{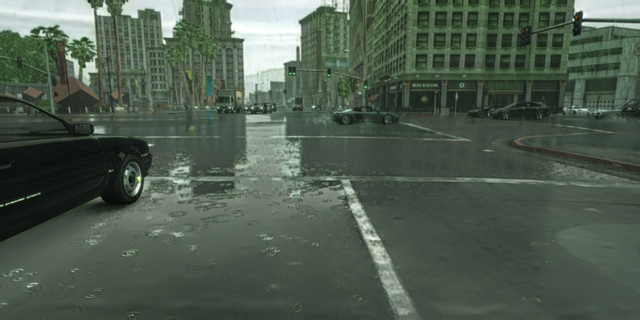}}&
		\bmvaHangBox{\includegraphics[width=.19\textwidth]{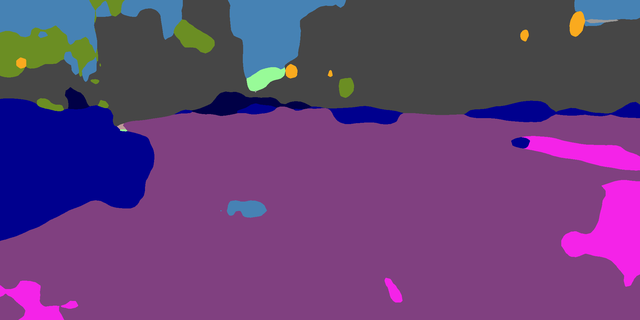}}&
		\bmvaHangBox{\includegraphics[width=.19\textwidth]{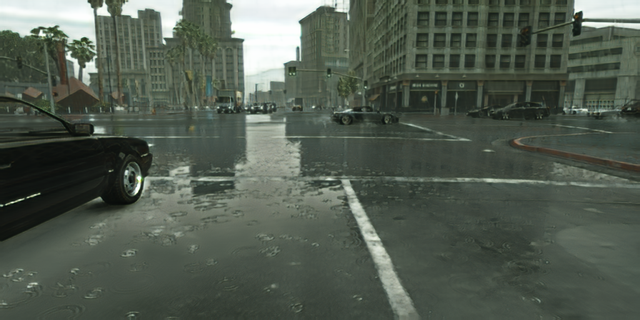}}\\
		\bmvaHangBox{\includegraphics[width=.19\textwidth]{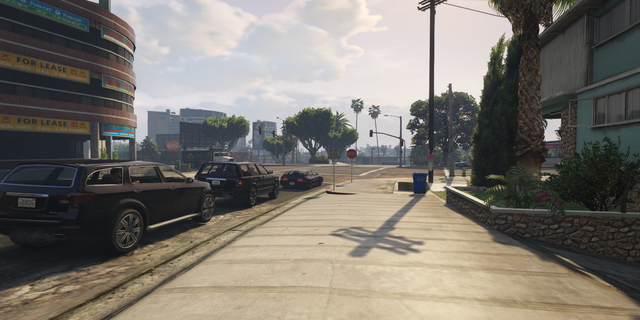}}&
		\bmvaHangBox{\includegraphics[width=.19\textwidth]{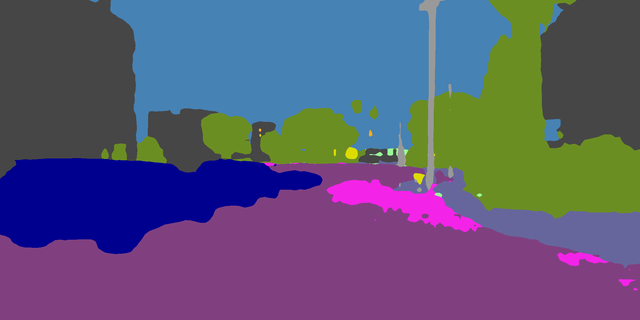}}&
		\bmvaHangBox{\includegraphics[width=.19\textwidth]{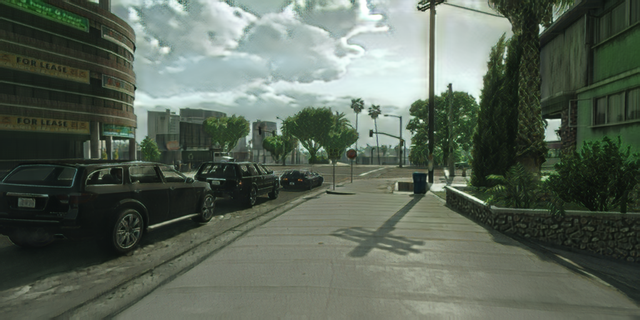}}&
		\bmvaHangBox{\includegraphics[width=.19\textwidth]{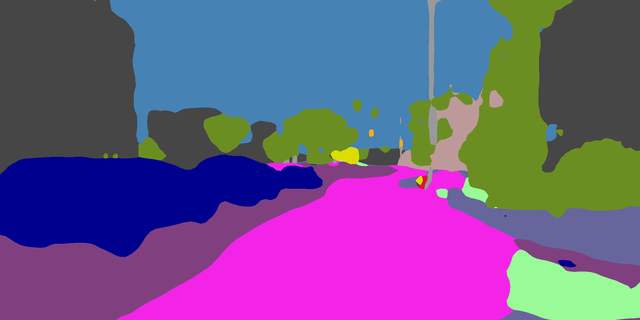}}&
		\bmvaHangBox{\includegraphics[width=.19\textwidth]{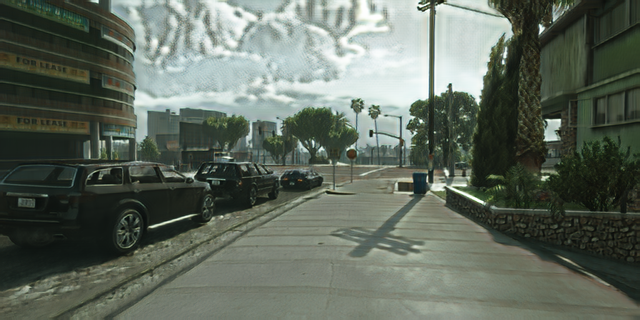}}\\
		GTA5 sample&GTA5 pred (D)&GTA5 $\rightarrow$ CS (D)&GTA5 pred (F)&GTA5 $\rightarrow$ CS (F)
	\end{tabular}
	\caption{\textbf{Translation from GTA5~\cite{gta} to Cityscapes~\cite{cityscapes}.} We take a sample from GTA5, get the predicted segmentation using $M$, and generate $X_{S \rightarrow T}$. We present the results obtained with both DeepLabV2~\cite{deeplab} and FCN8s~\cite{fcn} used as semantic guidance.}
	\label{fig:gta2cs}
\end{figure}

\section{Method}
\label{sec:method}
Our objective is to train a deep neural network $M$ to perform semantic segmentation on a target (real) dataset $T$. We assume we only have the target images $X_T$ without the target labels $Y_T$. In order to do this, we use synthetic data from a source (synthetic) dataset $S$, where we have both the images $X_S$ and the labels $Y_S$.

This problem setting is UDA for semantic segmentation, which means that we want to reduce the domain shift caused by the difference in visual appearance of the two domains.

As depicted in \figurename~\ref{fig:full_scheme}, we follow the recent work~\cite{cycada, crdoco, bdl} and take advantage of both pixel-level and feature-level alignment to reduce the domain shift. We can see them as two separate subtasks, but we will also show how they actually need to cooperate to improve each other in the following sections.

\subsection{Pixel-level alignment}
\label{sec:pixel-level}

\begin{figure}
	\centering
	\begin{tabular}{@{\hskip3pt}c@{\hskip3pt}c@{\hskip3pt}c@{\hskip3pt}c}
		\includegraphics[width=.24\linewidth]{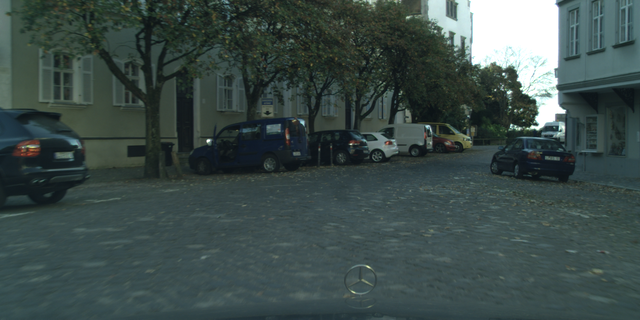}&
		\includegraphics[width=.24\linewidth]{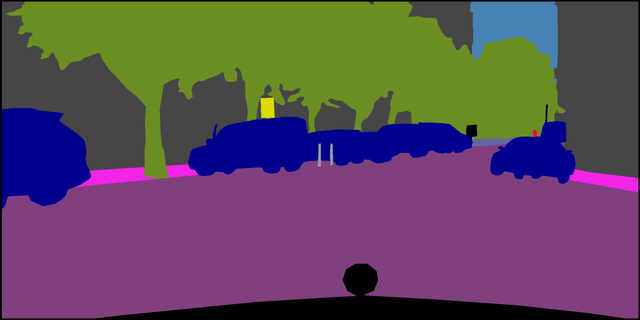}&
		\includegraphics[width=.24\linewidth]{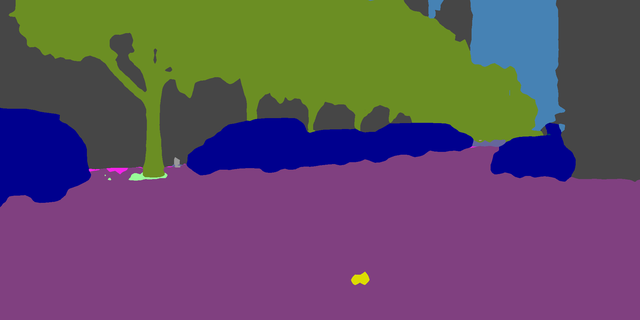}&
		\includegraphics[width=.24\linewidth]{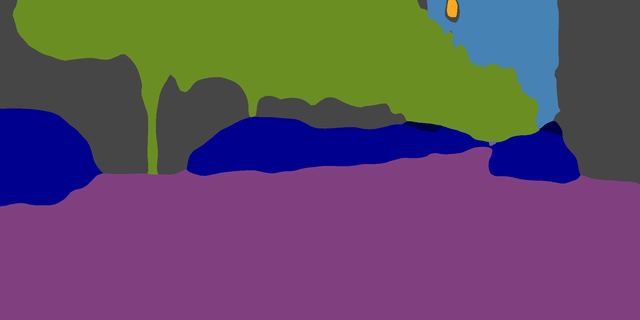}\\
		\includegraphics[width=.24\linewidth]{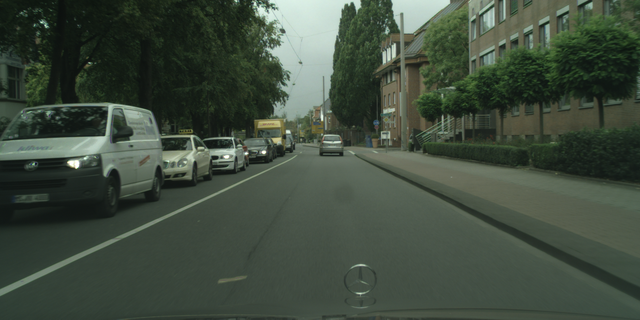}&
		\includegraphics[width=.24\linewidth]{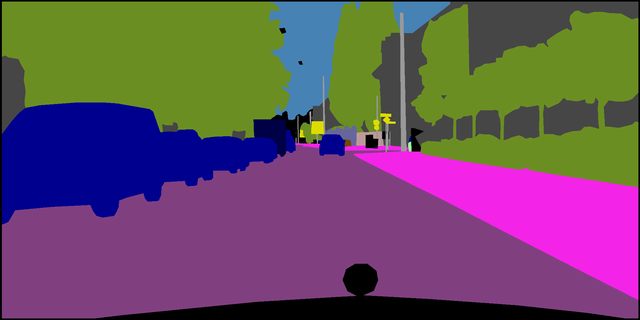}&
		\includegraphics[width=.24\linewidth]{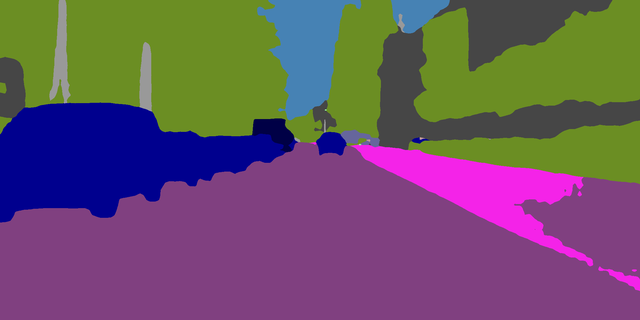}&
		\includegraphics[width=.24\linewidth]{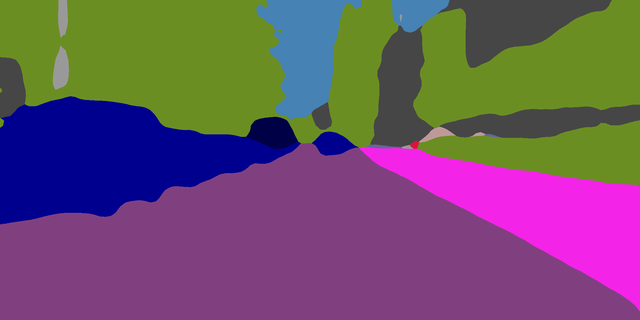}\\
		Test sample&Ground truth&GTA5 $\rightarrow$ CS w/ D&GTA5 $\rightarrow$ CS w/ F
	\end{tabular}
	\caption{\textbf{Semantic Segmentation adapted from GTA5~\cite{gta} to Cityscapes~\cite{cityscapes}.} We take a sample $X_T$ from the Cityscapes validation set and show the segmentation predictions $M(X_T)$ of both the adapted DeepLabV2~\cite{deeplab} and FCN8s~\cite{fcn} networks.}
	\label{fig:seg_comparison}
\end{figure}

For pixel-level alignment, we make use of an image-to-image translation network $F_{S \rightarrow T}$, which learns through adversarial training to visually align $X_S$ to $X_T$ by generating $X_{S \rightarrow T}=F_{S \rightarrow T}(X_S)$. Some visual examples of the results of this pipeline are depicted in \figurename~\ref{fig:gta2cs}. Inspired by~\cite{unit}~\cite{munit}, we assume that $S$ and $T$ share a common latent space $Z$ and design two coupled GANs to train the desired system.

The training objective for the image translation model is comprised of several loss functions computed as the sum of two components, one per domain. The final objective is then:

\begin{equation}
	\begin{aligned}
		\mathcal{L} &=
		\lambda_{recon} \mathcal{L}_{recon} +
		\lambda_{GAN} \mathcal{L}_{GAN} +
		\lambda_{CC_I} \mathcal{L}_{CC_I} +
		\lambda_{CC_H} \mathcal{L}_{CC_H} +
		\lambda_{SCE} \mathcal{L}_{SCE}
	\end{aligned}
\end{equation}

\paragraph{Image reconstruction}
We have an encoder for each domain, $E_S$ and $E_T$, coupled with a corresponding generator for each domain, $G_S$ and $G_T$, to form two Autoencoders. The encoders extract the latent code $z \sim Z$, which is fed to the generators along with the semantic features predicted by $M$. Therefore, a translated image is indicated as $x_{A \rightarrow B} = G_B(E_A(x_A), M(x_A))$ and the image reconstruction loss is:

\begin{equation}
		\mathcal{L}_{recon}^S = \mathbb{E}_{x_S \sim X_S} \,\, [||x_{S \rightarrow S} \, \, - x_S \:||] \qquad \qquad \mathcal{L}_{recon}^T = \mathbb{E}_{x_T \sim X_T} [||x_{T \rightarrow T} - x_T||]
\end{equation}

\paragraph{Adversarial loss}
By combining $E_S$ with $G_T$ and vice versa, we get the actual translation models $F_{S \rightarrow T}$ and $F_{T \rightarrow S}$, which are trained in an adversarial fashion to trick the corresponding discriminators $D_T$ and $D_S$:

\begin{equation}
	\begin{aligned}
		\mathcal{L}_{GAN}^S &= \frac{1}{2} \mathbb{E}_{x_S \sim X_S} \, \, [(D_S(x_S))^2] \, \, + \frac{1}{2} \mathbb{E}_{x_T \sim X_T}[(D_S(x_{T \rightarrow S}) - 1)^2]\\
		\mathcal{L}_{GAN}^T &= \frac{1}{2} \mathbb{E}_{x_T \sim X_T} [(D_T(x_T))^2] + \frac{1}{2} \mathbb{E}_{x_S \sim X_S}[(D_T(x_{S \rightarrow T}) - 1)^2]
	\end{aligned}
\end{equation}

\paragraph{Cycle consistency}
By combining $F_{S \rightarrow T}$ with $F_{T \rightarrow S}$ and viceversa we can now apply the cycle consistency loss to images and latent spaces:

\begin{equation}
	\begin{aligned}
		&\mathcal{L}_{CC_{I}}^S = \mathbb{E}_{x_S \sim X_S} [||x_{S \rightleftarrows T} - x_S||] \qquad \qquad \mathcal{L}_{CC_{I}}^T = \mathbb{E}_{x_T \sim X_T} [||x_{T \rightleftarrows S } - x_T||]\\
	    &\mathcal{L}_{CC_{H}}^S = \mathbb{E}_{z_S \sim Z} [||z_{S \rightarrow T} - z_S||] \qquad \qquad \; \mathcal{L}_{CC_{H}}^T = \mathbb{E}_{z_T \sim Z} [||z_{T \rightarrow S} - z_T||]\\
	\end{aligned}
\end{equation}

where $z_{S \rightarrow T}$ refers to the latent space extracted by $E_T$ from $x_{S \rightarrow T}$ and viceversa.

\paragraph{Symmetric cross-entropy}
Finally, we impose that the segmentation predicted for the translated image has to be consistent with the one predicted for the original one through a symmetric cross-entropy loss, which is made of two contributions. For the $S \rightarrow T$ case, the first contribution assumes that $M(x_{S \rightarrow T})$ is the ground truth label and tries to align $M(x_S)$ with it. The second contribution assumes that $M(x_S)$ is the ground truth label and tries to align $M(x_{S \rightarrow T})$ with it. The $T \rightarrow S$ case is symmetrical to the first one.

\begin{equation}
\label{eq:pixel-level}
	\begin{aligned}
		\mathcal{L}_{SCE}^S = &-\mathbb{E}_{x_S \sim X_S}\,\,[M(x_{S \rightarrow T}) \log M(x_S)\,] - \mathbb{E}_{x_S \sim X_S}\,\,[M(x_S) \log M(x_{S \rightarrow T})\,]\\
		\mathcal{L}_{SCE}^T = &-\mathbb{E}_{x_T \sim X_T}[M(x_{T \rightarrow S}) \log M(x_T)] - \mathbb{E}_{x_T \sim X_T}[M(x_T) \log M(x_{T \rightarrow S})]
	\end{aligned}
\end{equation}

\subsection{Semantically adaptive generator}
\label{sec:sem-ad-gen}

Recent generator architectures~\cite{munit, funit, stylegan} make use of AdaIN to remove the source style and inject the target one. However, we observe that the global denormalization performed by AdaIN might be suboptimal for the image translation task. This is why we redesigned our generator to adaptively denormalize each pixel based on its semantics.

We use $M$ to extract a segmentation map $m \in \mathbb{R}^{B \! \times \! C \! \times \! H \! \times \! W}$ from the input image, where $C$ is the number of classes. When feeding it to the generator, we choose to represent this semantic guidance as the unnormalized output of $M$. In the supplementary material we detail the reasons behind this choice and the other possibilities.

Given an input activation $x \in \mathbb{R}^{B \! \times \! C' \! \times \! H' \! \times \! W'}$, $m$ is resized to $H' \! \times \! W'$ and fed to the SPADE layer, which outputs $\gamma, \beta \in \mathbb{R}^{B \! \times \! C' \! \times \! H' \! \times \! W'}$.  We then normalize $x$ by using Instance Normalization and use $\gamma$ and $\beta$ to denormalize it:

\begin{equation}
	y_{b, c, h, w} = \gamma_{b, c, h, w} \frac{x_{b, c, h, w} - \mu_{b, c}}{\sigma_{b, c}} + \beta_{b, c, h, w}
\end{equation}

\subsection{Analysis}
Pixel-level alignment has given a great boost to the research in UDA problems, but the gap with the performance achievable with full supervision is still huge. We believe that the image translation methods still need a lot of improvements and this is why we focused on redesigning the generator to include a semantic conditioning.
Our claim is that adaptively denormalizing each pixel based on its class allows the translation model to produce results which are better for domain adaptation, since each region gets injected with features that are more consistent with its semantic. This connection strengthens the bridge with feature-level alignment (see \figurename~\ref{fig:i2i-flow}), which before our work was induced only by consistency losses.

\subsection{Feature-level alignment}
For feature-level alignment, we train $M$ on $X_T$ and $X_{S \rightarrow T}$ by combining supervision on $X_{S \rightarrow T}$, self-supervision on $X_T$ and adversarial learning. The loss, in this case, is given by

\begin{equation}
\label{eq:feature-level}
\mathcal{L} = \lambda_{seg} \mathcal{L}_{seg} + \lambda_{SSL} \mathcal{L}_{SSL} + \lambda_{adv} \mathcal{L}_{adv}
\end{equation}

We set $\lambda_{seg}=1$, $\lambda_{SSL}=1$, $\lambda_{adv}=10^{-3}$ for DeepLabV2, $\lambda_{adv}=10^{-4}$ for FCN8s and use the same optimization hyperparameters of~\cite{bdl} to train both networks.

\begin{table}
	\resizebox{\linewidth}{!}{
		\begin{tabular}{cccccccccccccccccccccc}
			\hline
			\multicolumn{21}{c}{GTA5 $\rightarrow$ Cityscapes}\\
			\hline
			\\
			Method & \rotatebox{90}{Arch.} & \rotatebox{90}{road} & \rotatebox{90}{sidewalk} & \rotatebox{90}{building} & \rotatebox{90}{wall} & \rotatebox{90}{fence} & \rotatebox{90}{pole} & \rotatebox{90}{light} & \rotatebox{90}{sign} & \rotatebox{90}{veget} & \rotatebox{90}{terrain} & \rotatebox{90}{sky} & \rotatebox{90}{person} & \rotatebox{90}{rider} & \rotatebox{90}{car} & \rotatebox{90}{truck} & \rotatebox{90}{bus} & \rotatebox{90}{train} & \rotatebox{90}{mbike} & \rotatebox{90}{bike} & mIoU \\ \hline
			Cycada~\cite{cycada} & D & 86.7 & 35.6 & 80.1 & 19.8 & 17.5 & \textbf{38.0} & 39.9 & \textbf{41.5} & 82.7 & 27.9 & 73.6 & \textbf{64.9} & 19 & 65.0 & 12.0 & 28.6 & 4.5 & 31.1 & \textbf{42.0} & 42.7 \\
			AdaptSegNet~\cite{adaptsegnet} & D & 86.5 & 25.9 & 79.8 & 22.1 & 20.0 & 23.6 & 33.1 & 21.8 & 81.8 & 25.9 & 75.9 & 57.3 & 26.2 & 76.3 & 29.8 & 32.1 & \textbf{7.2} & 29.5 & 32.5 & 41.4 \\
			DCAN~\cite{dcan} & D & 85.0 & 30.8 & 81.3 & 25.8 & 21.2 & 22.2 & 25.4 & 26.6 & 83.4 & 36.7 & 76.2 & 58.9 & 24.9 & 80.7 & 29.5 & 42.9 & 2.5 & 26.9 & 11.6 & 41.7 \\
			CLAN~\cite{clan} & D & 87.0 & 27.1 & 79.6 & 27.3 & 23.3 & 28.3 & 35.5 & 24.2 & 83.6 & 27.4 & 74.2 & 58.6 & 28.0 & 76.2 & 33.1 & 36.7 & 6.7 & 31.9 & 31.4 & 43.2 \\
			BDL~\cite{bdl} & D & 91.0 & \textbf{44.7} & 84.2 & 34.6 & \textbf{27.6} & 30.2 & 36.0 & 36.0 & 85.0 & 43.6 & 83.0 & 58.6 & 31.6 & 83.3 & \textbf{35.3} & \textbf{49.7} & 3.3 & 28.8 & 35.6 & 48.5 \\
			Ours & D & \textbf{91.2} & 43.3 & \textbf{85.2} & \textbf{38.6} & 25.9 & 34.7 & \textbf{41.3} & 41.0 & \textbf{85.5} & \textbf{46.0} & \textbf{86.5} & 61.7 & \textbf{33.8} & \textbf{85.5} & 34.4 & 48.7 & 0.0 & \textbf{36.1} & 37.8 & \textbf{50.4} \\ \hline
			Curriculum~\cite{cda} & F & 74.9 & 22.0 & 71.7 & 6.0 & 11.9 & 8.4 & 16.3 & 11.1 & 75.7 & 13.3 & 66.5 & 38.0 & 9.3 & 55.2 & 18.8 & 18.9 & 0.0 & 16.8 & 16.6 & 28.9 \\
			CBST~\cite{cbst} & F & 66.7 & 26.8 & 73.7 & 14.8 & 9.5 & \textbf{28.3} & 25.9 & 10.1 & 75.5 & 15.7 & 51.6 & 47.2 & 6.2 & 71.9 & 3.7 & 2.2 & 5.4 & 18.9 & 32.4 & 30.9 \\
			Cycada~\cite{cycada} & F & 85.2 & 37.2 & 76.5 & 21.8 & 15.0 & 23.8 & 22.9 & 21.5 & 80.5 & 31.3 & 60.7 & 50.5 & 9.0 & 76.9 & 17.1 & 28.2 & 4.5 & 9.8 & 0.0 & 35.4 \\
			DCAN~\cite{dcan} & F & 82.3 & 26.7 & 77.4 & 23.7 & 20.5 & 20.4 & \textbf{30.3} & 15.9 & 80.9 & 25.4 & 69.5 & 52.6 & 11.1 & 79.6 & 24.9 & 21.2 & 1.3 & 17.0 & 6.7 & 36.2 \\
			LSD~\cite{lsd} & F & 88.0 & 30.5 & 78.6 & 25.2 & 23.5 & 16.7 & 23.5 & 11.6 & 78.7 & 27.2 & 71.9 & 51.3 & 19.5 & 80.4 & 19.8 & 18.3 & 0.9 & 20.8 & 18.4 & 37.1 \\
			CLAN~\cite{clan} & F & 88.0 & 30.6 & 79.2 & 23.4 & 20.5 & 26.1 & 23.0 & 14.8 & 81.6 & 34.5 & 72.0 & 45.8 & 7.9 & 80.5 & \textbf{26.6} & 29.9 & 0.0 & 10.7 & 0.0 & 36.6 \\
			CrDoCo~\cite{crdoco} & F & 89.1 & 33.2 & 80.1 & 26.9 & 25.0 & 18.3 & 23.4 & 12.8 & 77.0 & 29.1 & 72.4 & \textbf{55.1} & 20.2 & 79.9 & 22.3 & 19.5 & 1.0 & 20.1 & 18.7 & 38.1 \\
			BDL~\cite{bdl} & F & 89.2 & 40.9 & 81.2 & 29.1 & 19.2 & 14.2 & 29.0 & 19.6 & 83.7 & 35.9 & 80.7 & 54.7 & 23.3 & \textbf{82.7} & 25.8 & 28.0 & 2.3 & 25.7 & 19.9 & 41.3 \\
			Ours & F & \textbf{91.1} & \textbf{46.4} & \textbf{82.9} & \textbf{33.2} & \textbf{27.9} & 20.6 & 29.0 & \textbf{28.2} & \textbf{84.5} & \textbf{40.9} & \textbf{82.3} & 52.4 & \textbf{24.4} & 81.2 & 21.8 & \textbf{44.8} & \textbf{31.5} & \textbf{26.5} & \textbf{33.7} & \textbf{46.5} \\
			\hline \\
		\end{tabular}
	}
	\caption{Results of adapting GTA5~\cite{gta} to Cityscapes~\cite{cityscapes}. D stands for DeepLabV2~\cite{deeplab} with ResNet101~\cite{resnet}, while F stands for FCN8s~\cite{fcn} with VGG16~\cite{vgg} as backbone network.}
	\label{tab:gta2cs}
\end{table}

\paragraph{Segmentation loss}
The main supervision for the segmentation task is given by training the network on $(X_{S \rightarrow T}, Y_S)$, where $X_{S \rightarrow T}$ are images translated from the synthetic to the real domain. This is formulated as the common cross-entropy loss:

\begin{equation}
\mathcal{L}_{seg} = -\mathbb{E}_{x \sim X_{S \rightarrow T}, y \sim Y_S} \sum_{k=1}^K{\mathbf{1}_{[k=y]} \log(M(x)_k)}
\end{equation}

\paragraph{Self-supervised segmentation}
Following~\cite{bdl}, we also adopt self-supervision to improve the adaptation model. To this end, we compute $M(X_T)$ and use as labels the high confidence predictions, creating $Y_T^{SSL}$:

\begin{equation}
\label{eq:pseudo-labels}
Y_T^{SSL} =
\begin{cases}
\arg{\underset{1 \leq k \leq K}{\max} \; M(X_T)_k}, & \text{if } M(X_T)_k \ge th_{SSL} \\
-1, & \text{otherwise}
\end{cases}
\end{equation}

where $K$ is the number of classes, $-1$ is the index ignored and $th_{SSL}$ is the confidence threshold, which we use to filter the uncertain predictions. In our experiments we set $th_{SSL}=0.9$.

This makes us able to compute a cross-entropy loss also on the target dataset:

\begin{equation}
\mathcal{L}_{SSL} = -\mathbb{E}_{x \sim X_T, y \sim Y_T^{SSL}} \sum_{k=1}^{K}{\mathbf{1}_{[k=y]} \log(M(x)_k)}
\end{equation}

\paragraph{Adversarial loss}
Supervision on pixel-level aligned images and self-supervision on target images are not enough to learn a full model. This is why we also make use of adversarial training by feeding the semantic maps to a discriminator $D_{seg}$, which has to distinguish the maps predicted by $M$ for $S$ and $T$, giving:

\begin{equation}
\mathcal{L}_{adv} = \mathbb{E}_{x_T \sim {X_T}}[\log (D_{seg}(M(x_T)))] + \mathbb{E}_{x_{S \rightarrow T} \sim {X_{S \rightarrow T}}}[\log (1- D_{seg}(M(x_{S \rightarrow T})))]
\end{equation}

This loss enforces an output space alignment~\cite{adaptsegnet}, which means that $M$ has to learn how to predict semantic maps with distributions that are aligned regardless of the input domain.

\section{Experiments}

\begin{table}
	\resizebox{\linewidth}{!}{
		\begin{tabular}{ccccccccccccccccccc}
			\hline
			\multicolumn{19}{c}{SYNTHIA $\rightarrow$ Cityscapes}\\
			\hline
			\\
			Method & \rotatebox{90}{Arch.} & \rotatebox{90}{road} & \rotatebox{90}{sidewalk} & \rotatebox{90}{building} & \rotatebox{90}{wall} & \rotatebox{90}{fence} & \rotatebox{90}{pole} & \rotatebox{90}{light} & \rotatebox{90}{sign} & \rotatebox{90}{veget} & \rotatebox{90}{sky} & \rotatebox{90}{person} & \rotatebox{90}{rider} & \rotatebox{90}{car} & \rotatebox{90}{bus} & \rotatebox{90}{mbike} & \rotatebox{90}{bike} & mIoU \\ \hline
			AdaptSegNet~\cite{adaptsegnet} & D & 79.2 & 37.2 & 78.8 & - & - & - & 9.9 & 10.5 & 78.2 & 80.5 & 53.5 & 19.6 & 67.0 & 29.5 & 21.6 & 31.3 & 45.9 \\
			CLAN~\cite{clan} & D & 81.3 & 37.0 & 80.1 & - & - & - & 16.1 & 13.7 & 78.2 & 81.5 & 53.4 & 21.2 & 73.0 & 32.9 & 22.6 & 30.7 & 47.8 \\
			BDL~\cite{bdl} & D & 86.0 & 46.7 & 80.3 & - & - & - & 14.1 & 11.6 & 79.2 & 81.3 & 54.1 & 27.9 & \textbf{73.7} & 42.2 & 25.7 & 45.3 & 51.4 \\
			Ours & D & \textbf{87.7} & \textbf{49.7} & \textbf{81.6} & - & - & - & \textbf{19.3} & \textbf{18.5} & \textbf{81.1} & \textbf{83.7} & \textbf{58.7} & \textbf{31.8} & 73.3 & \textbf{47.9} & \textbf{37.1} & \textbf{45.7} & \textbf{55.1} \\ \hline
			FCNsITW~\cite{fcnsitw} & F & 11.5 & 19.6 & 30.8 & 4.4 & 0.0 & 20.3 & 0.1 & 11.7 & 42.3 & 68.7 & 51.2 & 3.8 & 54.0 & 3.2 & 0.2 & 0.6 & 20.2 \\
			Curriculum~\cite{cda} & F & 65.2 & 26.1 & 74.9 & 0.1 & 0.5 & 10.7 & 3.5 & 3.0 & 76.1 & 70.6 & 47.1 & 8.2 & 43.2 & 20.7 & 0.7 & 13.1 & 29.0 \\
			CBST~\cite{cbst} & F & 69.6 & 28.7 & 69.5 & \textbf{12.1} & 0.1 & 25.4 & 11.9 & 13.6 & \textbf{82.0} & \textbf{81.9} & 49.1 & 14.5 & 66.0 & 6.6 & 3.7 & 32.4 & 35.4 \\
			DCAN~\cite{dcan} & F & 79.9 & 30.4 & 70.8 & 1.6 & \textbf{0.6} & 22.3 & 6.7 & 23.0 & 76.9 & 73.9 & 41.9 & 16.7 & 61.7 & 11.5 & 10.3 & 38.6 & 35.4 \\
			CLAN~\cite{clan} & F & 80.4 & 30.7 & 74.7 & - & - & - & 1.4 & 8.0 & 77.1 & 79.0 & 46.5 & 8.9 & 73.8 & 18.2 & 2.2 & 9.9 & 39.3 \\
			CrDoCo~\cite{crdoco} & F & \textbf{84.9} & 32.8 & \textbf{80.1} & 4.3 & 0.4 & \textbf{29.4} & 14.2 & 21.0 & 79.2 & 78.3 & 50.2 & 15.9 & 69.8 & 23.4 & 11.0 & 15.6 & 38.2 \\
			BDL~\cite{bdl} & F & 72.0 & 30.3 & 74.5 & 0.1 & 0.3 & 24.6 & 10.2 & 25.2 & 80.5 & 80.0 & 54.7 & \textbf{23.2} & 72.7 & \textbf{24.0} & 7.5 & 44.9 & 39.0 \\
			Ours & F & 79.1 & \textbf{34.0} & 78.3 & 0.3 & \textbf{0.6} & 26.7 & \textbf{15.9} & \textbf{29.5} & 81.0 & 81.1 & \textbf{55.5} & 21.9 & \textbf{77.2} & 23.5 & \textbf{11.8} & \textbf{47.5} & \textbf{41.5} \\
			\hline \\
		\end{tabular}
	}
	\caption{Results of adapting SYNTHIA~\cite{synthia} to Cityscapes~\cite{synthia}. D stands for DeepLabV2~\cite{deeplab} with ResNet101~\cite{resnet}, while F stands for FCN8s~\cite{fcn} with VGG16~\cite{vgg} as backbone network.}
	\label{tab:synthia2cs}
\end{table}

We present our experimental results for the synthetic to real adaptation using two dataset settings: GTA5~\cite{gta} to Cityscapes~\cite{cityscapes} and SYNTHIA~\cite{synthia} to Cityscapes. We evaluate the mean intersection-over-union (IoU) on the Cityscapes validation set and show how our method outperforms the current state-of-the-art by adopting the same segmentation models. Finally, we conduct an ablation study to highlight the value of our contributions.

\paragraph{Segmentation network}
We choose to adapt two segmentation networks: DeepLabV2~\cite{deeplab} with ResNet101~\cite{resnet} and FCN8s~\cite{fcn} with VGG16~\cite{vgg}.  Both networks are trained on images downsampled to 1024x512 with batch size 1.

We initialize the segmentation networks from~\cite{bdl} to speed up the training process. In order to show the independence from this initialization, we also conduct one experiment where we train DeepLabV2 from scratch for the GTA$\rightarrow$Cityscapes task, and we find this to be in line with the results that we get by initializing it with~\cite{bdl}.

\paragraph{Translation network}
For the translation part, we describe the architecture of the encoders, generators and discriminators.

The encoder is made by few downsampling blocks, followed by residual blocks for further processing of the latent code and they all use IN~\cite{instancenorm}. Symmetrically, the generators take in the latent code and process it with residual blocks, where IN and SPADE are combined to normalize the feature maps. These are followed by upsampling blocks with Layer Normalization~\cite{layernorm}. We found LN to better preserve the style in the generated activations.

In each domain we have discriminators for multiple scales~\cite{pix2pixhd}, each being a Patch Discriminator~\cite{pix2pix, li2016precomputed}. The GAN~\cite{gan} objective we choose is the one proposed in LSGAN~\cite{lsgan}. We apply Spectral Normalization~\cite{spectralnorm} to all the models described here.

When training the translation model we resize the input images to 1024x512 and take 512x512 random crops out of them. We use Adam~\cite{adam} as optimizer with $\beta_1=0.9$ and $\beta_2=0.99$. We apply TTUR~\cite{ttur} and set the initial learning rate to be $10^{-4}$. The learning rate is scheduled to decay to $0$ after 1000000 iterations with a 'poly' scheduling where the power is $0.9$. The batch size is 1 for all the experiments. The loss weights are set to $\lambda_{recon}=10$, $\lambda_{GAN}=1$, $\lambda_{CC_I}=10$, $\lambda_{CC_H}=1$, $\lambda_{SCE}=10$.

\paragraph{Bidirectional learning}
Pixel-level and feature-level alignment are not performed in an end-to-end fashion. Besides being highly expensive in terms of memory requirements, we found this approach to be very unstable and it did not lead to good results.

We adopt a policy similar to~\cite{bdl} and iteratively recreate $X_{S \rightarrow T}$ when $M$ stops improving on the target dataset. Before each training of the segmentation network, we also generate new pseudo-labels $Y_T^{SSL}$. We found this procedure to significantly improve the final mIoU compared to a single iteration of pixel-level and feature-level alignment.

\subsection{Comparison with State of the Art}

\paragraph{GTA5 to Cityscapes}
For the GTA5~\cite{gta} to Cityscapes~\cite{cityscapes} task, we evaluate on all the 19 classes used in the Cityscapes benchmark since the datasets are fully compatible. Some visual results for this setting are presented in \figurename~\ref{fig:seg_comparison}. In this case, the upper bounds in terms of mIoU are 65.1 for DeepLabV2~\cite{deeplab} and 60.3 for FCN8s~\cite{fcn}, which are the results achievable by training with the target labels. In \tablename~\ref{tab:gta2cs} we compare our results with the related work. In terms of mIoU, we get respectively +1.9\% and +4.2\% over the state-of-the-art with the two networks.

\paragraph{SYNTHIA to Cityscapes}
SYNTHIA~\cite{synthia} has been adopted in the past by the other works for its overlapping with 16 of the Cityscapes classes. For the SYNTHIA to Cityscapes task we compare our results with the state-of-the-art in \tablename~\ref{tab:synthia2cs} and present some visual results in the supplementary material. For a fair comparison, the results of the DeepLabV2 architecture are limited to the 13 classes adopted by the other works~\cite{adaptsegnet, clan, bdl}. The upper bounds in terms of mIoU are 71.7 for DeepLabV2 and 59.5 for FCN8s. In the case of DeepLabV2 we surpass the current state-of-the-art in mIoU by +3.7\%. For FCN8s, instead, we get +2.5\% on the mIoU.

\subsection{Ablation study}
\label{sec:ablation}

\begin{table}
	\begin{center}
		\resizebox{.5\linewidth}{!}{
			\begin{tabular}{ccccc}
				\hline
				SPADE & $\mathcal{L}_{SCE}$ & mIoU & Gain & Gap to UB \\ \hline
				& & 49.2 & 15.6 & 15.9 \\
				\checkmark & & 49.5 & 15.9 & 15.6 \\
				& \checkmark & 49.5 & 15.9 & 15.6 \\
				\checkmark & \checkmark & 50.4 & 16.8 & 14.7 \\
				\hline \\
			\end{tabular}
		}
		\caption{\textbf{Ablation study}. We report the mIoU, the gain \textit{wrt} the lower bound (\textit{i.e.} training naively on source), the gap \textit{wrt} the upper bound (\textit{i.e.} training on target).}
		\label{tab:ablation}
	\end{center}
\end{table}

In order to weight our contribution, we perform an ablation study of the proposed method (see \tablename~\ref{tab:ablation}). For each experiment, we report 3 values: the mIoU; the gain \textit{wrt} the lower bound, which is a naive training on the source dataset; the remaining gap \textit{wrt} the upper bound, which is the result for training with target labels (called oracle prediction).

The experiments are conducted with DeepLabV2~\cite{deeplab} for the GTA5~\cite{gta} to Cityscapes~\cite{cityscapes} task, for which the lower bound is 33.6 and the upper bound is 65.1.

We first show the baseline results that we get by using the generator with no Symmetric Cross-Entropy $\mathcal{L}_{SCE}$ and no semantic guidance. In this setting, the residual blocks of the generator use IN~\cite{instancenorm} layers and the image-to-image translation is completely unrelated to the semantic segmentation. Secondly, we add the semantic guidance with the SPADE~\cite{spade} layer. This setting can still benefit from the semantic guidance in the translation, but loses the ability to enforce the cross-domain consistency for the segmentation task. Then we swap back the SPADE layer with IN and enable $\mathcal{L}_{SCE}$. This setting resembles the one used in~\cite{crdoco}, where the architecture of CycleGAN~\cite{cyclegan} is replaced by ours. Finally, we show that the best results are achieved by the combination of the two elements, which completely bridges the translation and segmentation tasks and is the final setting of our work.

We can see that when we remove SPADE or $\mathcal{L}_{SCE}$ the mIoU drops, suggesting that they both have an important contribution to get the best result.

\subsection{Generated image quality}
We also report the quality of the images generated by our image-to-image translation model. In \tablename~\ref{tab:quality} we report the Inception Score (IS)~\cite{inception-score} of the images $X_{S \rightarrow T}$ and the Fr\'echet Inception Distance (FID)~\cite{ttur} with the Cityscapes training set. Although the IS of the produced images is low in every setting, the FID results indicate that the semantic guidance induced by DeepLabV2 is the one that best visually aligns the synthetic domain to Cityscapes. The images translated from SYNTHIA, however, have a much greater distance from Cityscapes than the ones translated from GTA5, regardless of the network used as semantic guidance. We note that this is possibly due to the bigger initial gap in visual appearance between the two domains, since the FID between the original SYNTHIA and Cityscapes is 156.92, while the FID between the original GTA5 and Cityscapes is only 62.42.

\begin{table}
	\begin{center}
		\resizebox{.5\linewidth}{!}{
			\begin{tabular}{ccccc}
				\hline
				Setting & Network & IS & FID \\ \hline
				GTA5 $\rightarrow$ Cityscapes & DeepLabV2 & 4.9 & 27.9 \\
				GTA5 $\rightarrow$ Cityscapes & FCN8s & 4.8 & 40.3 \\
				SYNTHIA $\rightarrow$ Cityscapes & DeepLabV2 & 5.0 & 100.8 \\
				SYNTHIA $\rightarrow$ Cityscapes & FCN8s & 4.9 & 113.7 \\
				\hline \\
			\end{tabular}
		}
	\caption{\textbf{Image quality evaluation}. We report the Inception Score (IS)~\cite{inception-score} and the Fr\'echet Inception Distance (FID)~\cite{ttur} of the images generated in each setting of our experiments.}
	\label{tab:quality}
	\end{center}
\end{table}

\bibliography{egbib}

\begin{thebibliography}{52}
\providecommand{\natexlab}[1]{#1}
\providecommand{\url}[1]{\texttt{#1}}
\expandafter\ifx\csname urlstyle\endcsname\relax
  \providecommand{\doi}[1]{doi: #1}\else
  \providecommand{\doi}{doi: \begingroup \urlstyle{rm}\Url}\fi

\bibitem[Ba et~al.(2016)Ba, Kiros, and Hinton]{layernorm}
Jimmy~Lei Ba, Jamie~Ryan Kiros, and Geoffrey~E Hinton.
\newblock Layer normalization.
\newblock \emph{arXiv preprint arXiv:1607.06450}, 2016.

\bibitem[Chen et~al.(2017{\natexlab{a}})Chen, Papandreou, Kokkinos, Murphy, and
  Yuille]{deeplab}
Liang-Chieh Chen, George Papandreou, Iasonas Kokkinos, Kevin Murphy, and Alan~L
  Yuille.
\newblock Deeplab: Semantic image segmentation with deep convolutional nets,
  atrous convolution, and fully connected crfs.
\newblock \emph{IEEE transactions on pattern analysis and machine
  intelligence}, 40\penalty0 (4):\penalty0 834--848, 2017{\natexlab{a}}.

\bibitem[Chen et~al.(2018)Chen, Liu, Wang, Wassell, and Chetty]{raan}
Qingchao Chen, Yang Liu, Zhaowen Wang, Ian Wassell, and Kevin Chetty.
\newblock Re-weighted adversarial adaptation network for unsupervised domain
  adaptation.
\newblock In \emph{Proceedings of the IEEE Conference on Computer Vision and
  Pattern Recognition}, pages 7976--7985, 2018.

\bibitem[Chen et~al.(2017{\natexlab{b}})Chen, Chen, Chen, Tsai, Frank~Wang, and
  Sun]{nomoredis}
Yi-Hsin Chen, Wei-Yu Chen, Yu-Ting Chen, Bo-Cheng Tsai, Yu-Chiang Frank~Wang,
  and Min Sun.
\newblock No more discrimination: Cross city adaptation of road scene
  segmenters.
\newblock In \emph{Proceedings of the IEEE International Conference on Computer
  Vision}, pages 1992--2001, 2017{\natexlab{b}}.

\bibitem[Chen et~al.(2019)Chen, Lin, Yang, and Huang]{crdoco}
Yun-Chun Chen, Yen-Yu Lin, Ming-Hsuan Yang, and Jia-Bin Huang.
\newblock Crdoco: Pixel-level domain transfer with cross-domain consistency.
\newblock In \emph{Proceedings of the IEEE Conference on Computer Vision and
  Pattern Recognition}, pages 1791--1800, 2019.

\bibitem[Cordts et~al.(2016)Cordts, Omran, Ramos, Rehfeld, Enzweiler, Benenson,
  Franke, Roth, and Schiele]{cityscapes}
Marius Cordts, Mohamed Omran, Sebastian Ramos, Timo Rehfeld, Markus Enzweiler,
  Rodrigo Benenson, Uwe Franke, Stefan Roth, and Bernt Schiele.
\newblock The cityscapes dataset for semantic urban scene understanding.
\newblock In \emph{Proceedings of the IEEE conference on computer vision and
  pattern recognition}, pages 3213--3223, 2016.

\bibitem[Diederik et~al.(2014)Diederik, Welling, et~al.]{vae_kingma}
P~Kingma Diederik, Max Welling, et~al.
\newblock Auto-encoding variational bayes.
\newblock In \emph{Proceedings of the International Conference on Learning
  Representations (ICLR)}, 2014.

\bibitem[Dundar et~al.(2018)Dundar, Liu, Wang, Zedlewski, and
  Kautz]{domstylization}
Aysegul Dundar, Ming-Yu Liu, Ting-Chun Wang, John Zedlewski, and Jan Kautz.
\newblock Domain stylization: A strong, simple baseline for synthetic to real
  image domain adaptation.
\newblock \emph{arXiv preprint arXiv:1807.09384}, 2018.

\bibitem[Ganin and Lempitsky(2015)]{ganin2015unsupervised}
Yaroslav Ganin and Victor Lempitsky.
\newblock Unsupervised domain adaptation by backpropagation.
\newblock In \emph{Proceedings of the 32nd International Conference on
  International Conference on Machine Learning-Volume 37}, pages 1180--1189.
  JMLR. org, 2015.

\bibitem[Ganin et~al.(2016)Ganin, Ustinova, Ajakan, Germain, Larochelle,
  Laviolette, Marchand, and Lempitsky]{ganin2016domain}
Yaroslav Ganin, Evgeniya Ustinova, Hana Ajakan, Pascal Germain, Hugo
  Larochelle, Fran{\c{c}}ois Laviolette, Mario Marchand, and Victor Lempitsky.
\newblock Domain-adversarial training of neural networks.
\newblock \emph{The Journal of Machine Learning Research}, 17\penalty0
  (1):\penalty0 2096--2030, 2016.

\bibitem[Geng et~al.(2011)Geng, Tao, and Xu]{daml}
Bo~Geng, Dacheng Tao, and Chao Xu.
\newblock Daml: Domain adaptation metric learning.
\newblock \emph{IEEE Transactions on Image Processing}, 20\penalty0
  (10):\penalty0 2980--2989, 2011.

\bibitem[Goodfellow et~al.(2014)Goodfellow, Pouget-Abadie, Mirza, Xu,
  Warde-Farley, Ozair, Courville, and Bengio]{gan}
Ian Goodfellow, Jean Pouget-Abadie, Mehdi Mirza, Bing Xu, David Warde-Farley,
  Sherjil Ozair, Aaron Courville, and Yoshua Bengio.
\newblock Generative adversarial nets.
\newblock In \emph{Advances in neural information processing systems}, pages
  2672--2680, 2014.

\bibitem[He et~al.(2016)He, Zhang, Ren, and Sun]{resnet}
Kaiming He, Xiangyu Zhang, Shaoqing Ren, and Jian Sun.
\newblock Deep residual learning for image recognition.
\newblock In \emph{Proceedings of the IEEE conference on computer vision and
  pattern recognition}, pages 770--778, 2016.

\bibitem[Heusel et~al.(2017)Heusel, Ramsauer, Unterthiner, Nessler, and
  Hochreiter]{ttur}
Martin Heusel, Hubert Ramsauer, Thomas Unterthiner, Bernhard Nessler, and Sepp
  Hochreiter.
\newblock Gans trained by a two time-scale update rule converge to a local nash
  equilibrium.
\newblock In \emph{Advances in Neural Information Processing Systems}, pages
  6626--6637, 2017.

\bibitem[Hoffman et~al.(2016)Hoffman, Wang, Yu, and Darrell]{fcnsitw}
Judy Hoffman, Dequan Wang, Fisher Yu, and Trevor Darrell.
\newblock Fcns in the wild: Pixel-level adversarial and constraint-based
  adaptation.
\newblock \emph{arXiv preprint arXiv:1612.02649}, 2016.

\bibitem[Hoffman et~al.(2018)Hoffman, Tzeng, Park, Zhu, Isola, Saenko, Efros,
  and Darrell]{cycada}
Judy Hoffman, Eric Tzeng, Taesung Park, Jun-Yan Zhu, Phillip Isola, Kate
  Saenko, Alexei Efros, and Trevor Darrell.
\newblock Cycada: Cycle-consistent adversarial domain adaptation.
\newblock In \emph{Proceedings of the 35th International Conference on Machine
  Learning}, 2018.

\bibitem[Huang and Belongie(2017)]{adain}
Xun Huang and Serge Belongie.
\newblock Arbitrary style transfer in real-time with adaptive instance
  normalization.
\newblock In \emph{Proceedings of the IEEE International Conference on Computer
  Vision}, pages 1501--1510, 2017.

\bibitem[Huang et~al.(2018)Huang, Liu, Belongie, and Kautz]{munit}
Xun Huang, Ming-Yu Liu, Serge Belongie, and Jan Kautz.
\newblock Multimodal unsupervised image-to-image translation.
\newblock In \emph{Proceedings of the European Conference on Computer Vision
  (ECCV)}, pages 172--189, 2018.

\bibitem[Ioffe and Szegedy(2015)]{batchnorm}
Sergey Ioffe and Christian Szegedy.
\newblock Batch normalization: Accelerating deep network training by reducing
  internal covariate shift.
\newblock In \emph{International Conference on Machine Learning}, pages
  448--456, 2015.

\bibitem[Isola et~al.(2017)Isola, Zhu, Zhou, and Efros]{pix2pix}
Phillip Isola, Jun-Yan Zhu, Tinghui Zhou, and Alexei~A Efros.
\newblock Image-to-image translation with conditional adversarial networks.
\newblock In \emph{Proceedings of the IEEE conference on computer vision and
  pattern recognition}, pages 1125--1134, 2017.

\bibitem[Karras et~al.(2019)Karras, Laine, and Aila]{stylegan}
Tero Karras, Samuli Laine, and Timo Aila.
\newblock A style-based generator architecture for generative adversarial
  networks.
\newblock In \emph{Proceedings of the IEEE Conference on Computer Vision and
  Pattern Recognition}, pages 4401--4410, 2019.

\bibitem[Kingma and Ba(2014)]{adam}
Diederik~P Kingma and Jimmy Ba.
\newblock Adam: A method for stochastic optimization.
\newblock \emph{arXiv preprint arXiv:1412.6980}, 2014.

\bibitem[Li and Wand(2016)]{li2016precomputed}
Chuan Li and Michael Wand.
\newblock Precomputed real-time texture synthesis with markovian generative
  adversarial networks.
\newblock In \emph{European Conference on Computer Vision}, pages 702--716.
  Springer, 2016.

\bibitem[Li et~al.(2019)Li, Yuan, and Vasconcelos]{bdl}
Yunsheng Li, Lu~Yuan, and Nuno Vasconcelos.
\newblock Bidirectional learning for domain adaptation of semantic
  segmentation.
\newblock In \emph{Proceedings of the IEEE Conference on Computer Vision and
  Pattern Recognition}, pages 6936--6945, 2019.

\bibitem[Liu and Tuzel(2016)]{cogan}
Ming-Yu Liu and Oncel Tuzel.
\newblock Coupled generative adversarial networks.
\newblock In \emph{Advances in neural information processing systems}, pages
  469--477, 2016.

\bibitem[Liu et~al.(2017)Liu, Breuel, and Kautz]{unit}
Ming-Yu Liu, Thomas Breuel, and Jan Kautz.
\newblock Unsupervised image-to-image translation networks.
\newblock In \emph{Advances in neural information processing systems}, pages
  700--708, 2017.

\bibitem[Liu et~al.(2019)Liu, Huang, Mallya, Karras, Aila, Lehtinen, and
  Kautz]{funit}
Ming-Yu Liu, Xun Huang, Arun Mallya, Tero Karras, Timo Aila, Jaakko Lehtinen,
  and Jan Kautz.
\newblock Few-shot unsupervised image-to-image translation.
\newblock In \emph{IEEE International Conference on Computer Vision (ICCV)},
  2019.

\bibitem[Long et~al.(2015{\natexlab{a}})Long, Shelhamer, and Darrell]{fcn}
Jonathan Long, Evan Shelhamer, and Trevor Darrell.
\newblock Fully convolutional networks for semantic segmentation.
\newblock In \emph{Proceedings of the IEEE conference on computer vision and
  pattern recognition}, pages 3431--3440, 2015{\natexlab{a}}.

\bibitem[Long et~al.(2015{\natexlab{b}})Long, Cao, Wang, and
  Jordan]{learning_transferable}
Mingsheng Long, Yue Cao, Jianmin Wang, and Michael~I Jordan.
\newblock Learning transferable features with deep adaptation networks.
\newblock In \emph{Proceedings of the 32nd International Conference on
  International Conference on Machine Learning-Volume 37}, pages 97--105. JMLR.
  org, 2015{\natexlab{b}}.

\bibitem[Long et~al.(2016)Long, Zhu, Wang, and Jordan]{long2016unsupervised}
Mingsheng Long, Han Zhu, Jianmin Wang, and Michael~I Jordan.
\newblock Unsupervised domain adaptation with residual transfer networks.
\newblock In \emph{Advances in Neural Information Processing Systems}, pages
  136--144, 2016.

\bibitem[Luo et~al.(2019)Luo, Zheng, Guan, Yu, and Yang]{clan}
Yawei Luo, Liang Zheng, Tao Guan, Junqing Yu, and Yi~Yang.
\newblock Taking a closer look at domain shift: Category-level adversaries for
  semantics consistent domain adaptation.
\newblock In \emph{Proceedings of the IEEE Conference on Computer Vision and
  Pattern Recognition}, pages 2507--2516, 2019.

\bibitem[Mao et~al.(2017)Mao, Li, Xie, Lau, Wang, and Paul~Smolley]{lsgan}
Xudong Mao, Qing Li, Haoran Xie, Raymond~YK Lau, Zhen Wang, and Stephen
  Paul~Smolley.
\newblock Least squares generative adversarial networks.
\newblock In \emph{Proceedings of the IEEE International Conference on Computer
  Vision}, pages 2794--2802, 2017.

\bibitem[Miyato et~al.(2018)Miyato, Kataoka, Koyama, and Yoshida]{spectralnorm}
Takeru Miyato, Toshiki Kataoka, Masanori Koyama, and Yuichi Yoshida.
\newblock Spectral normalization for generative adversarial networks.
\newblock In \emph{International Conference on Learning Representations}, 2018.
\newblock URL \url{https://openreview.net/forum?id=B1QRgziT-}.

\bibitem[Park et~al.(2019)Park, Liu, Wang, and Zhu]{spade}
Taesung Park, Ming-Yu Liu, Ting-Chun Wang, and Jun-Yan Zhu.
\newblock Semantic image synthesis with spatially-adaptive normalization.
\newblock In \emph{Proceedings of the IEEE Conference on Computer Vision and
  Pattern Recognition}, pages 2337--2346, 2019.

\bibitem[Rezende et~al.(2014)Rezende, Mohamed, and
  Wierstra]{rezende2014stochastic}
Danilo~Jimenez Rezende, Shakir Mohamed, and Daan Wierstra.
\newblock Stochastic backpropagation and approximate inference in deep
  generative models.
\newblock In \emph{International Conference on Machine Learning}, pages
  1278--1286, 2014.

\bibitem[Richter et~al.(2016)Richter, Vineet, Roth, and Koltun]{gta}
Stephan~R Richter, Vibhav Vineet, Stefan Roth, and Vladlen Koltun.
\newblock Playing for data: Ground truth from computer games.
\newblock In \emph{European conference on computer vision}, pages 102--118.
  Springer, 2016.

\bibitem[Ros et~al.(2016)Ros, Sellart, Materzynska, Vazquez, and
  Lopez]{synthia}
German Ros, Laura Sellart, Joanna Materzynska, David Vazquez, and Antonio~M
  Lopez.
\newblock The synthia dataset: A large collection of synthetic images for
  semantic segmentation of urban scenes.
\newblock In \emph{Proceedings of the IEEE conference on computer vision and
  pattern recognition}, pages 3234--3243, 2016.

\bibitem[Salimans et~al.(2016)Salimans, Goodfellow, Zaremba, Cheung, Radford,
  and Chen]{inception-score}
Tim Salimans, Ian Goodfellow, Wojciech Zaremba, Vicki Cheung, Alec Radford, and
  Xi~Chen.
\newblock Improved techniques for training gans.
\newblock In \emph{Advances in neural information processing systems}, pages
  2234--2242, 2016.

\bibitem[Sankaranarayanan et~al.(2018)Sankaranarayanan, Balaji, Jain, Nam~Lim,
  and Chellappa]{lsd}
Swami Sankaranarayanan, Yogesh Balaji, Arpit Jain, Ser Nam~Lim, and Rama
  Chellappa.
\newblock Learning from synthetic data: Addressing domain shift for semantic
  segmentation.
\newblock In \emph{Proceedings of the IEEE Conference on Computer Vision and
  Pattern Recognition}, pages 3752--3761, 2018.

\bibitem[Simonyan and Zisserman(2014)]{vgg}
Karen Simonyan and Andrew Zisserman.
\newblock Very deep convolutional networks for large-scale image recognition.
\newblock \emph{arXiv preprint arXiv:1409.1556}, 2014.

\bibitem[Sun and Saenko(2016)]{coral}
Baochen Sun and Kate Saenko.
\newblock Deep coral: Correlation alignment for deep domain adaptation.
\newblock In \emph{European Conference on Computer Vision}, pages 443--450.
  Springer, 2016.

\bibitem[Tsai et~al.(2018)Tsai, Hung, Schulter, Sohn, Yang, and
  Chandraker]{adaptsegnet}
Yi-Hsuan Tsai, Wei-Chih Hung, Samuel Schulter, Kihyuk Sohn, Ming-Hsuan Yang,
  and Manmohan Chandraker.
\newblock Learning to adapt structured output space for semantic segmentation.
\newblock In \emph{Proceedings of the IEEE Conference on Computer Vision and
  Pattern Recognition}, pages 7472--7481, 2018.

\bibitem[Tzeng et~al.(2015)Tzeng, Hoffman, Darrell, and
  Saenko]{tzeng2015simultaneous}
Eric Tzeng, Judy Hoffman, Trevor Darrell, and Kate Saenko.
\newblock Simultaneous deep transfer across domains and tasks.
\newblock In \emph{Proceedings of the IEEE International Conference on Computer
  Vision}, pages 4068--4076, 2015.

\bibitem[Tzeng et~al.(2017)Tzeng, Hoffman, Saenko, and Darrell]{adda}
Eric Tzeng, Judy Hoffman, Kate Saenko, and Trevor Darrell.
\newblock Adversarial discriminative domain adaptation.
\newblock In \emph{Proceedings of the IEEE Conference on Computer Vision and
  Pattern Recognition}, pages 7167--7176, 2017.

\bibitem[Ulyanov et~al.(2016{\natexlab{a}})Ulyanov, Lebedev, Vedaldi, and
  Lempitsky]{texture}
Dmitry Ulyanov, Vadim Lebedev, Andrea Vedaldi, and Victor~S Lempitsky.
\newblock Texture networks: Feed-forward synthesis of textures and stylized
  images.
\newblock In \emph{ICML}, volume~1, page~4, 2016{\natexlab{a}}.

\bibitem[Ulyanov et~al.(2016{\natexlab{b}})Ulyanov, Vedaldi, and
  Lempitsky]{instancenorm}
Dmitry Ulyanov, Andrea Vedaldi, and Victor Lempitsky.
\newblock Instance normalization: The missing ingredient for fast stylization.
\newblock \emph{arXiv preprint arXiv:1607.08022}, 2016{\natexlab{b}}.

\bibitem[Ulyanov et~al.(2017)Ulyanov, Vedaldi, and Lempitsky]{improved-texture}
Dmitry Ulyanov, Andrea Vedaldi, and Victor Lempitsky.
\newblock Improved texture networks: Maximizing quality and diversity in
  feed-forward stylization and texture synthesis.
\newblock In \emph{Proceedings of the IEEE Conference on Computer Vision and
  Pattern Recognition}, pages 6924--6932, 2017.

\bibitem[Wang et~al.(2018)Wang, Liu, Zhu, Tao, Kautz, and Catanzaro]{pix2pixhd}
Ting-Chun Wang, Ming-Yu Liu, Jun-Yan Zhu, Andrew Tao, Jan Kautz, and Bryan
  Catanzaro.
\newblock High-resolution image synthesis and semantic manipulation with
  conditional gans.
\newblock In \emph{The IEEE Conference on Computer Vision and Pattern
  Recognition (CVPR)}, June 2018.

\bibitem[Wu et~al.(2018)Wu, Han, Lin, Gokhan~Uzunbas, Goldstein, Nam~Lim, and
  Davis]{dcan}
Zuxuan Wu, Xintong Han, Yen-Liang Lin, Mustafa Gokhan~Uzunbas, Tom Goldstein,
  Ser Nam~Lim, and Larry~S Davis.
\newblock Dcan: Dual channel-wise alignment networks for unsupervised scene
  adaptation.
\newblock In \emph{Proceedings of the European Conference on Computer Vision
  (ECCV)}, pages 518--534, 2018.

\bibitem[Zhang et~al.(2017)Zhang, David, and Gong]{cda}
Yang Zhang, Philip David, and Boqing Gong.
\newblock Curriculum domain adaptation for semantic segmentation of urban
  scenes.
\newblock In \emph{Proceedings of the IEEE International Conference on Computer
  Vision}, pages 2020--2030, 2017.

\bibitem[Zhu et~al.(2017)Zhu, Park, Isola, and Efros]{cyclegan}
Jun-Yan Zhu, Taesung Park, Phillip Isola, and Alexei~A Efros.
\newblock Unpaired image-to-image translation using cycle-consistent
  adversarial networks.
\newblock In \emph{Proceedings of the IEEE international conference on computer
  vision}, pages 2223--2232, 2017.

\bibitem[Zou et~al.(2018)Zou, Yu, Vijaya~Kumar, and Wang]{cbst}
Yang Zou, Zhiding Yu, BVK Vijaya~Kumar, and Jinsong Wang.
\newblock Unsupervised domain adaptation for semantic segmentation via
  class-balanced self-training.
\newblock In \emph{Proceedings of the European Conference on Computer Vision
  (ECCV)}, pages 289--305, 2018.

\end{thebibliography}

\clearpage

\appendix

\section{Representation of the semantic input}
The image synthesis network of SPADE takes as input a \textit{one-hot} encoding of the ground truth semantic segmentation. Here, instead, we use the unnormalized output of $M$ for every translation that we perform. This is a consequence of the cycle consistency constraints.

As explained in the main article, we have to perform both the $S \rightleftarrows T$ and $T \rightleftarrows S$ cycles, which is why we have to train both $G_S$ and $G_T$ by feeding them semantic maps aligned with the input images. In UDA problems, we do not have access to $Y_T$, which is why we use $M(X_T)$ for the $T \rightleftarrows S$ cycle.

However, we note that the refined output classes predicted by $M$ are far from the ground truth and cannot give an accurate conditioning, especially in the target domain when the segmentation is still in the initial training phases. Because of this, we choose to use as semantic guidance the unnormalized output of $M$. This representation has the advantage of carrying the confidence of the prediction, which could potentially be used by the SPADE layers to avoid denormalizing a region with the incorrect class (\eg on the borders of objects, where the segmentation tends to fail more easily).

In the $S \rightleftarrows T$ cycle, we could use $Y_S$ as semantic guidance, but this would lead to inconsistent input distributions for the SPADE layers, which is why we adopt $M(X_S)$ as semantic guidance in this case too.

\newpage

\section{Detailed architecture}

\begin{table}[htb!]
	\centering
	\label{tab:architecture}
	\smallskip
	\resizebox{0.9\linewidth}{!}{
		\begin{tabular}{ccccccccc}
			\hline
			\multicolumn{9}{c}{Encoder}\\
			\hline
			\\
			\makecell{Kernel\\size} & Stride & \makecell{Input\\channels} & \makecell{Output\\channels} & \makecell{Output\\upsampling} & Residual & \makecell{Activation\\function} & Normalization & \makecell{Spectral\\normalization} \\ \hline
			7 & 1 & 3 & 64 & - & - & ReLU & IN & \checkmark \\
			4 & 2 & 64 & 128 & - & - & ReLU & IN & \checkmark  \\
			4 & 2 & 128 & 256 & - & - & ReLU & IN & \checkmark  \\
			3 & 1 & 256 & 256 & - & \checkmark & ReLU & IN & \checkmark  \\
			3 & 1 & 256 & 256 & - & \checkmark & ReLU & IN & \checkmark  \\
			3 & 1 & 256 & 256 & - & \checkmark & ReLU & IN & \checkmark  \\
			3 & 1 & 256 & 256 & - & \checkmark & ReLU & IN & \checkmark  \\
			\hline
			\multicolumn{9}{c}{Generator}\\
			\hline
			\\
			\makecell{Kernel\\size} & Stride & \makecell{Input\\channels} & \makecell{Output\\channels} & \makecell{Output\\upsampling} & Residual & \makecell{Activation\\function} & Normalization & \makecell{Spectral\\normalization} \\ \hline
			3 & 1 & 256 & 256 & - & \checkmark & ReLU & IN+SPADE & \checkmark  \\
			3 & 1 & 256 & 256 & - & \checkmark & ReLU & IN+SPADE & \checkmark  \\
			3 & 1 & 256 & 256 & - & \checkmark & ReLU & IN+SPADE & \checkmark  \\
			3 & 1 & 256 & 256 & \checkmark & \checkmark & ReLU & IN+SPADE & \checkmark  \\
			5 & 1 & 256 & 128 & \checkmark & - & ReLU & LN & \checkmark \\
			5 & 1 & 128 & 64 & - & - & ReLU & LN & \checkmark  \\
			7 & 1 & 64 & 3 & - & - & Tanh & - & \checkmark  \\
			\hline
			\multicolumn{9}{c}{Discriminator (x3)}\\
			\hline
			\\
			\makecell{Kernel\\size} & Stride & \makecell{Input\\channels} & \makecell{Output\\channels} & \makecell{Output\\upsampling} & Residual & \makecell{Activation\\function} & Normalization & \makecell{Spectral\\normalization} \\ \hline
			4 & 2 & 3 & 64 & - & - & LReLU$_{0.2}$ & - & \checkmark  \\
			4 & 2 & 64 & 128 & - & - & LReLU$_{0.2}$ & - & \checkmark  \\
			4 & 2 & 128 & 256 & - & - & LReLU$_{0.2}$ & - & \checkmark  \\
			4 & 2 & 256 & 512 & - & - & LReLU$_{0.2}$ & - & \checkmark  \\
			1 & 1 & 512 & 1 & - & - & - & - & \checkmark  \\
			\hline
		\end{tabular}
	}
	\medskip
	\caption{\textbf{Detailed architecture of encoders, generators and discriminators in the image-to-image translation step.} The architectures follow the schemes adopted by CycleGAN and UNIT. \textit{Output upsampling} indicates that we use a $2\times$ nearest-neighbor upsampling of the output feature maps. \textit{Residual} indicates that the layer is actually a residual block, not a simple convolutional one. LReLU$_{0.2}$ indicates the Leaky Rectified Linear Unit with slope $\alpha=0.2$.}
\end{table}

\newpage

\section{Fake segmentation}
\begin{figure}[!htb]
	\centering
	\begin{tabular}{@{\hskip3pt}c@{\hskip3pt}c}
		\includegraphics[width=.45\linewidth]{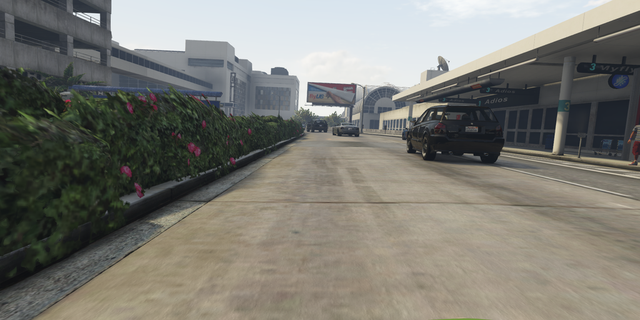}&
		\includegraphics[width=.45\linewidth]{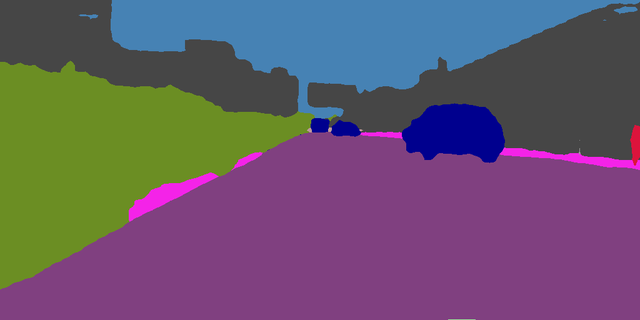}\\
		Source for segmentation&Semantic segmentation\\
		\includegraphics[width=.45\linewidth]{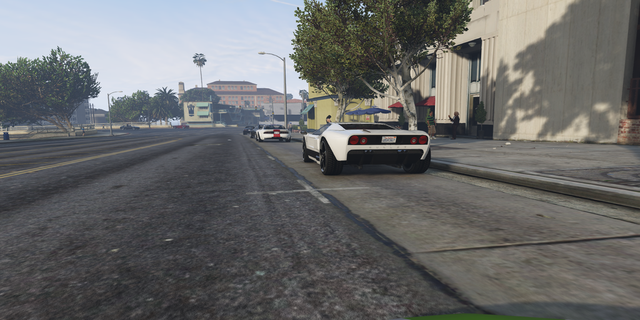}&
		\includegraphics[width=.45\linewidth]{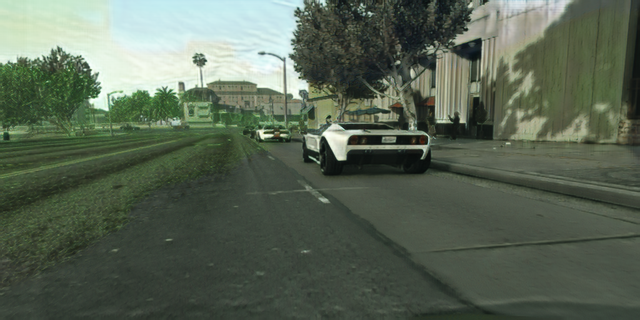}\\
		Source for translation&Translated image\\
	\end{tabular}
	\caption{\textbf{Fake segmentation for image-to-image translation.} We take two different samples $X_S^1$ (a) and $X_S^2$ (c) from GTA5. We then use $M$ to get the predicted segmentation $M(X_S^1)$ (b) and use it as semantic guidance for the translation of $X_S^1$ to get $X_{S \rightarrow T}=F_{S \rightarrow T}(X_S^1, M(X_S^2))$. The result (d) emphasizes the effect of the semantic guidance in our image-to-image translation method.}
	\label{fig:fake-seg}
\end{figure}

The effect of using the SPADE layers in the image-to-image translation model can be seen better when there is a mismatch between the source image and the semantic guidance. To show this effect, we feed the SPADE layers with a segmentation map extracted from an image that is different from the one being translated. In \figurename~\ref{fig:fake-seg}, we can see how the denormalization wrongly creates some features in the region of the image they do not belong to (\textit{i.e.} green on the road).

\newpage

\section{Additional results}

\begin{figure}[!htb]
	\centering
	\begin{tabular}{@{\hskip2pt}c@{\hskip2pt}c@{\hskip2pt}c@{\hskip2pt}c@{\hskip2pt}c}
		\includegraphics[width=.19\linewidth]{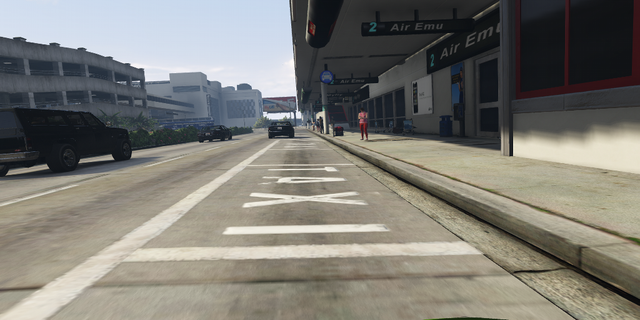}&
		\includegraphics[width=.19\linewidth]{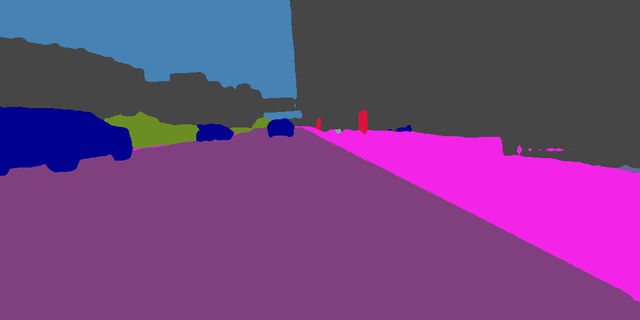}&
		\includegraphics[width=.19\linewidth]{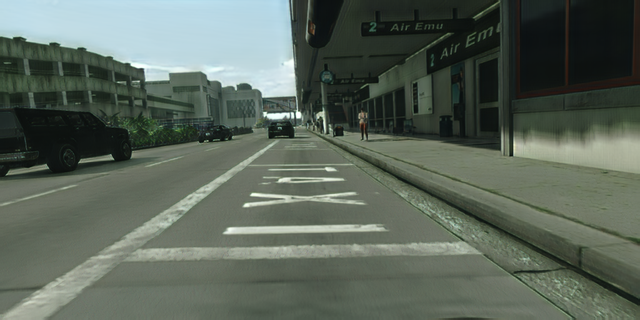}&
		\includegraphics[width=.19\linewidth]{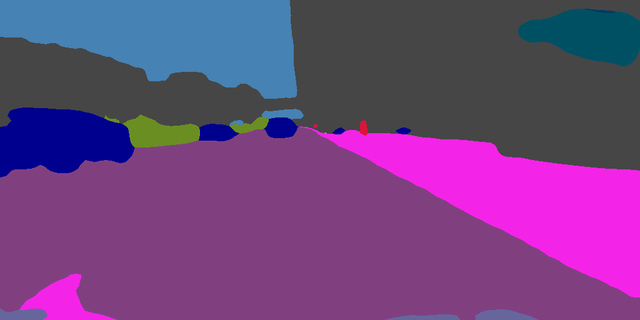}&
		\includegraphics[width=.19\linewidth]{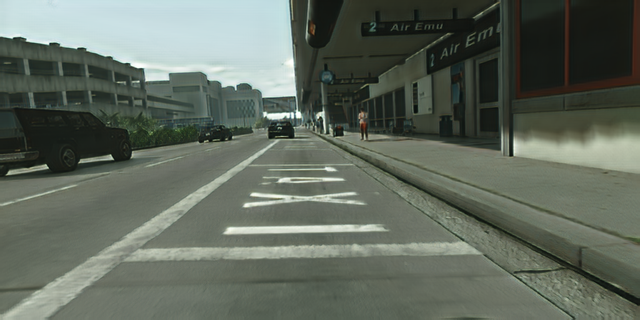}\\
		\includegraphics[width=.19\linewidth]{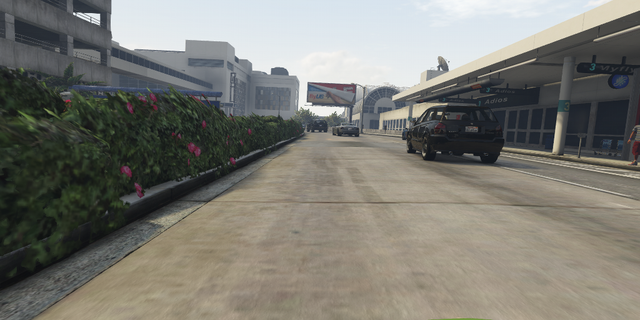}&
		\includegraphics[width=.19\linewidth]{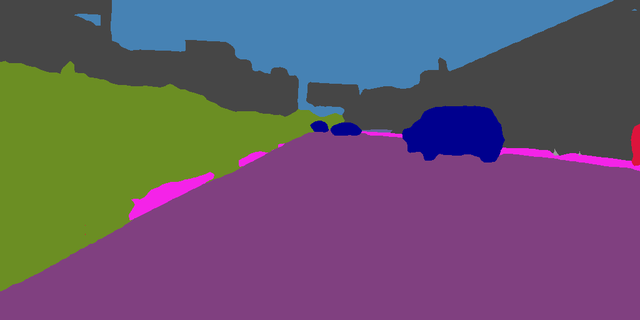}&
		\includegraphics[width=.19\linewidth]{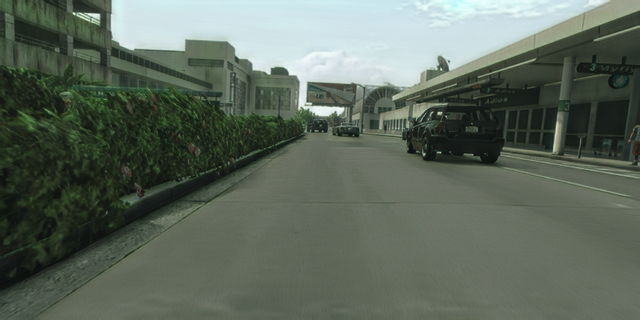}&
		\includegraphics[width=.19\linewidth]{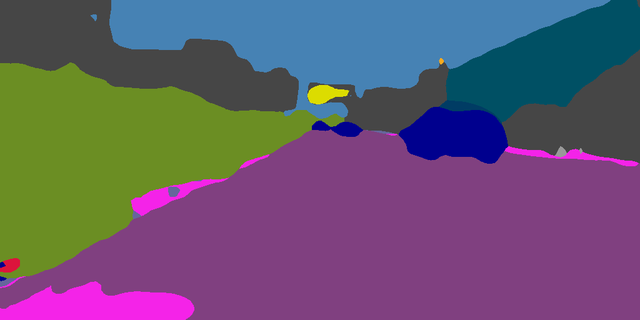}&
		\includegraphics[width=.19\linewidth]{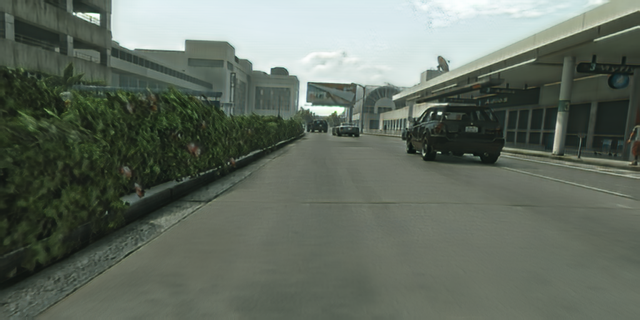}\\
		\includegraphics[width=.19\linewidth]{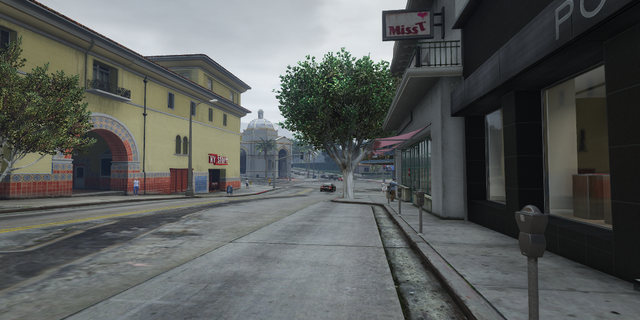}&
		\includegraphics[width=.19\linewidth]{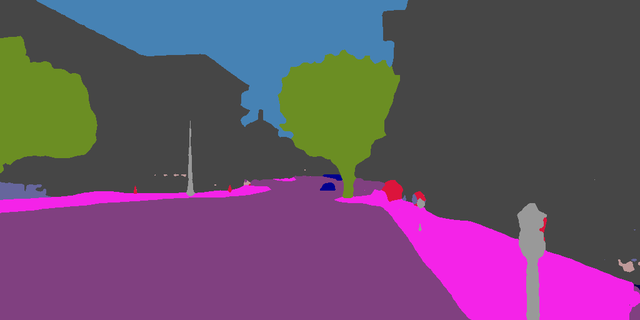}&
		\includegraphics[width=.19\linewidth]{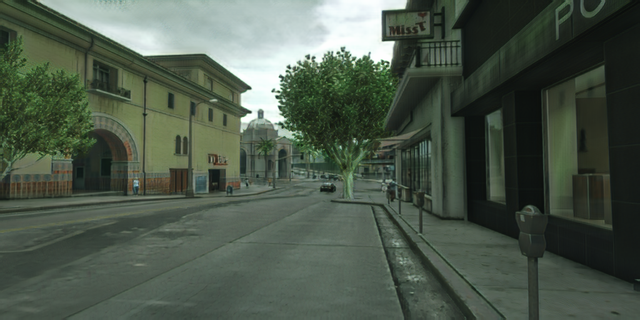}&
		\includegraphics[width=.19\linewidth]{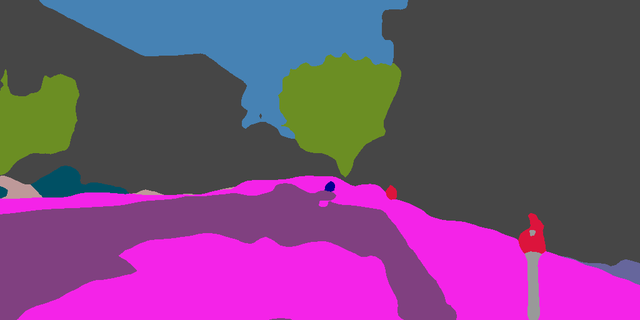}&
		\includegraphics[width=.19\linewidth]{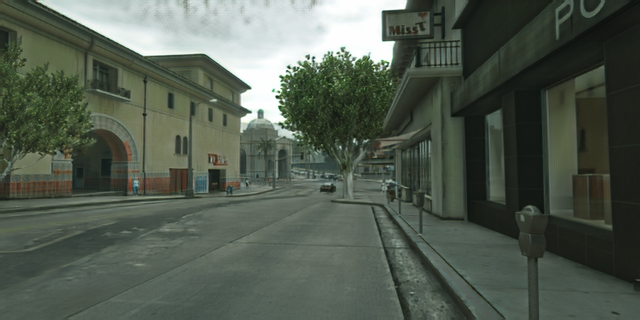}\\
		\includegraphics[width=.19\linewidth]{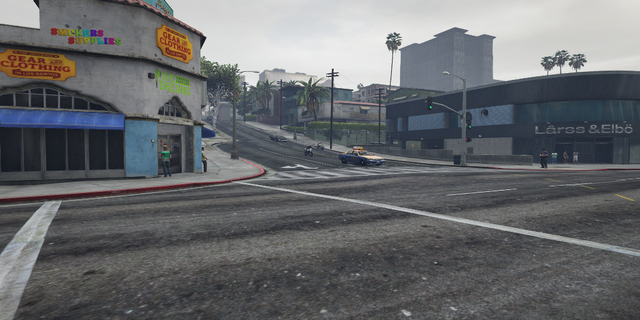}&
		\includegraphics[width=.19\linewidth]{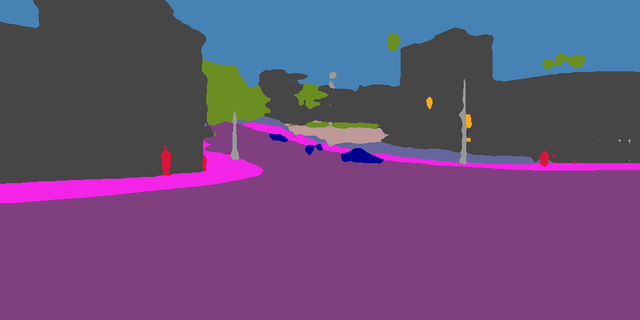}&
		\includegraphics[width=.19\linewidth]{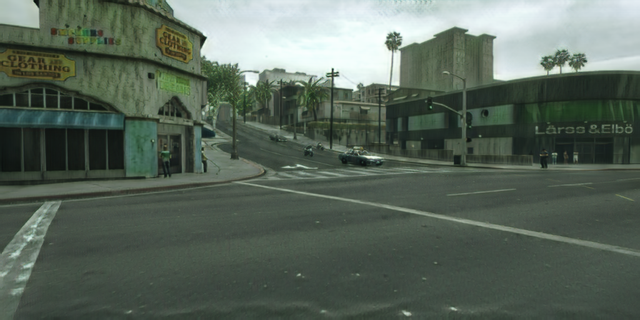}&
		\includegraphics[width=.19\linewidth]{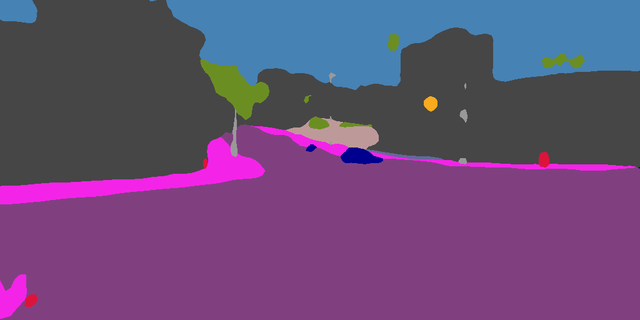}&
		\includegraphics[width=.19\linewidth]{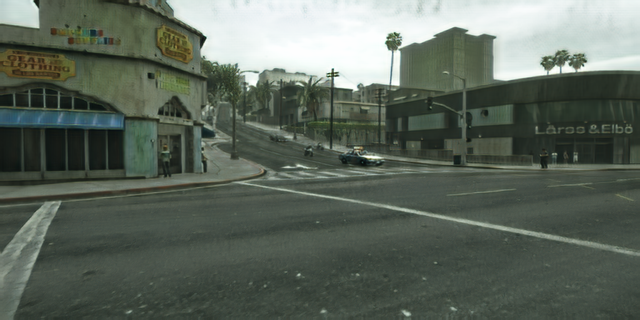}\\
		\includegraphics[width=.19\linewidth]{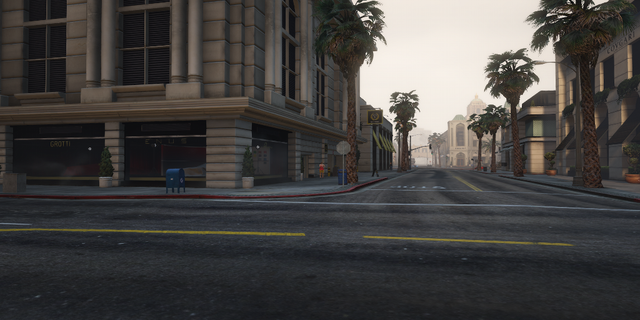}&
		\includegraphics[width=.19\linewidth]{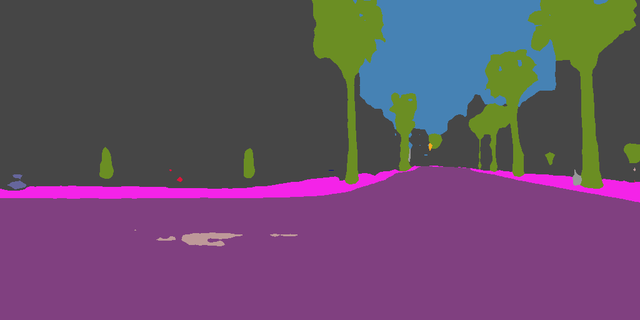}&
		\includegraphics[width=.19\linewidth]{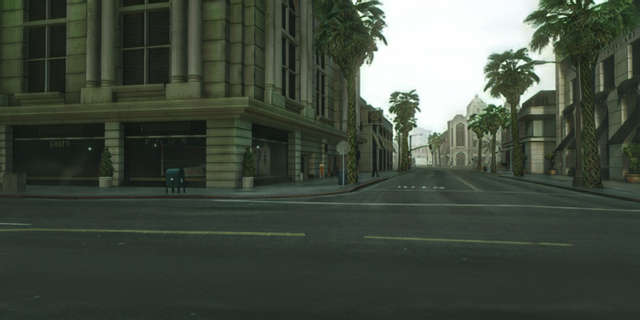}&
		\includegraphics[width=.19\linewidth]{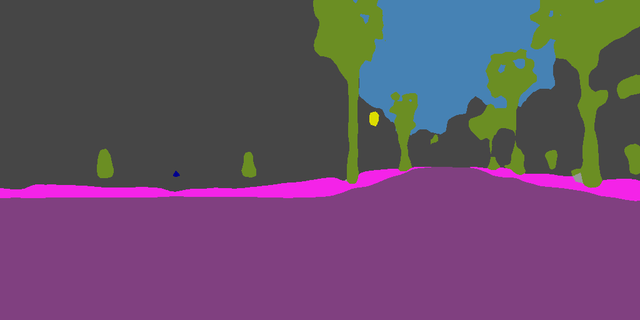}&
		\includegraphics[width=.19\linewidth]{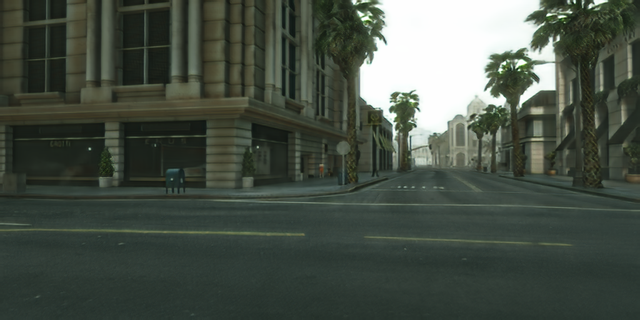}\\
		\includegraphics[width=.19\linewidth]{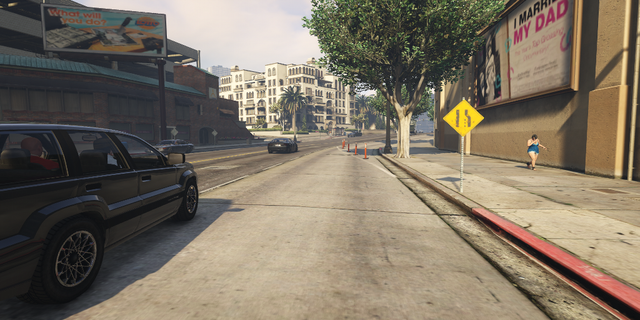}&
		\includegraphics[width=.19\linewidth]{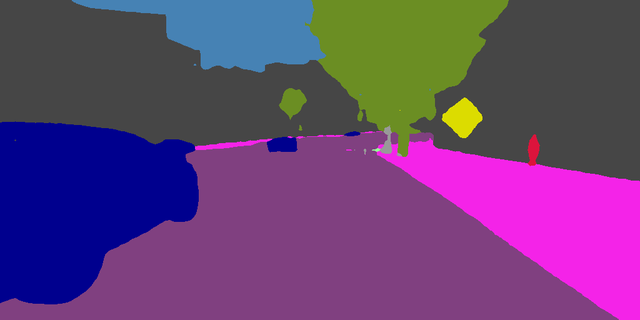}&
		\includegraphics[width=.19\linewidth]{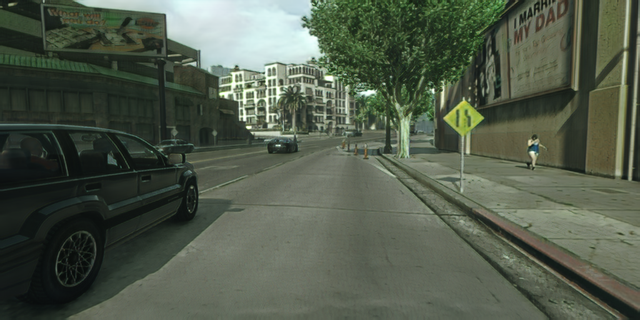}&
		\includegraphics[width=.19\linewidth]{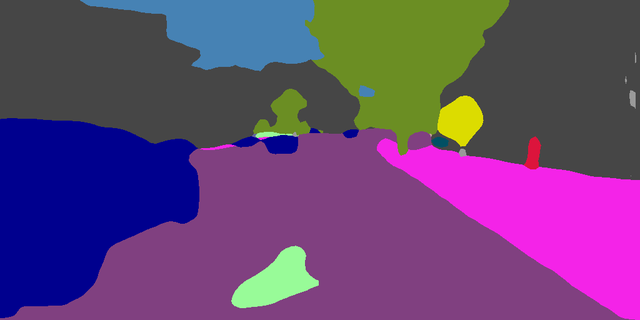}&
		\includegraphics[width=.19\linewidth]{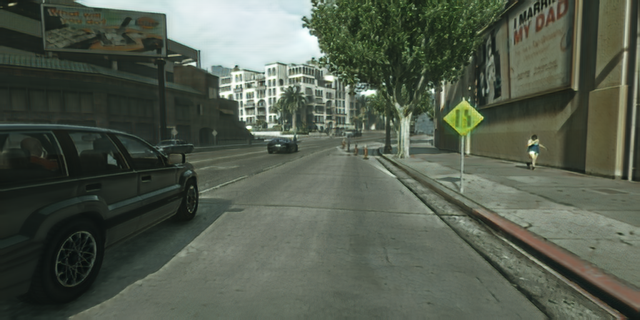}\\
		\includegraphics[width=.19\linewidth]{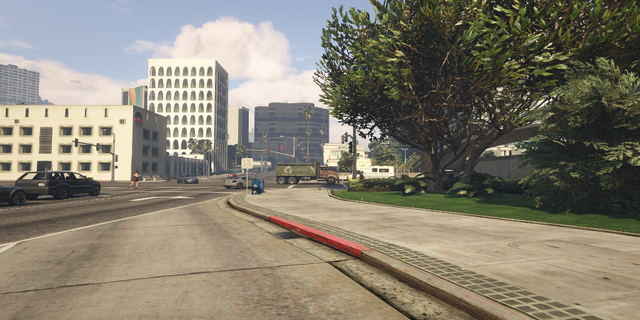}&
		\includegraphics[width=.19\linewidth]{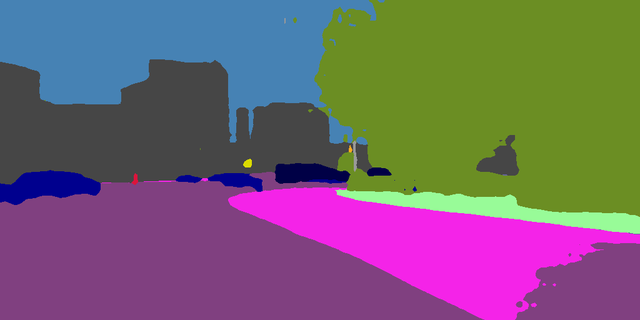}&
		\includegraphics[width=.19\linewidth]{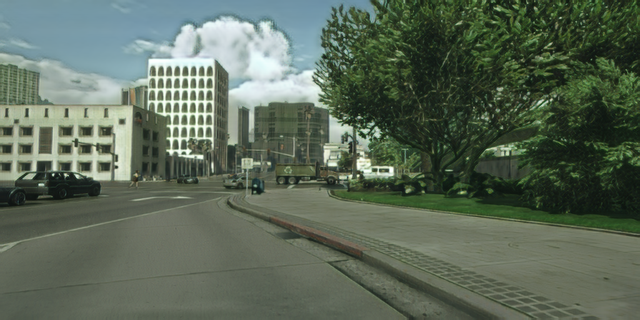}&
		\includegraphics[width=.19\linewidth]{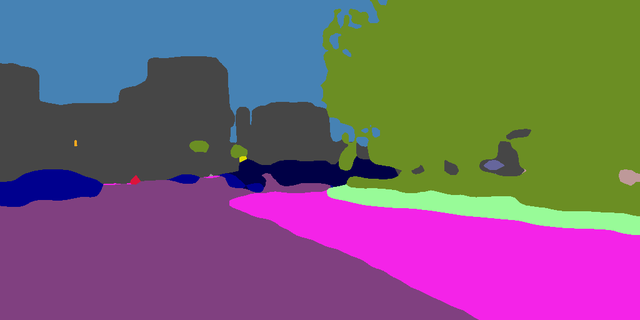}&
		\includegraphics[width=.19\linewidth]{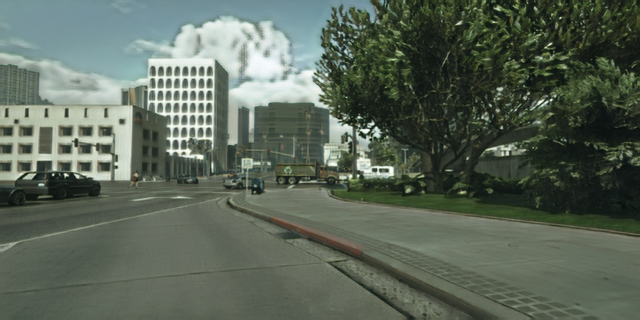}\\
		\includegraphics[width=.19\linewidth]{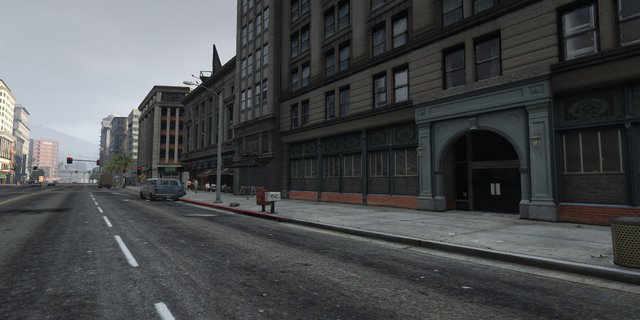}&
		\includegraphics[width=.19\linewidth]{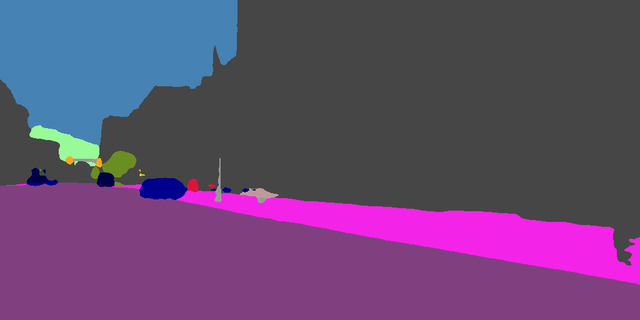}&
		\includegraphics[width=.19\linewidth]{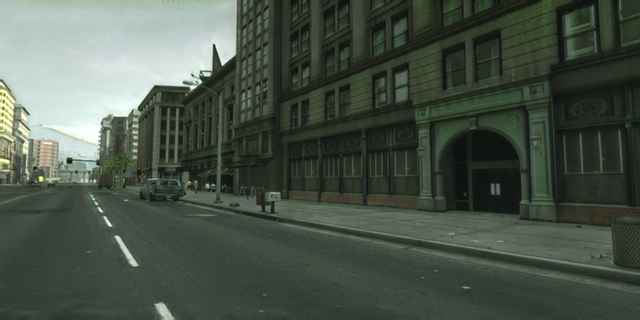}&
		\includegraphics[width=.19\linewidth]{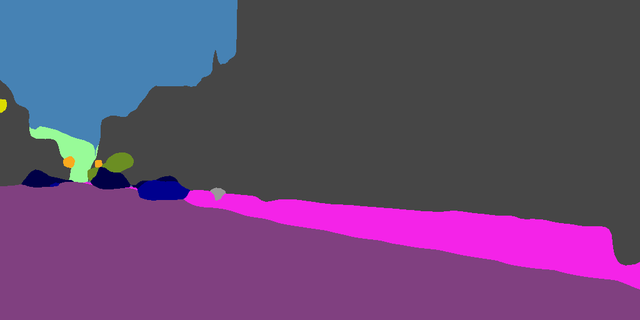}&
		\includegraphics[width=.19\linewidth]{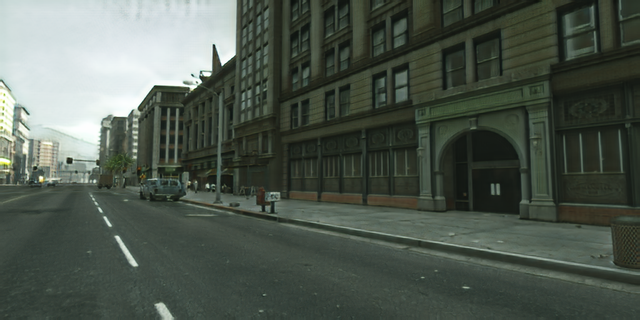}\\
		\includegraphics[width=.19\linewidth]{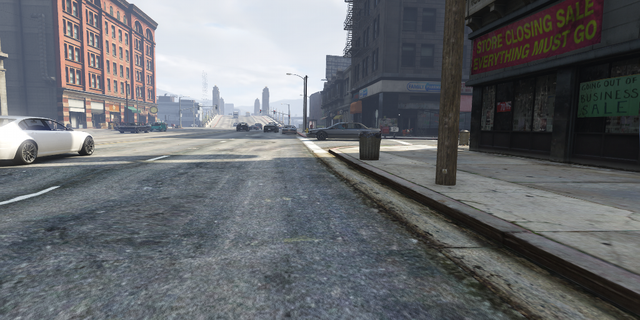}&
		\includegraphics[width=.19\linewidth]{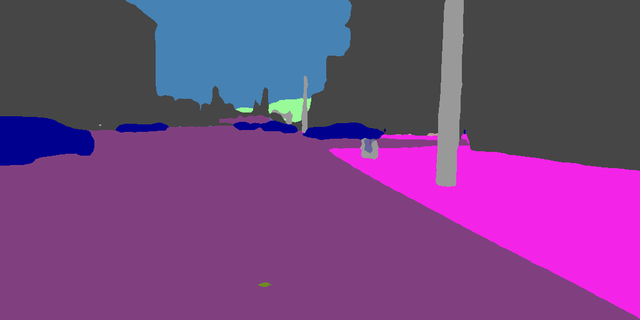}&
		\includegraphics[width=.19\linewidth]{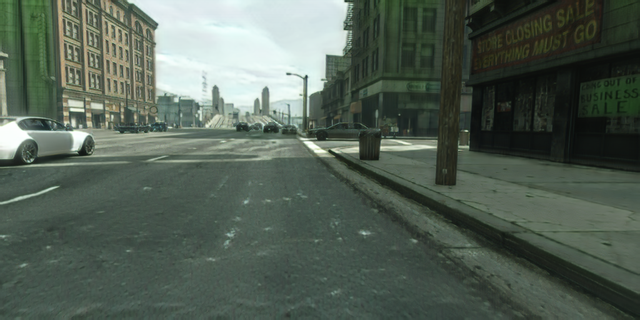}&
		\includegraphics[width=.19\linewidth]{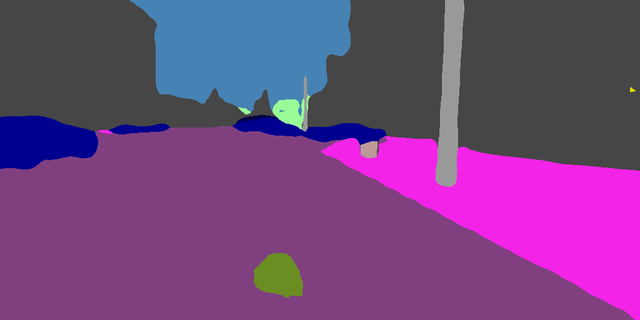}&
		\includegraphics[width=.19\linewidth]{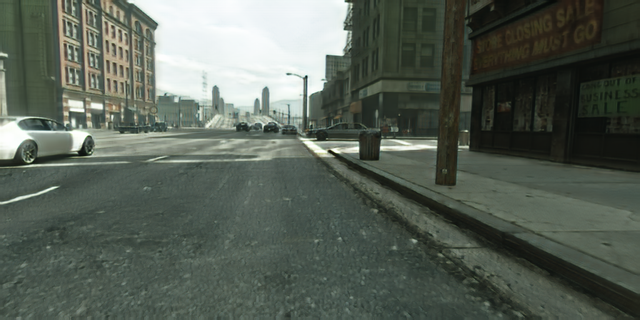}\\
		\includegraphics[width=.19\linewidth]{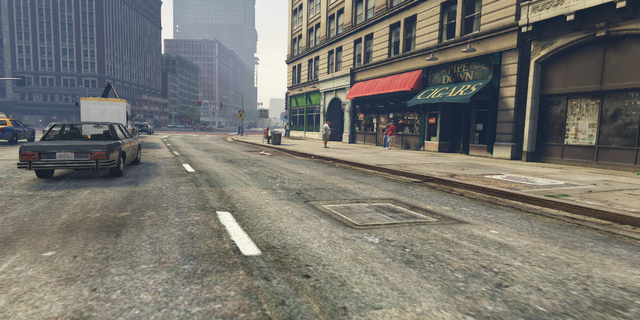}&
		\includegraphics[width=.19\linewidth]{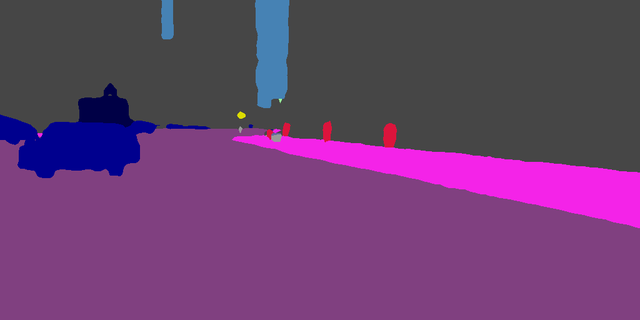}&
		\includegraphics[width=.19\linewidth]{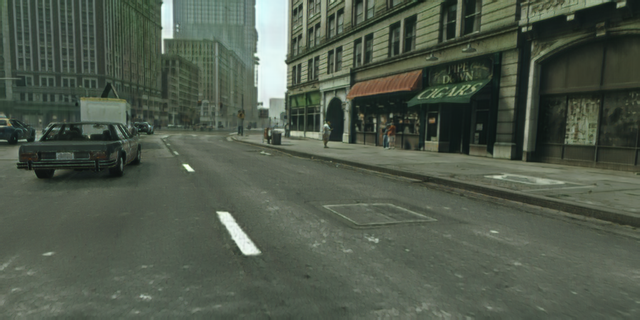}&
		\includegraphics[width=.19\linewidth]{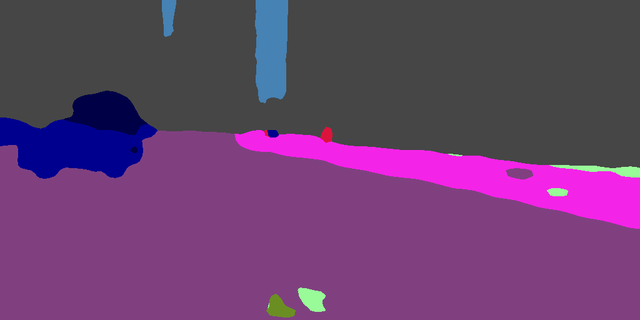}&
		\includegraphics[width=.19\linewidth]{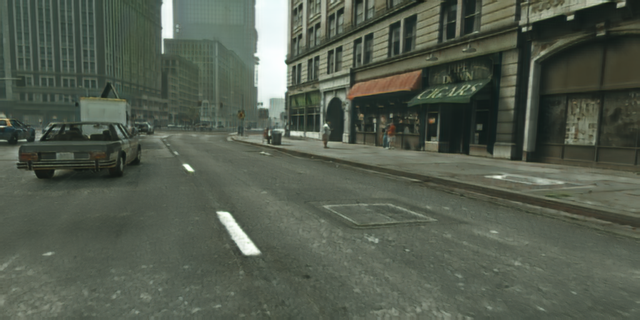}\\
		\includegraphics[width=.19\linewidth]{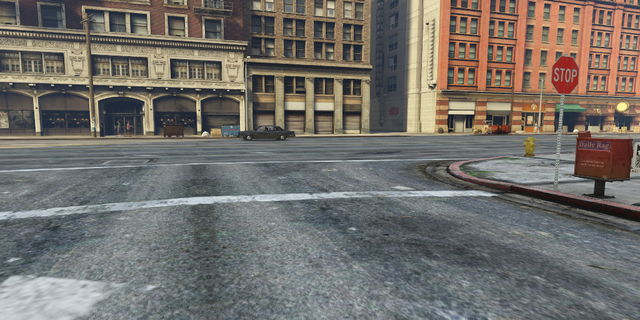}&
		\includegraphics[width=.19\linewidth]{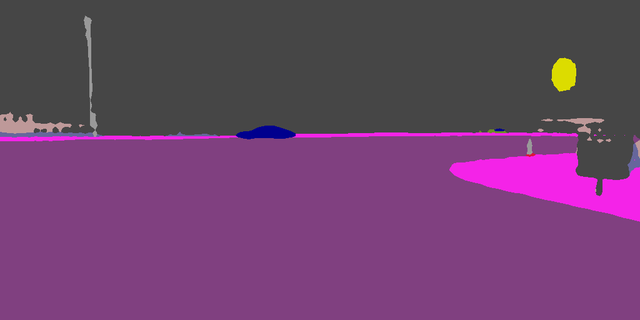}&
		\includegraphics[width=.19\linewidth]{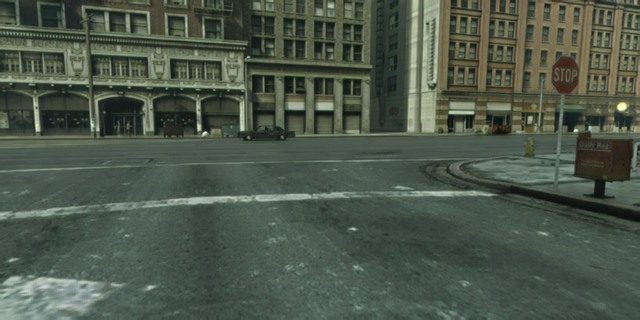}&
		\includegraphics[width=.19\linewidth]{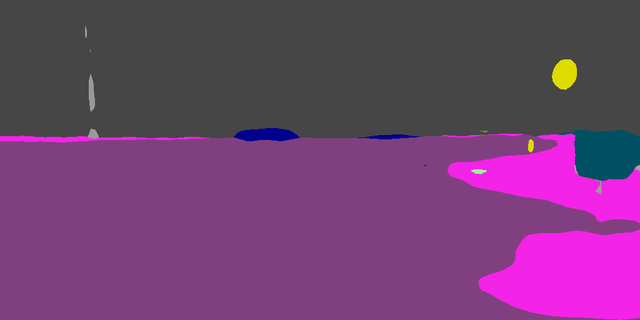}&
		\includegraphics[width=.19\linewidth]{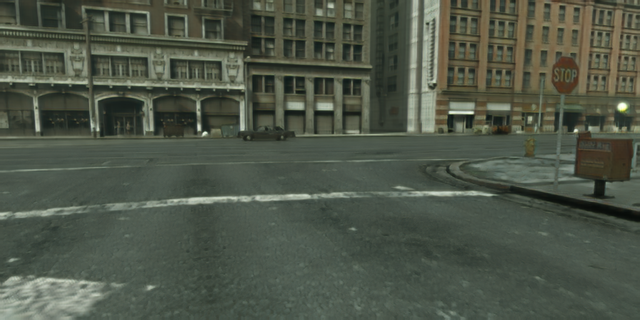}\\
		\makecell{GTA5\\sample}&\makecell{GTA5 pred\\(D)}&\makecell{GTA5$\rightarrow$CS\\(D)}&\makecell{GTA5 pred\\(F)}&\makecell{GTA5$\rightarrow$CS\\(F)}\\\medskip
	\end{tabular}
	\caption{\textbf{Additional translations from GTA5 to Cityscapes.} We take a sample $X_S$ from GTA5, get the predicted segmentation using $M$, and generate $X_{S \rightarrow T}$. We present the results obtained with both DeepLabV2 and FCN8s used as semantic guidance.}
	\label{fig:gta2cs}
\end{figure}

\begin{figure}[!htb]
	\centering
	\begin{tabular}{@{\hskip2pt}c@{\hskip2pt}c@{\hskip2pt}c@{\hskip2pt}c@{\hskip2pt}c}
		\includegraphics[width=.19\linewidth]{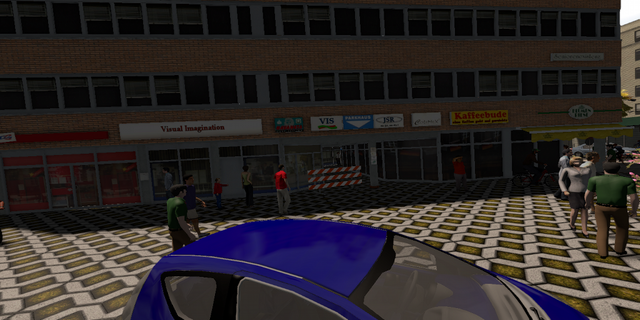}&
		\includegraphics[width=.19\linewidth]{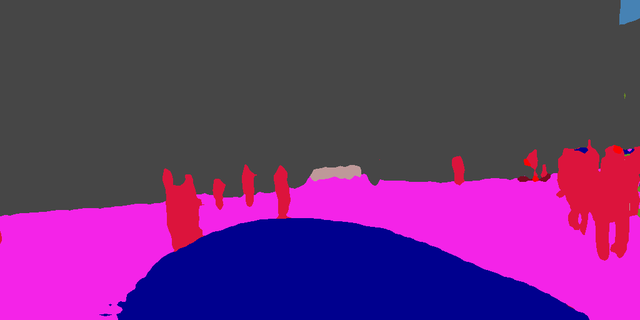}&
		\includegraphics[width=.19\linewidth]{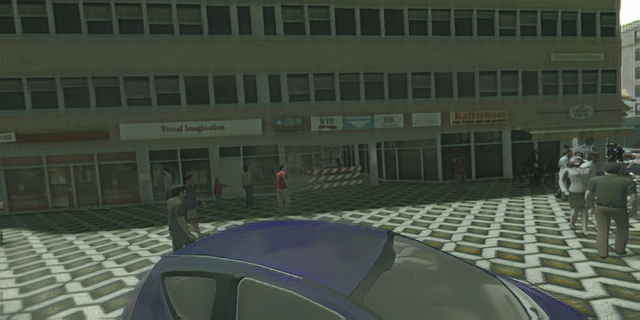}&
		\includegraphics[width=.19\linewidth]{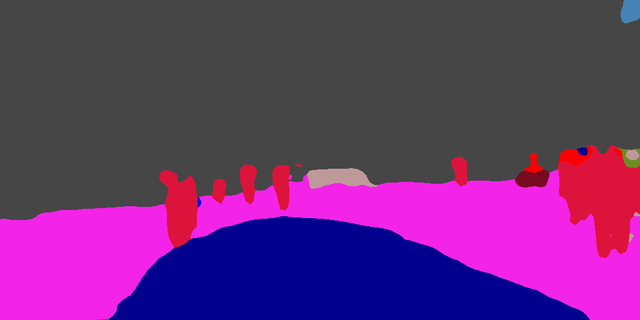}&
		\includegraphics[width=.19\linewidth]{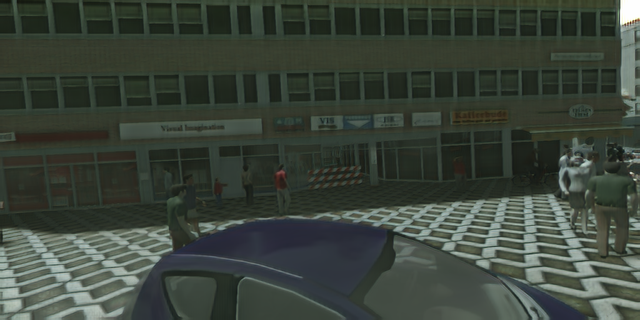}\\
		\includegraphics[width=.19\linewidth]{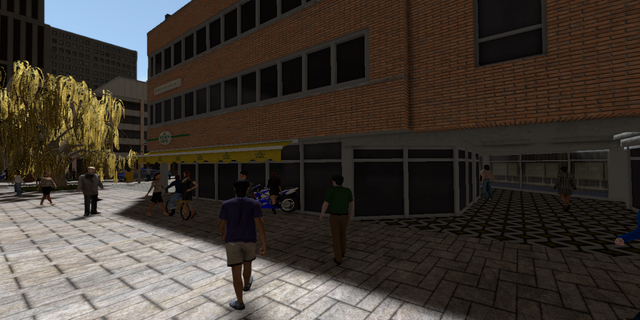}&
		\includegraphics[width=.19\linewidth]{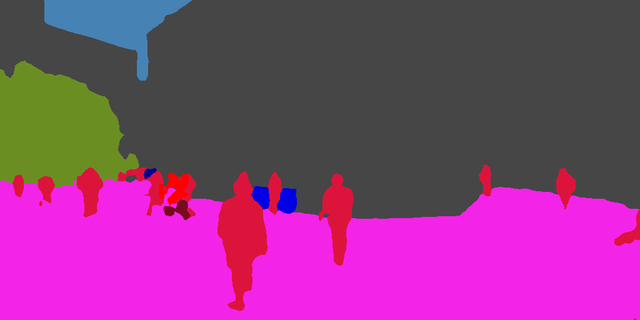}&
		\includegraphics[width=.19\linewidth]{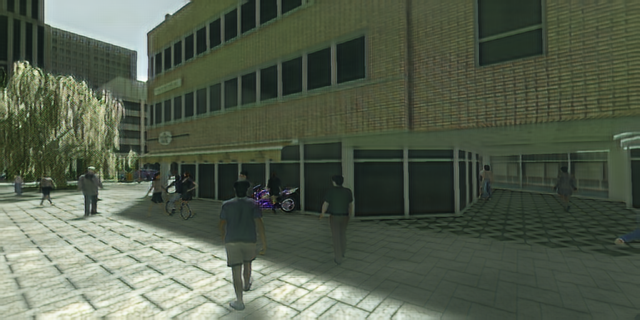}&
		\includegraphics[width=.19\linewidth]{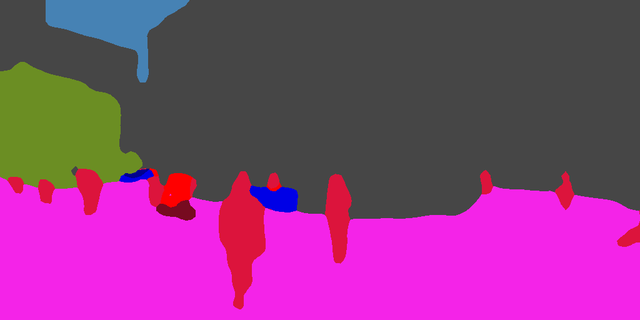}&
		\includegraphics[width=.19\linewidth]{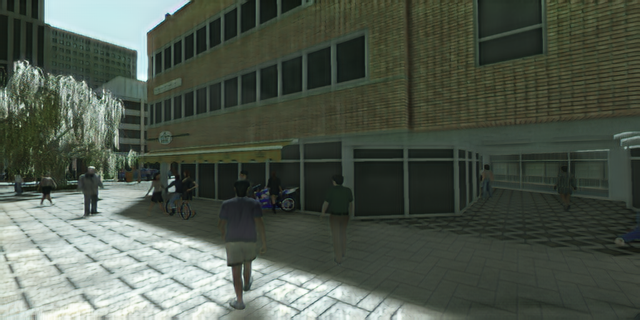}\\
		\includegraphics[width=.19\linewidth]{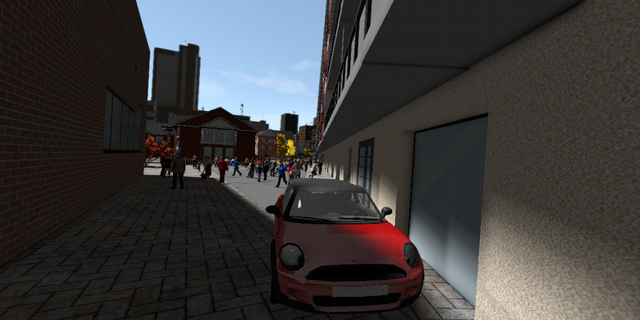}&
		\includegraphics[width=.19\linewidth]{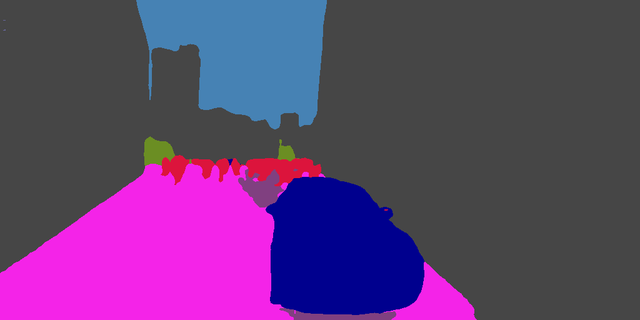}&
		\includegraphics[width=.19\linewidth]{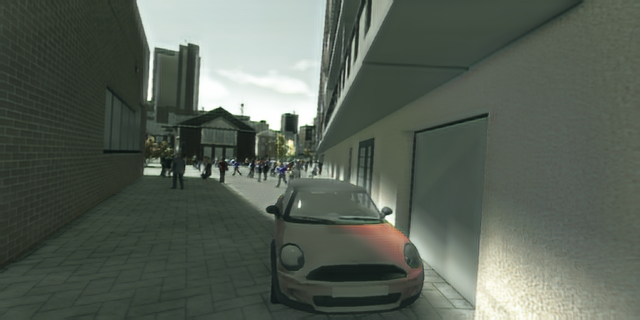}&
		\includegraphics[width=.19\linewidth]{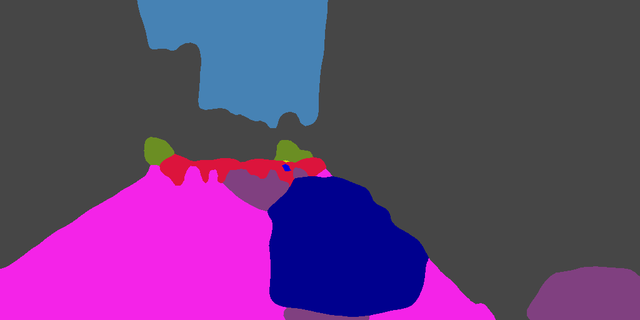}&
		\includegraphics[width=.19\linewidth]{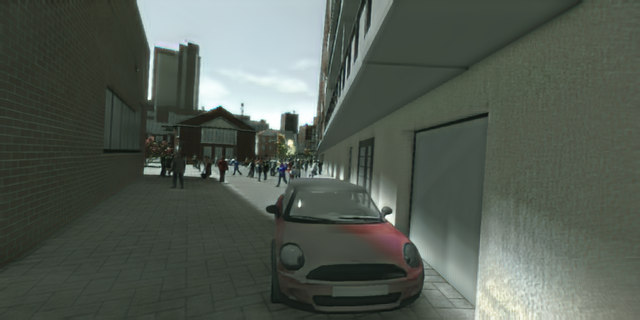}\\
		\includegraphics[width=.19\linewidth]{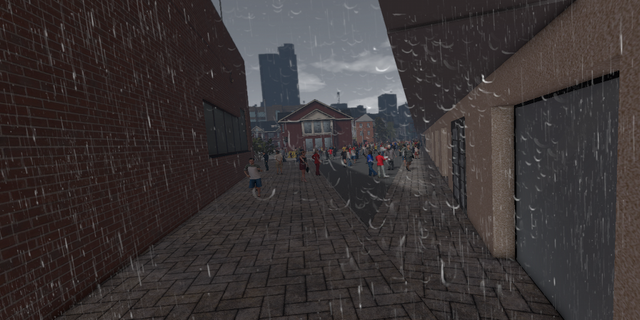}&
		\includegraphics[width=.19\linewidth]{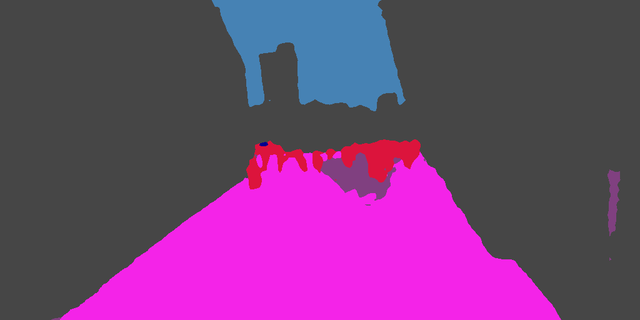}&
		\includegraphics[width=.19\linewidth]{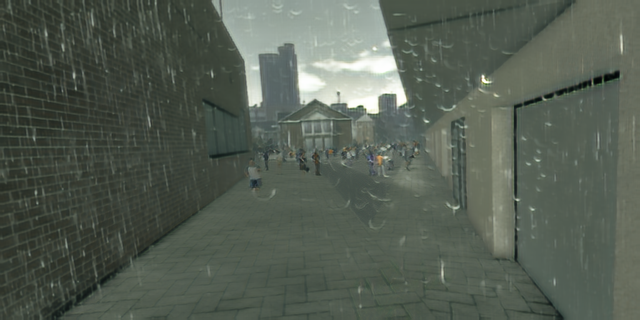}&
		\includegraphics[width=.19\linewidth]{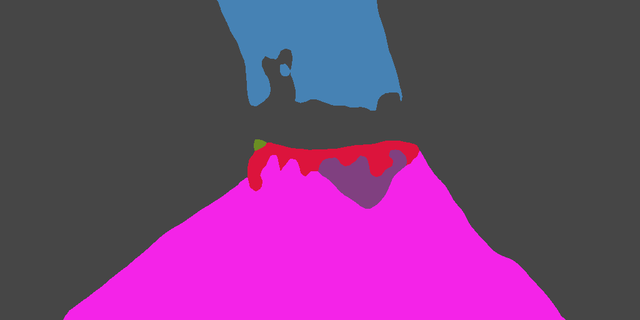}&
		\includegraphics[width=.19\linewidth]{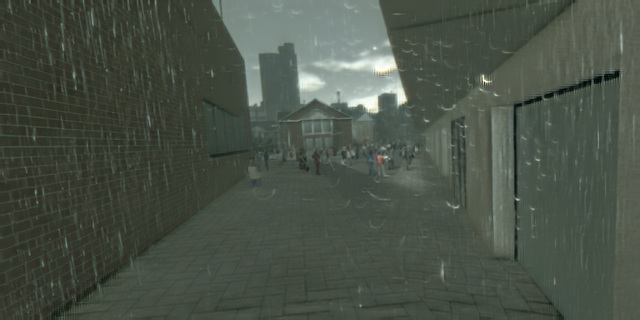}\\
		\includegraphics[width=.19\linewidth]{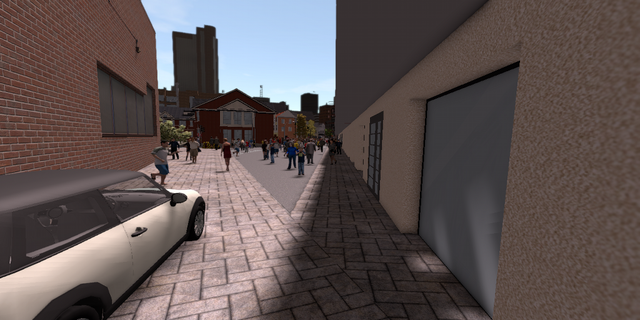}&
		\includegraphics[width=.19\linewidth]{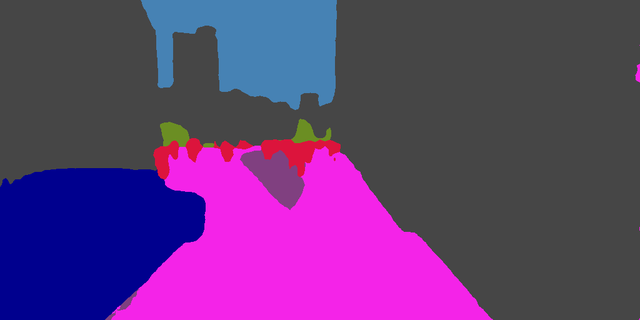}&
		\includegraphics[width=.19\linewidth]{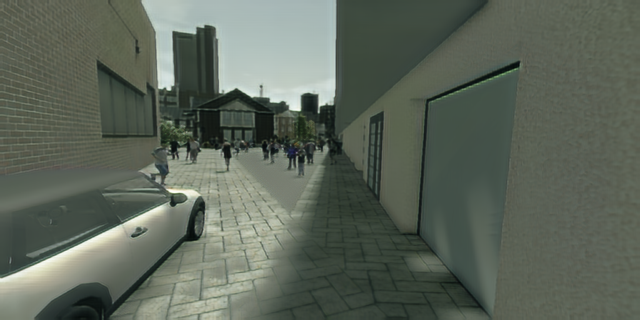}&
		\includegraphics[width=.19\linewidth]{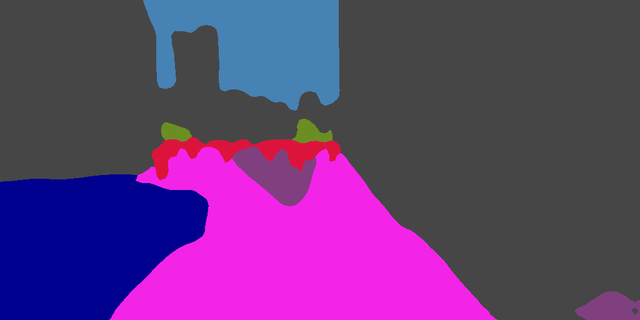}&
		\includegraphics[width=.19\linewidth]{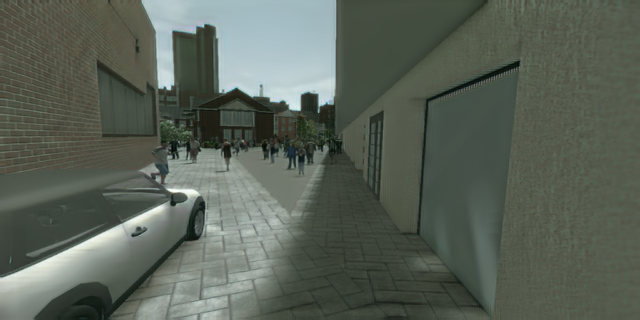}\\
		\includegraphics[width=.19\linewidth]{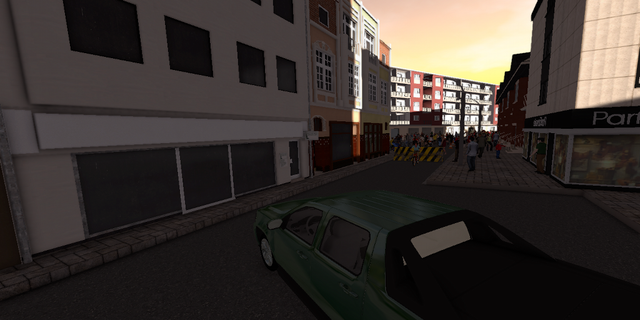}&
		\includegraphics[width=.19\linewidth]{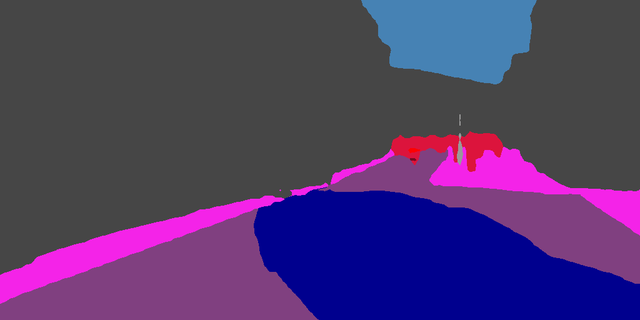}&
		\includegraphics[width=.19\linewidth]{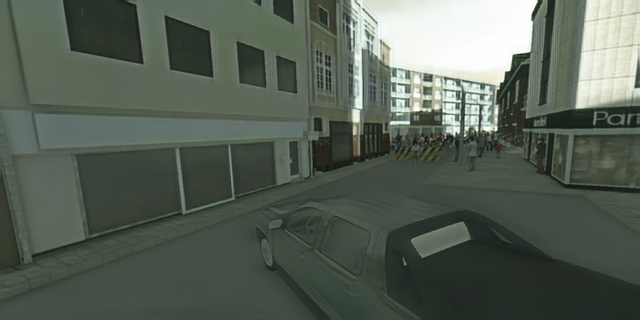}&
		\includegraphics[width=.19\linewidth]{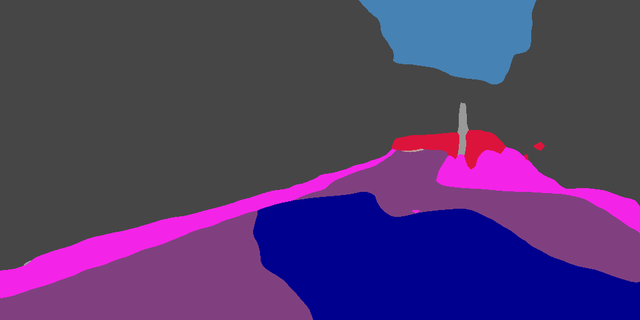}&
		\includegraphics[width=.19\linewidth]{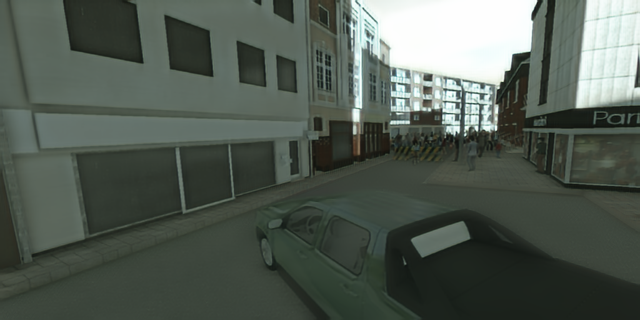}\\
		\includegraphics[width=.19\linewidth]{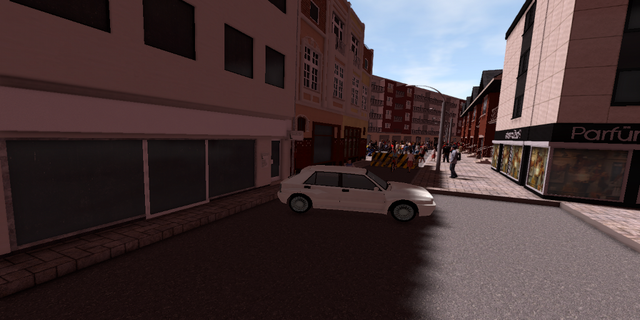}&
		\includegraphics[width=.19\linewidth]{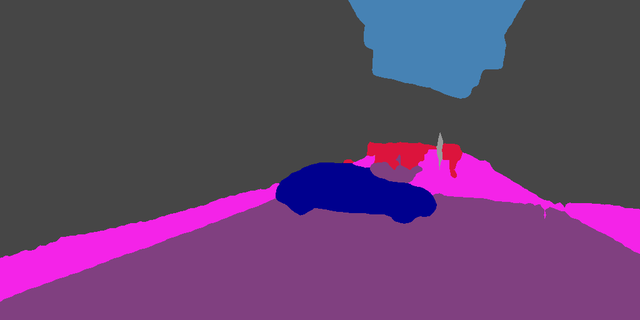}&
		\includegraphics[width=.19\linewidth]{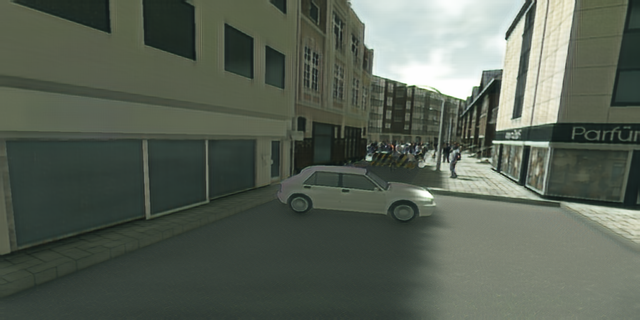}&
		\includegraphics[width=.19\linewidth]{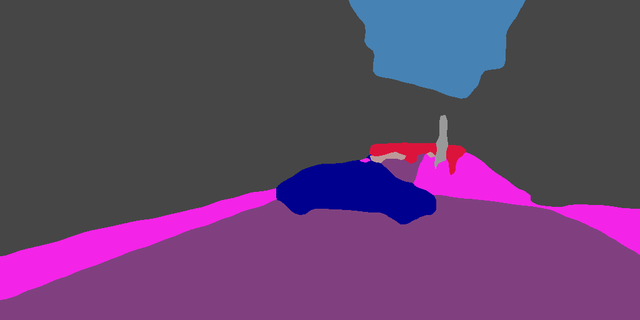}&
		\includegraphics[width=.19\linewidth]{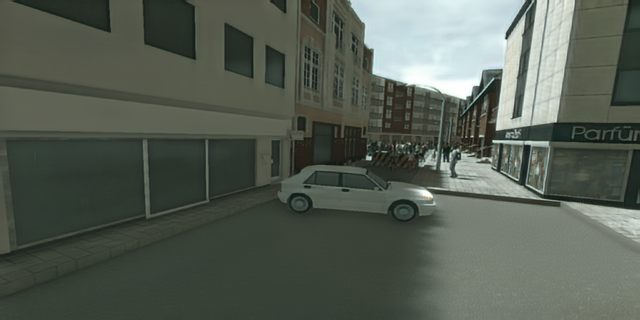}\\
		\includegraphics[width=.19\linewidth]{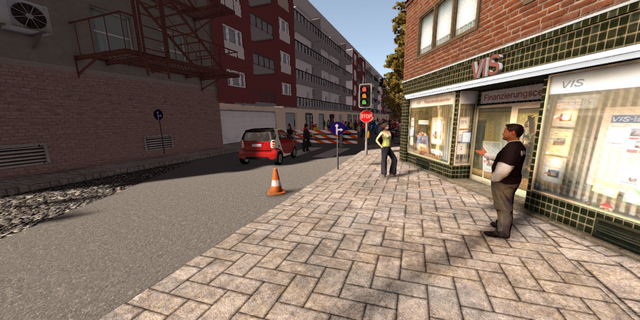}&
		\includegraphics[width=.19\linewidth]{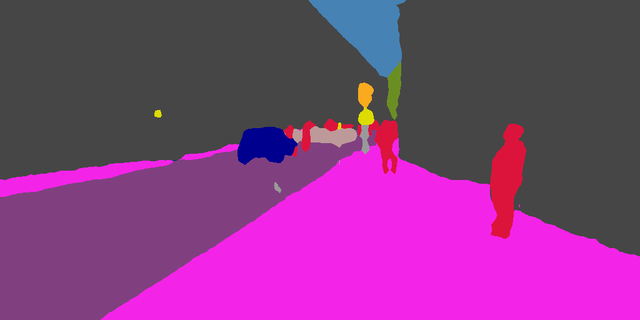}&
		\includegraphics[width=.19\linewidth]{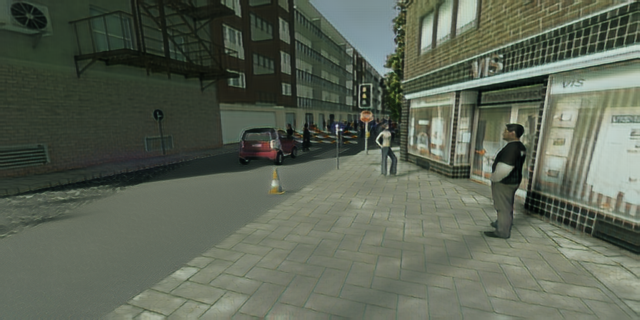}&
		\includegraphics[width=.19\linewidth]{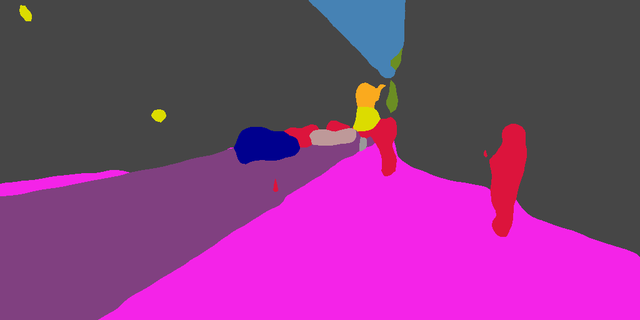}&
		\includegraphics[width=.19\linewidth]{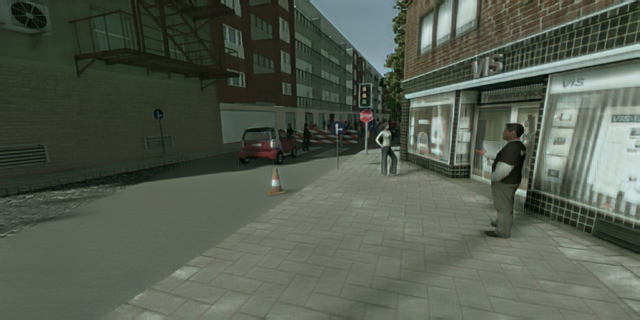}\\
		\includegraphics[width=.19\linewidth]{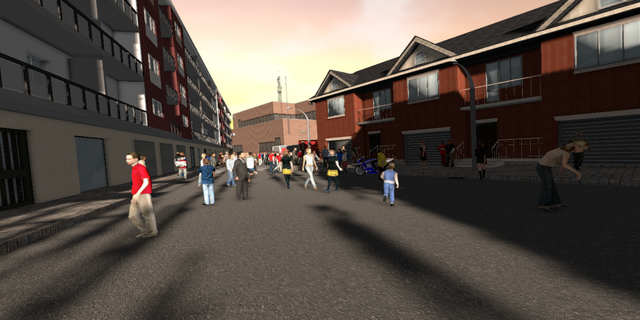}&
		\includegraphics[width=.19\linewidth]{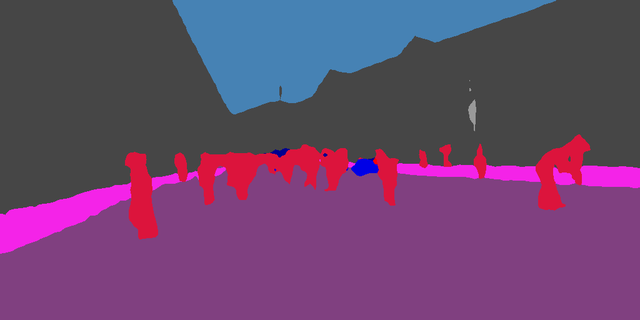}&
		\includegraphics[width=.19\linewidth]{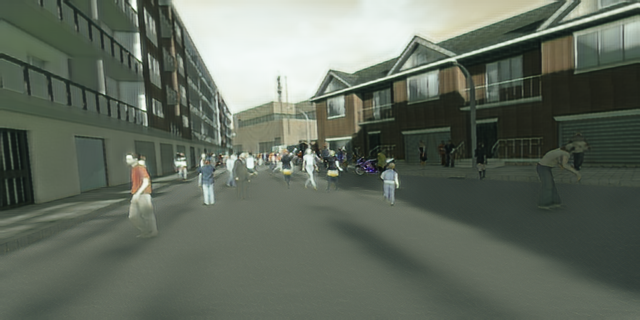}&
		\includegraphics[width=.19\linewidth]{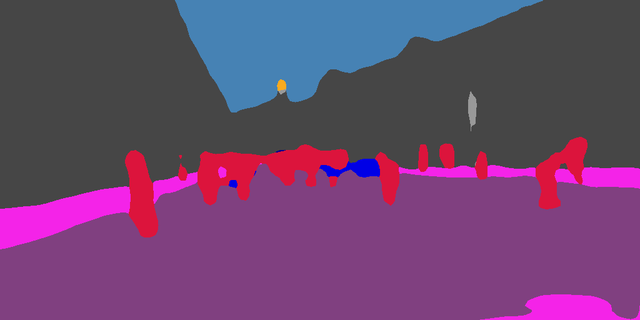}&
		\includegraphics[width=.19\linewidth]{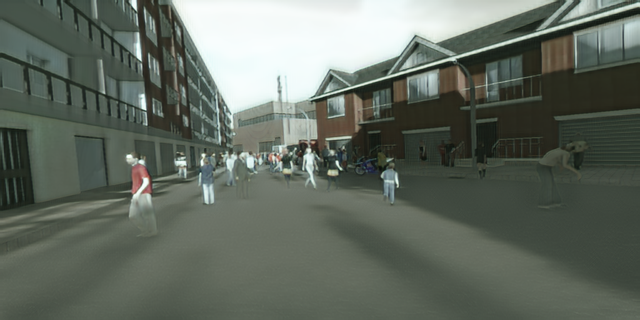}\\
		\makecell{SYNTHIA\\sample}&\makecell{SYNTHIA pred\\(D)}&\makecell{SYNTHIA$\rightarrow$CS\\(D)}&\makecell{SYNTHIA pred\\(F)}&\makecell{SYNTHIA$\rightarrow$CS\\(F)}\\\medskip
	\end{tabular}
	\caption{\textbf{Translations from SYNTHIA to Cityscapes.} We take a sample $X_S$ from SYNTHIA, get the predicted segmentation using $M$, and generate $X_{S \rightarrow T}$. We present the results obtained with both DeepLabV2 and FCN8s used as semantic guidance.}
	\label{fig:synthia2cs}
\end{figure}

\begin{figure}[!htb]
	\centering
	\begin{tabular}{@{\hskip2pt}c@{\hskip2pt}c@{\hskip2pt}c@{\hskip2pt}c@{\hskip2pt}c}
		\includegraphics[width=.19\linewidth]{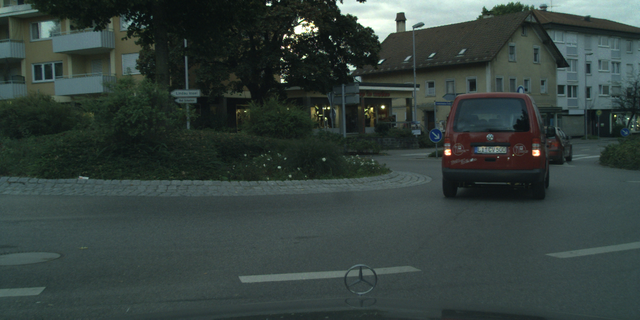}&
		\includegraphics[width=.19\linewidth]{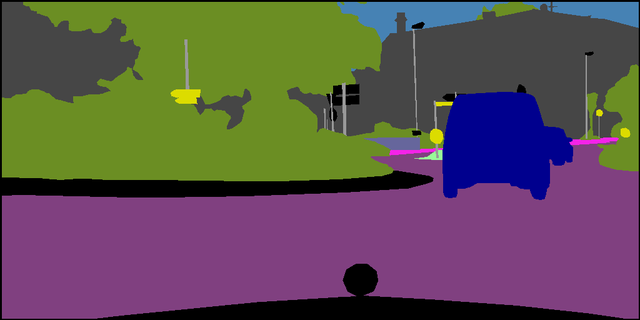}&
		\includegraphics[width=.19\linewidth]{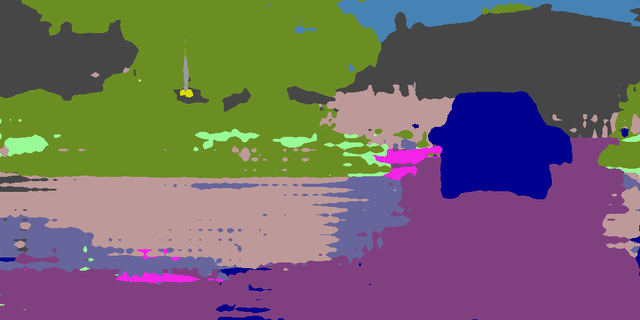}&
		\includegraphics[width=.19\linewidth]{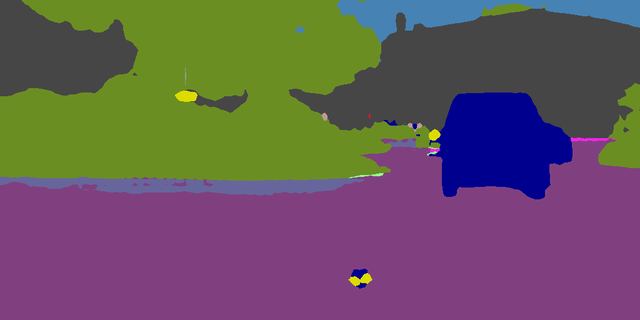}&
		\includegraphics[width=.19\linewidth]{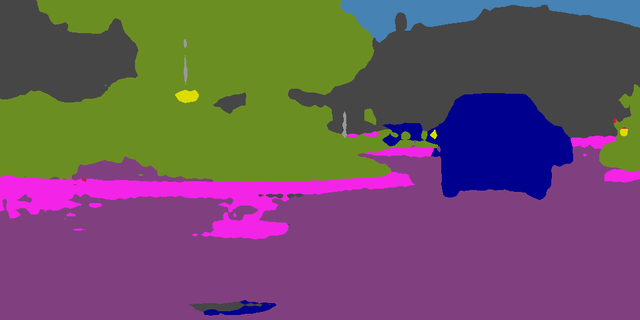}\\
		\includegraphics[width=.19\linewidth]{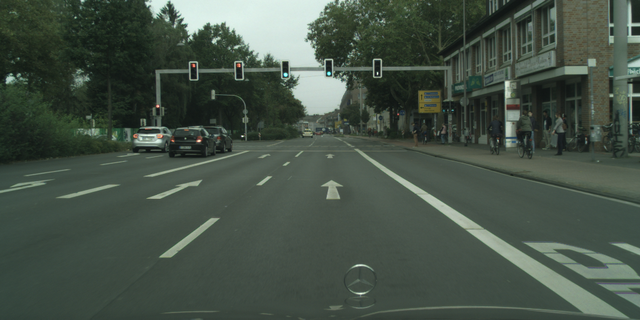}&
		\includegraphics[width=.19\linewidth]{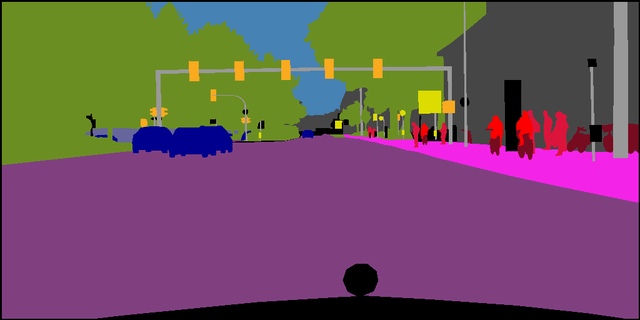}&
		\includegraphics[width=.19\linewidth]{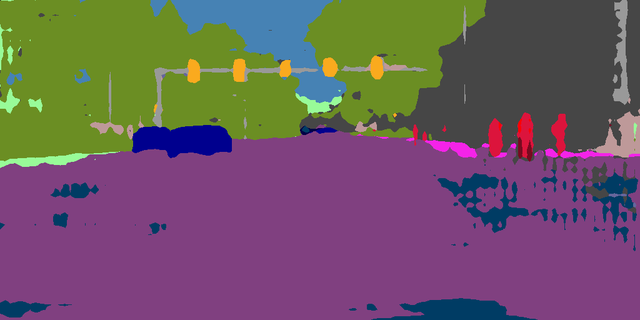}&
		\includegraphics[width=.19\linewidth]{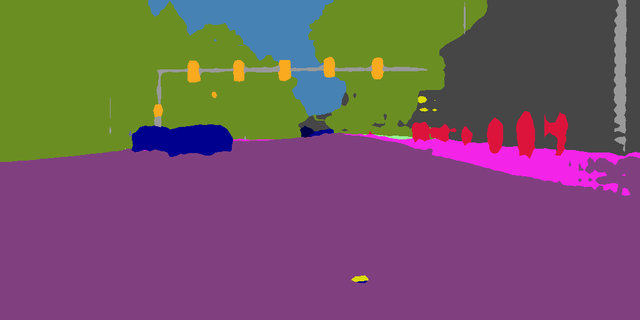}&
		\includegraphics[width=.19\linewidth]{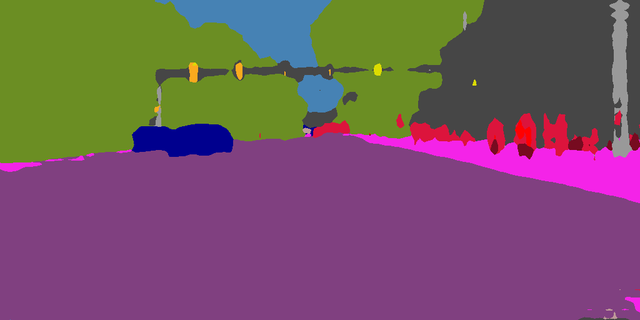}\\
		\includegraphics[width=.19\linewidth]{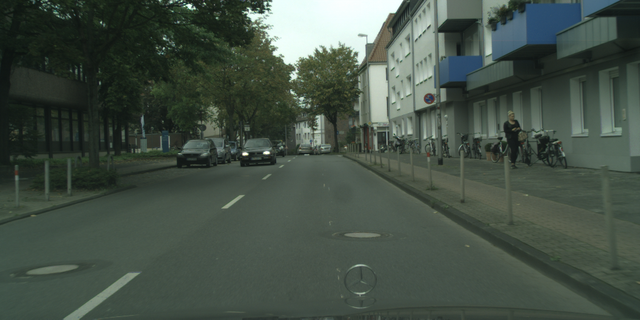}&
		\includegraphics[width=.19\linewidth]{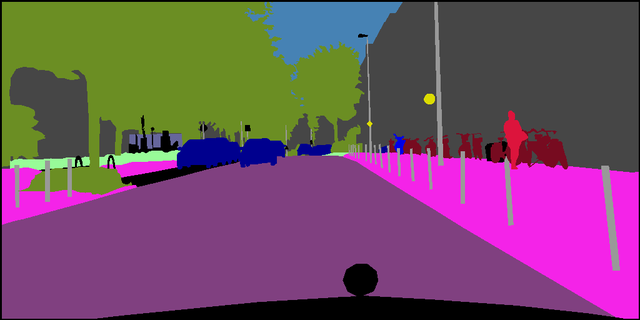}&
		\includegraphics[width=.19\linewidth]{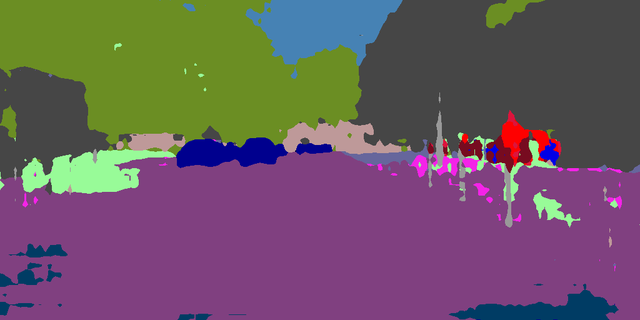}&
		\includegraphics[width=.19\linewidth]{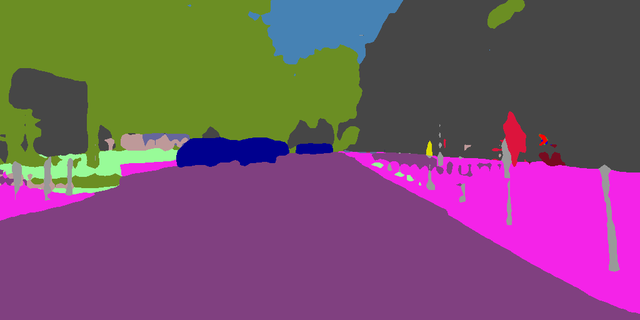}&
		\includegraphics[width=.19\linewidth]{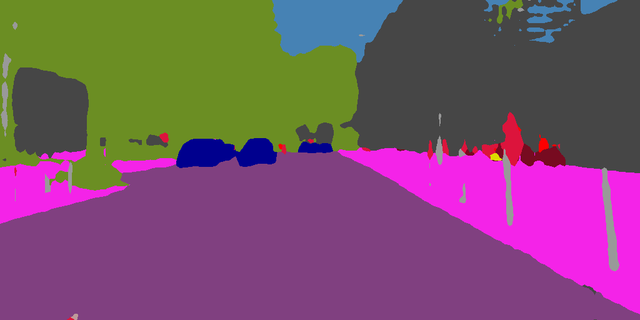}\\
		\includegraphics[width=.19\linewidth]{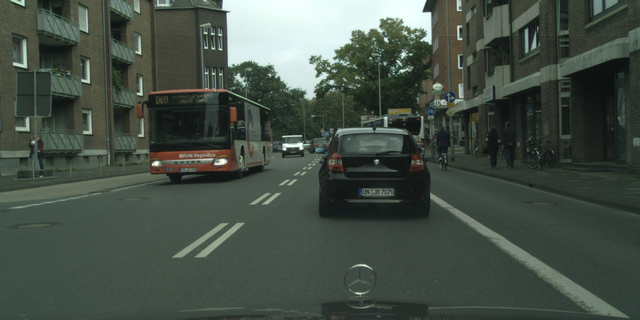}&
		\includegraphics[width=.19\linewidth]{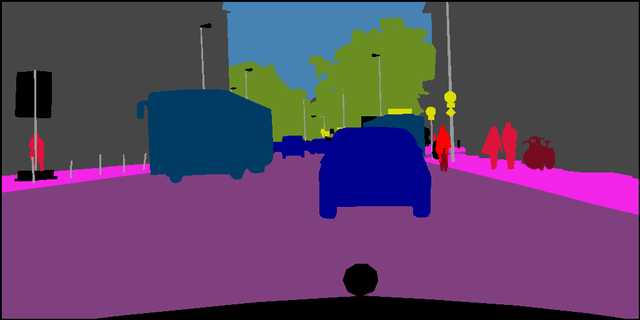}&
		\includegraphics[width=.19\linewidth]{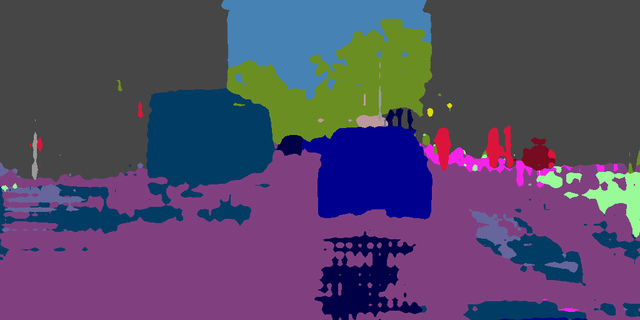}&
		\includegraphics[width=.19\linewidth]{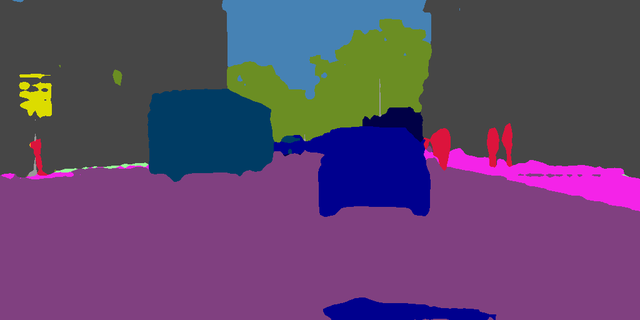}&
		\includegraphics[width=.19\linewidth]{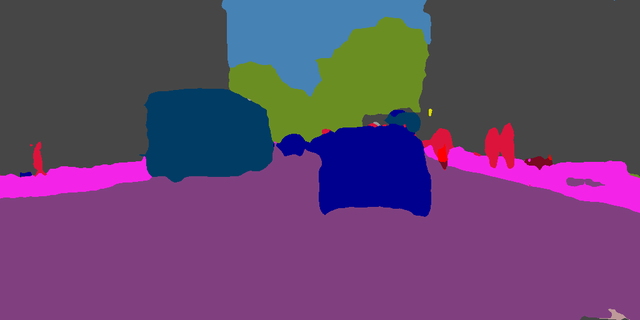}\\
		\includegraphics[width=.19\linewidth]{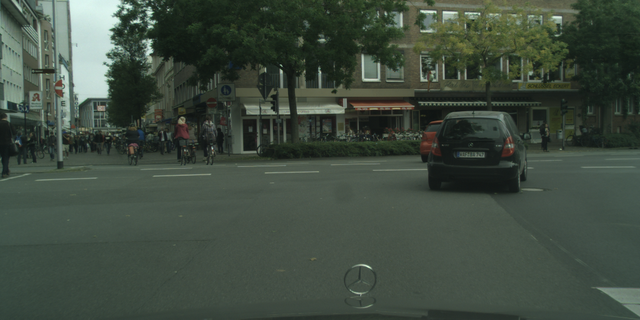}&
		\includegraphics[width=.19\linewidth]{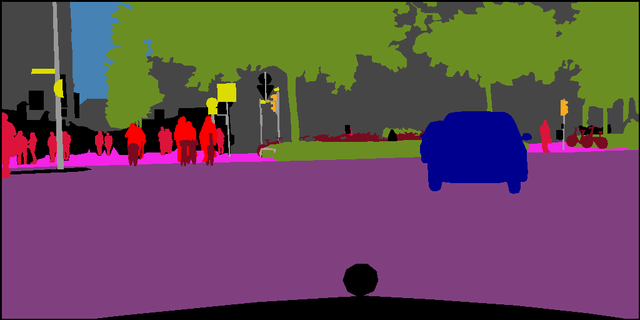}&
		\includegraphics[width=.19\linewidth]{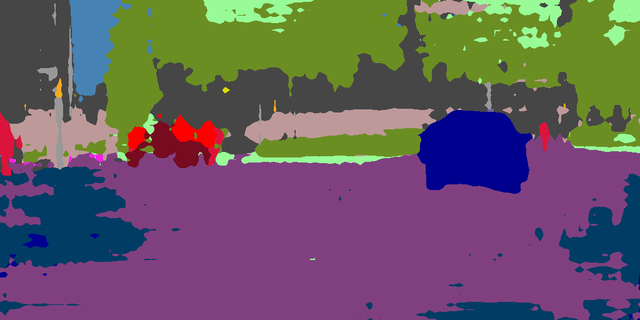}&
		\includegraphics[width=.19\linewidth]{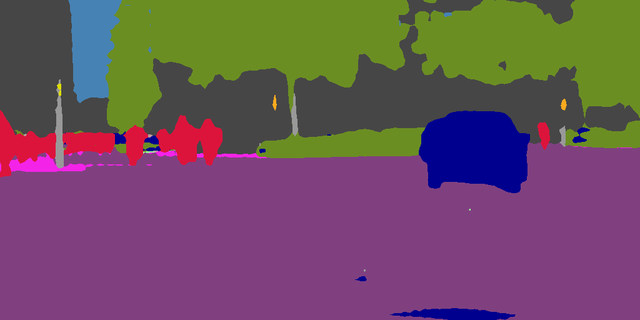}&
		\includegraphics[width=.19\linewidth]{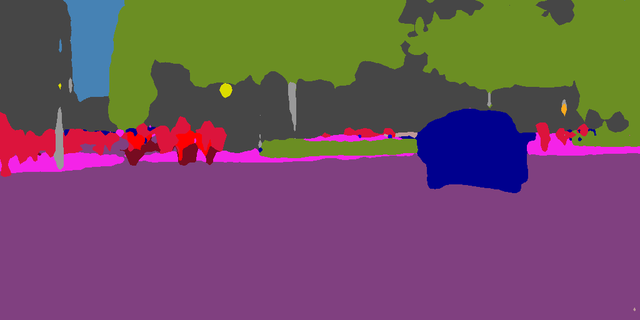}\\
		\includegraphics[width=.19\linewidth]{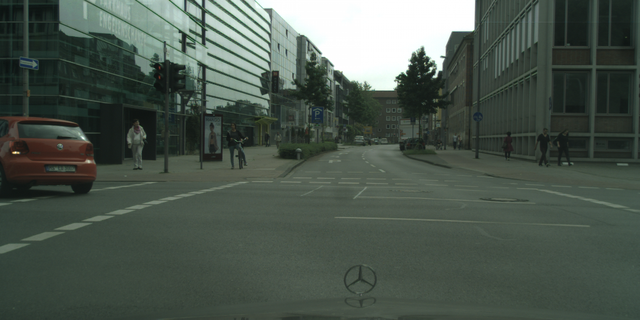}&
		\includegraphics[width=.19\linewidth]{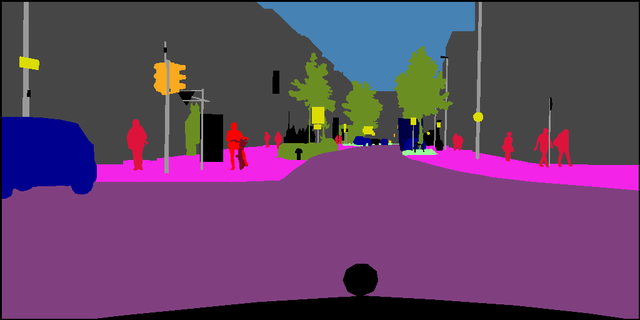}&
		\includegraphics[width=.19\linewidth]{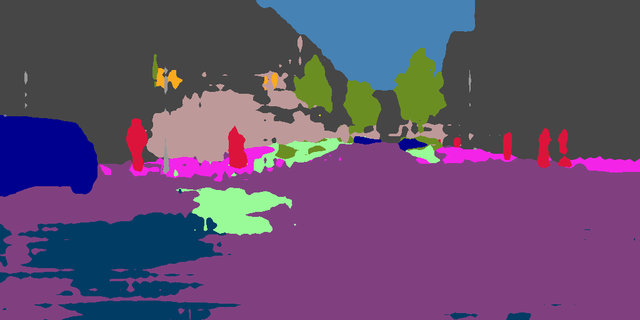}&
		\includegraphics[width=.19\linewidth]{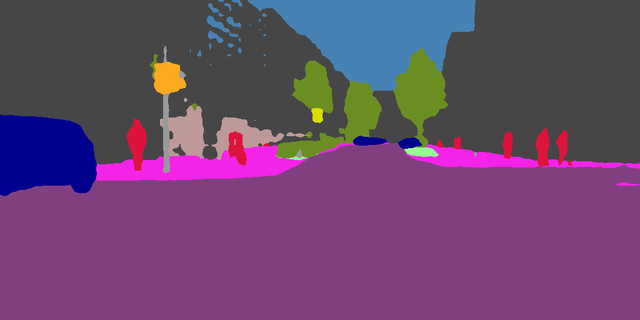}&
		\includegraphics[width=.19\linewidth]{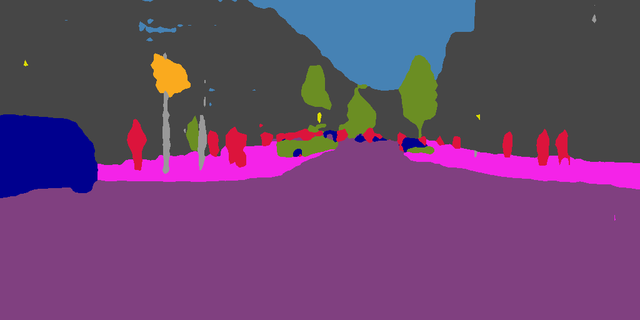}\\
		\includegraphics[width=.19\linewidth]{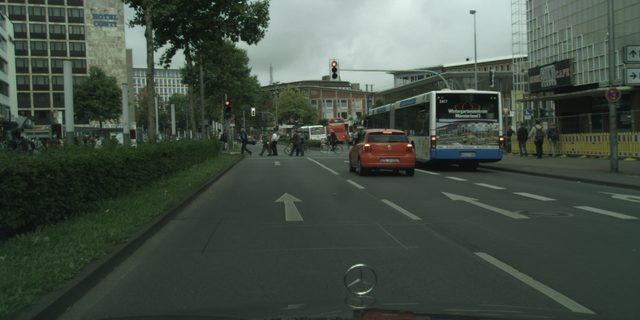}&
		\includegraphics[width=.19\linewidth]{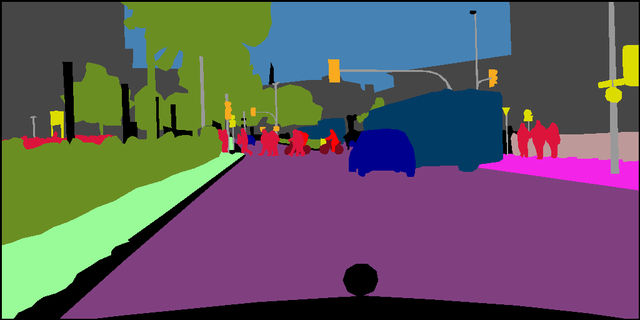}&
		\includegraphics[width=.19\linewidth]{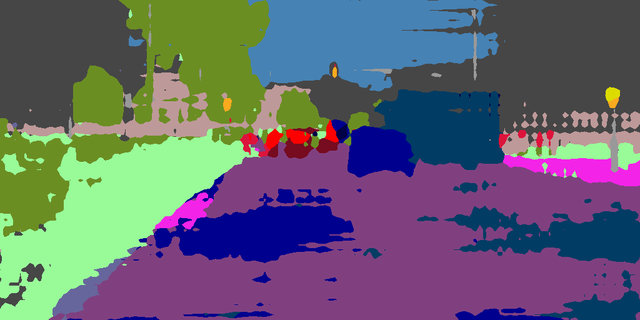}&
		\includegraphics[width=.19\linewidth]{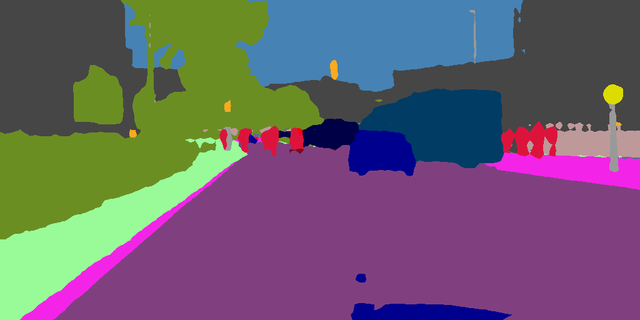}&
		\includegraphics[width=.19\linewidth]{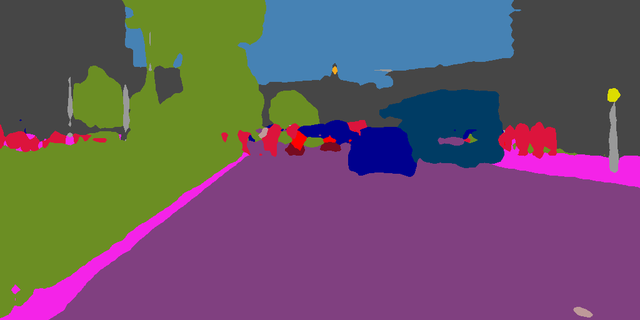}\\
		\includegraphics[width=.19\linewidth]{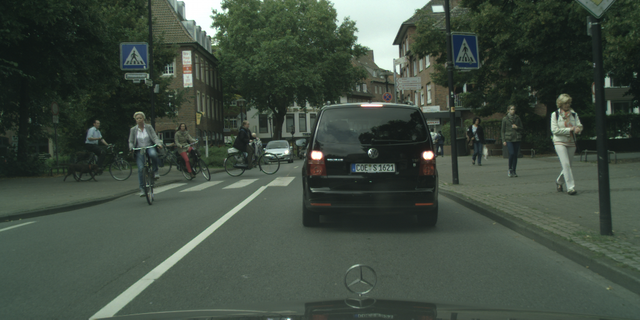}&
		\includegraphics[width=.19\linewidth]{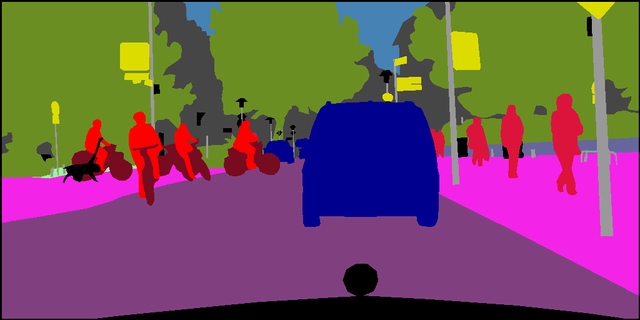}&
		\includegraphics[width=.19\linewidth]{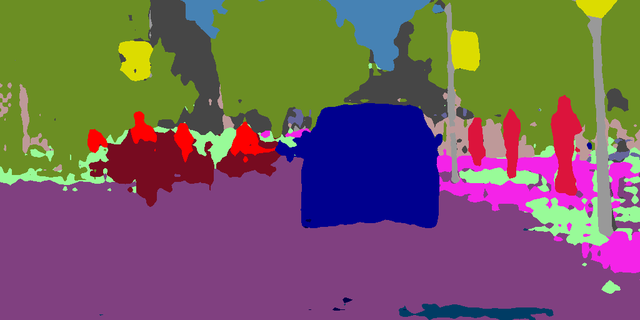}&
		\includegraphics[width=.19\linewidth]{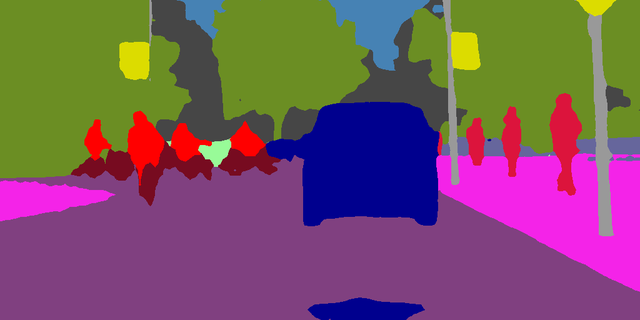}&
		\includegraphics[width=.19\linewidth]{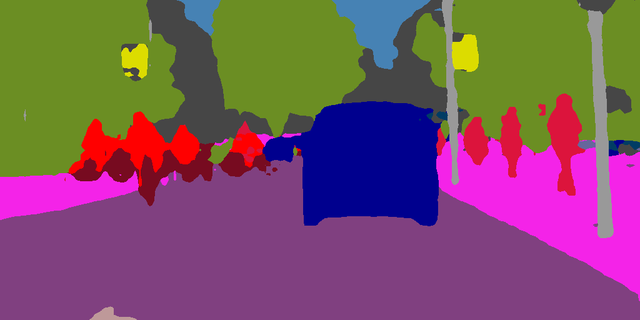}\\
		\includegraphics[width=.19\linewidth]{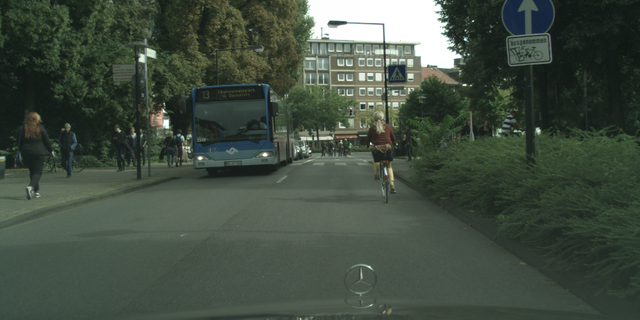}&
		\includegraphics[width=.19\linewidth]{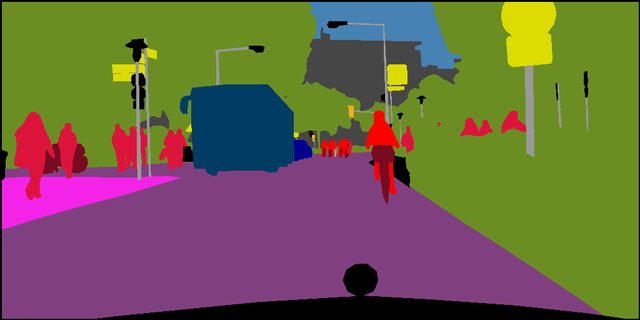}&
		\includegraphics[width=.19\linewidth]{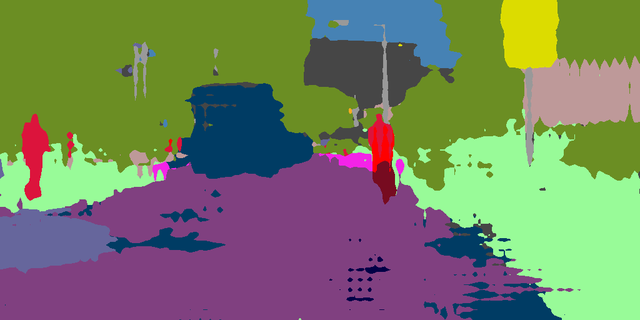}&
		\includegraphics[width=.19\linewidth]{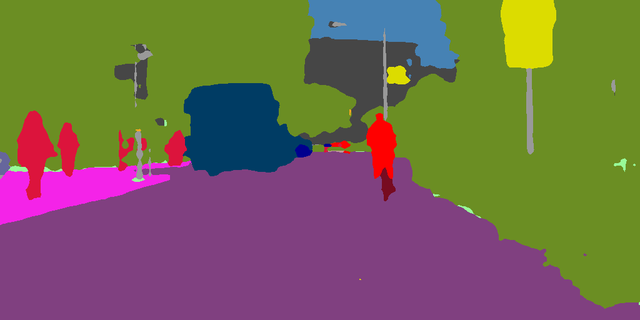}&
		\includegraphics[width=.19\linewidth]{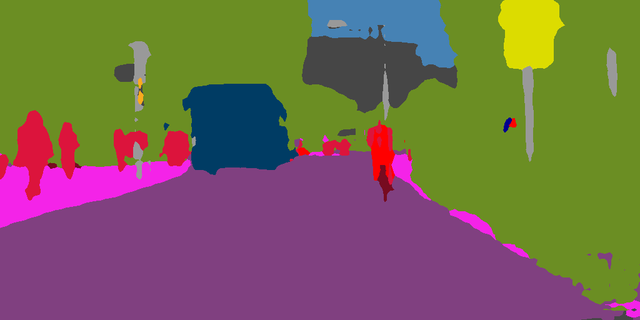}\\
		\includegraphics[width=.19\linewidth]{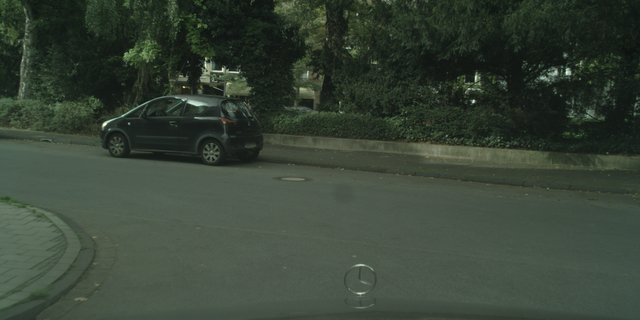}&
		\includegraphics[width=.19\linewidth]{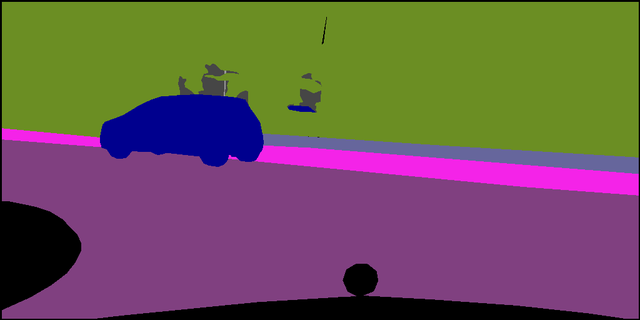}&
		\includegraphics[width=.19\linewidth]{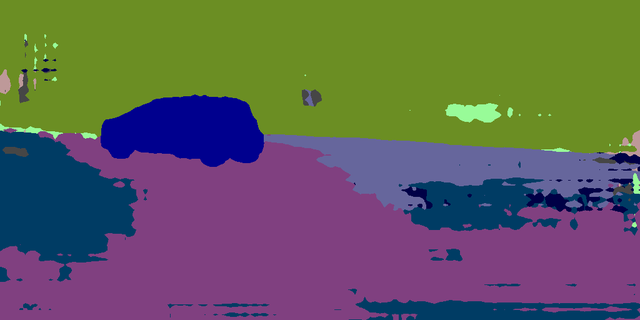}&
		\includegraphics[width=.19\linewidth]{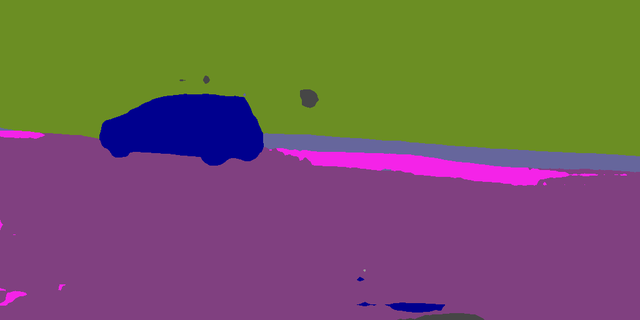}&
		\includegraphics[width=.19\linewidth]{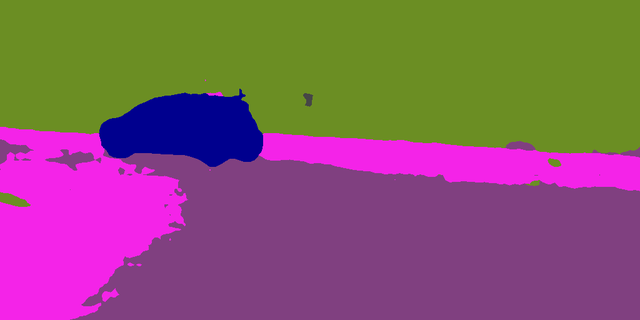}\\
		\includegraphics[width=.19\linewidth]{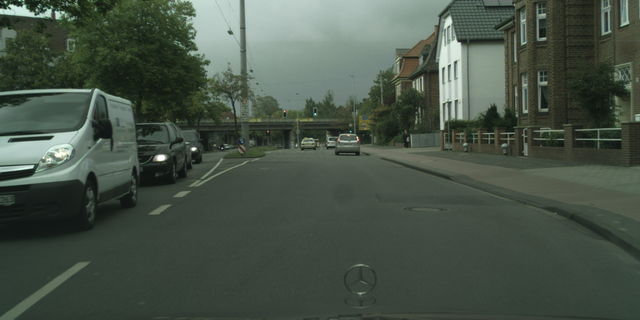}&
		\includegraphics[width=.19\linewidth]{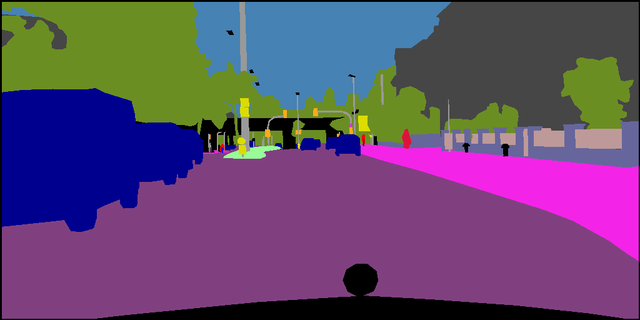}&
		\includegraphics[width=.19\linewidth]{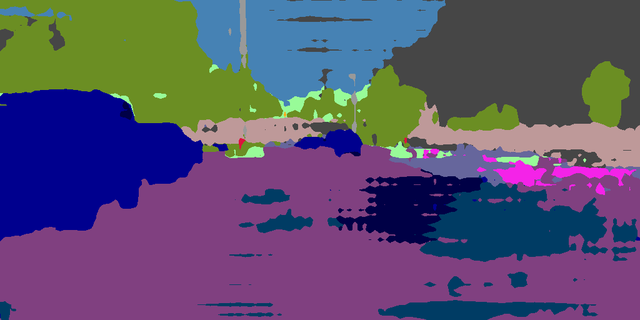}&
		\includegraphics[width=.19\linewidth]{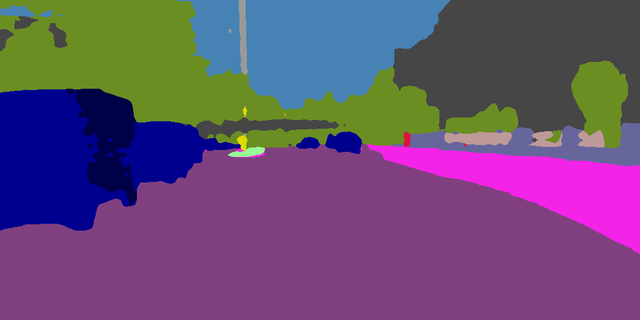}&
		\includegraphics[width=.19\linewidth]{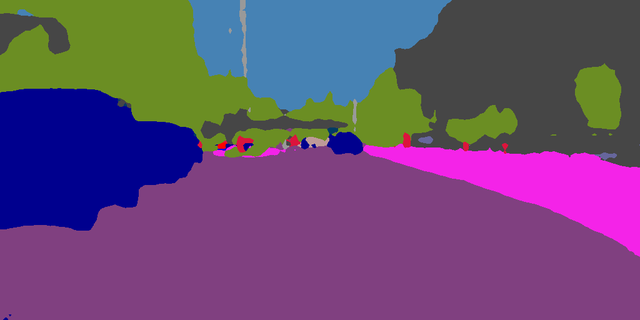}\\
		\includegraphics[width=.19\linewidth]{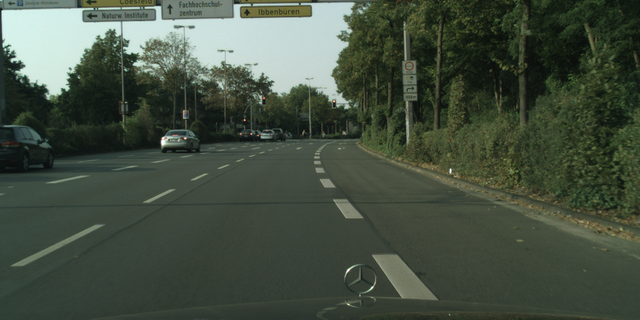}&
		\includegraphics[width=.19\linewidth]{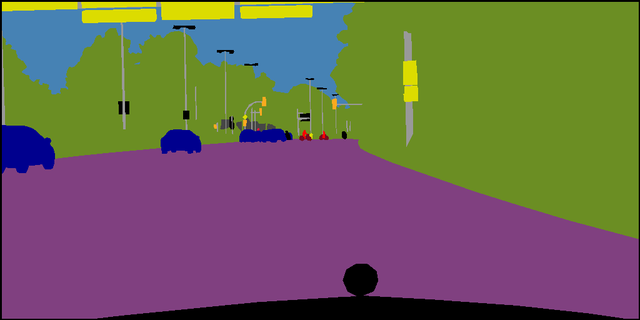}&
		\includegraphics[width=.19\linewidth]{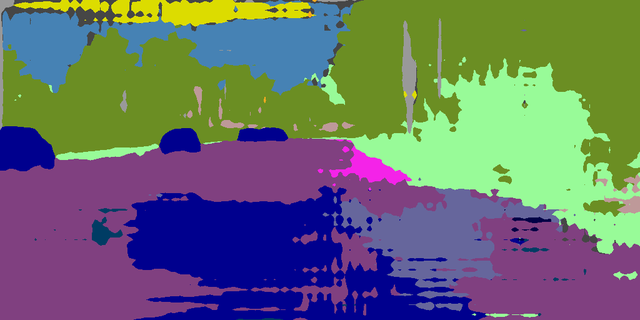}&
		\includegraphics[width=.19\linewidth]{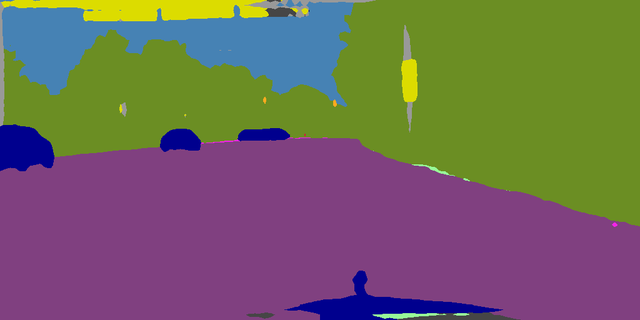}&
		\includegraphics[width=.19\linewidth]{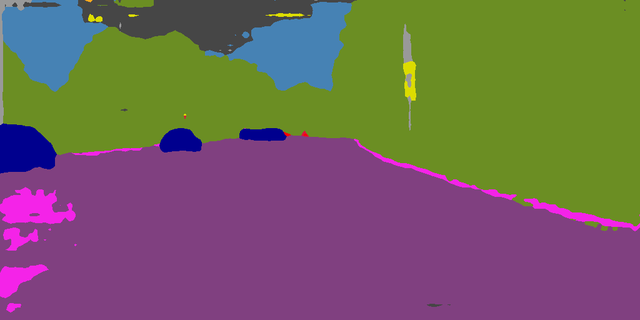}\\
		\makecell{CS\\sample}&\makecell{Ground\\truth}&\makecell{No adaptation\\(GTA)}&\makecell{GTA5$\rightarrow$CS}&\makecell{SYNTHIA$\rightarrow$CS}\\\medskip
	\end{tabular}
	\caption{\textbf{Additional segmentation results.} We take a sample $X_T$ from the Cityscapes validation set and get the predicted segmentation using $M$. Here we show the different results obtainable with $M$ being DeepLabV2. First we show the results obtained with $M$ trained with no adaptation on GTA5, then the results obtained by adapting GTA5 and SYNTHIA.}
	\label{fig:segmentation}
\end{figure}

\end{document}